\pdfoutput=1
\documentclass[dvipsnames]{article}
\usepackage[final]{colm2026_conference}

\usepackage{microtype}
\usepackage{hyperref}
\usepackage{url}
\usepackage{booktabs}
\usepackage{floatrow}
\definecolor{darkblue}{rgb}{0, 0, 0.5}
\hypersetup{colorlinks=true, citecolor=darkblue, linkcolor=darkblue, urlcolor=darkblue,
  pdftitle={Understanding In-context Learning of Addition via Activation Subspaces},
  pdfauthor={Xinyan Hu, Kayo Yin, Michael I. Jordan, Jacob Steinhardt, Lijie Chen}}

\usepackage{latexsym}
\usepackage{amssymb}
\usepackage{amsmath}

\usepackage[T1]{fontenc}

\usepackage[utf8]{inputenc}

\usepackage{inconsolata}

\usepackage{graphicx}
\usepackage{tikz}
\usetikzlibrary{matrix, positioning, arrows.meta}

\renewcommand{\paragraph}[1]{\vspace{0.5em}\noindent\textbf{#1}}

\usepackage{xcolor}
\definecolor{red}{HTML}{CC0000}
\definecolor{blue}{HTML}{0066CC}

\usepackage{wrapfig} 
\usepackage{caption}
\usepackage{subcaption}
\usepackage{enumitem}
\usepackage{threeparttable}
\usepackage{tablefootnote}

\title{Understanding In-context Learning of Addition via Activation Subspaces}

\author{Xinyan Hu\textsuperscript{$*$}\textsuperscript{$1$}, Kayo Yin\textsuperscript{$1$}, Michael I. Jordan\textsuperscript{$1,2$}, Jacob Steinhardt\textsuperscript{\dag}\textsuperscript{$1$}, Lijie Chen\textsuperscript{\dag}\textsuperscript{$1$}
\\
\textsuperscript{$1$}UC Berkeley, \textsuperscript{$2$}INRIA
}

\begin{document}

\maketitle

\begin{abstract}
To perform few-shot learning, language models extract signals from a few input-label pairs, aggregate them into a learned prediction rule, and apply this rule to new inputs. How is this implemented in the forward pass of modern transformer models? To explore this question, we study a structured family of few-shot learning tasks for which the true prediction rule is to add an integer $k$ to the input. We introduce a novel method that localizes the model's few-shot learning ability to only a few attention heads. This method and the findings generalize to four additional task families spanning arithmetic and semantic tasks. We then perform an in-depth analysis of individual heads via dimensionality reduction and decomposition of the heads' output spaces. For example, in Llama-3-8B-Instruct, we reduce the mechanism underlying these tasks to just three attention heads with six-dimensional subspaces, in which four dimensions track the units digit using trigonometric functions with periods $2$, $5$, and $10$, while two dimensions track magnitude using low-frequency components. To deepen our understanding of this mechanism, we also derive a mathematical identity relating the ``aggregator'' and ``extractor'' subspaces of attention heads, allowing us to track the flow of information from individual examples to a final aggregated concept. Our results demonstrate how tracking low-dimensional subspaces of localized heads throughout a forward pass can provide insight into fine-grained computational structures in language models. Our code is available at \url{https://github.com/xyVickyHu/addition-subspaces}.

\end{abstract}

\def\thefootnote{$*$}\footnotetext{Correspondence to: Xinyan Hu \texttt{<xinyanhu@berkeley.edu>}.}\def\thefootnote{\arabic{footnote}}
\def\thefootnote{\dag}\footnotetext{Equal advising.}\def\thefootnote{\arabic{footnote}}

\section{Introduction}

Large language models (LLMs) exhibit in-context learning (ICL) abilities; for instance, they can few-shot learn new tasks from a small number of demonstrations in the prompt. To understand this ability, past works have constructed detailed models of ICL for small synthetic language models~\citep{garg2022can, akyurek2023learningalgorithmincontextlearning, vonoswald2023transformerslearnincontextgradient} as well as coarser-grained analyses of large pretrained models~\citep{ih, tv, fv}. However, little is known about the fine-grained computational structure of ICL for large models.

ICL extracts task information from demonstrations and applies the aggregated information to input queries. Previous work~\citep{fv} constructed a vector (i.e., the ``function'' vector) from demonstrations that encodes the task information---for instance, adding it to the residual stream on zero-shot inputs recovers ICL behavior. However, two questions remain open: (1) How precisely do function vectors encode task information? (2) How do models aggregate information from ICL examples to form such function vectors?

\begin{figure}[t]
    \centering
    \makebox[\linewidth][c]{\includegraphics[width=1.15\linewidth]{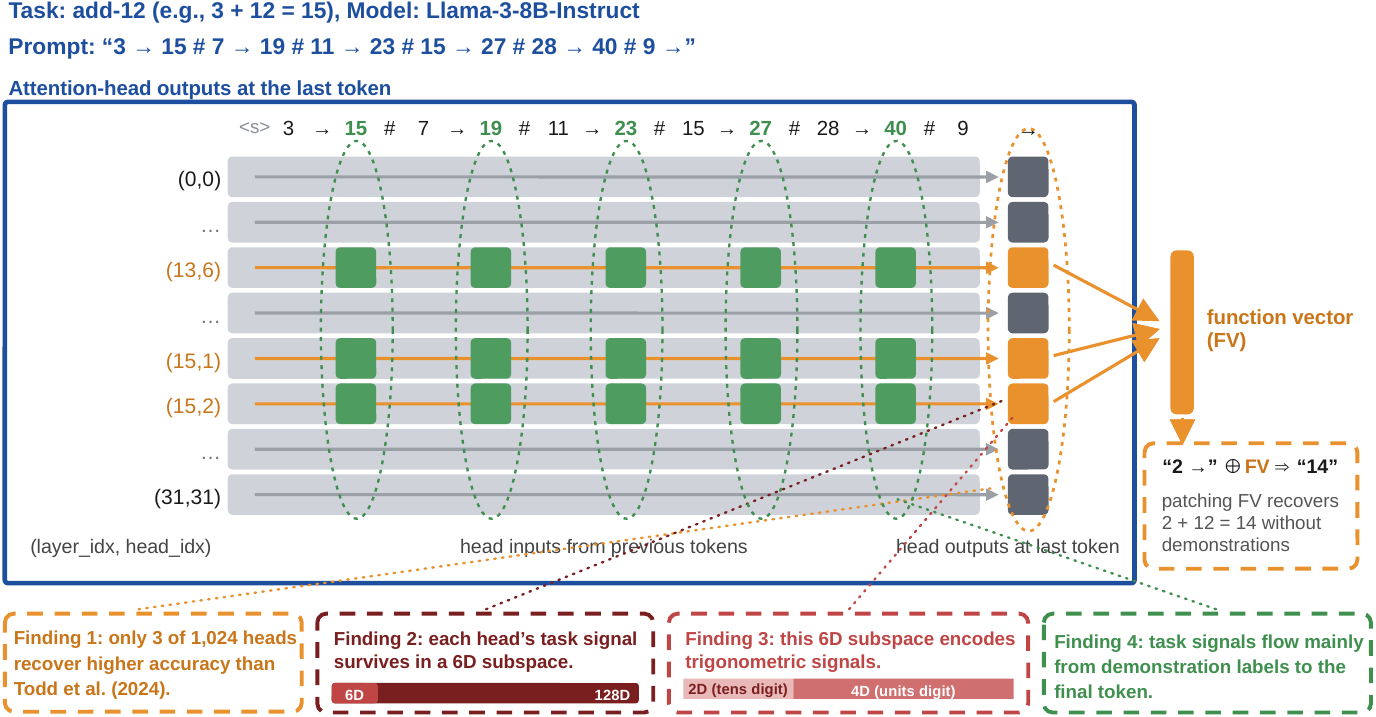}}
    \caption{
    Key findings illustrated on Llama-3-8B-Instruct and a five-shot \textit{add-12} prompt as an example:
    (1) only three attention heads carry most of the \textit{add-k} task signal (\S\ref{sec:id}); (2) each head's signal survives in a six-dimensional subspace of its outputs (\S\ref{sec:6dim}); (3) that subspace trigonometrically encodes $k$ with four dimensions for the units digit and two dimensions for the tens digit (\S\ref{sec:periodic}, \S\ref{sec:4+2dim}); and (4) signals flow mainly from demonstration labels (\S\ref{sec:extract}).}
    \label{fig:main}
\end{figure}

To address these questions, we first develop a novel general pipeline for localizing few-shot ability to a small number of attention heads and low-dimensional subspaces, and furthermore perform a case study for arithmetic tasks (i.e., learning to add a constant $k$ to the input) for mechanistic depth. This family of tasks has the advantage of providing a large number of tasks (different integers $k$) that all share the same input domain, which is helpful to rule out domain-based shortcuts when analyzing ICL mechanisms (\S\ref{sec:model-task}). Arithmetic also has the advantage of being well studied in other (non-in-context) settings~\citep{zhou2024pre, kantamneni2025language}, allowing us to situate our results with other known mechanisms. Finally, most LLMs perform this task reliably (e.g., $87\%$ accuracy for Llama-3 and $90\%$ for Qwen-2.5). Figure~\ref{fig:main} gives an overview of our key findings on this task family.

We first introduce a head-localization pipeline that combines a novel sparse optimization objective with statistical refinements. Our approach is inspired by function vectors~\citep{fv}, which mimic ICL behavior when patched into the residual stream (\S\ref{sec:ap-and-fv}). Whereas \citet{fv} rank heads independently by their \textit{individual} effects on ICL performance, we jointly optimize coefficients $c_h \in [0,1]$ over all heads to identify a sparse, weighted \textit{combination} of head outputs that maximizes intervention accuracy (\S\ref{sec:33head}).
Our method successfully finds a small number of heads that recover most of the ICL performance (\S\ref{sec:3head}); for example, for addition tasks, Llama-3-8B-Instruct achieves 68\% accuracy using function vectors from just three heads (78\% of the original ICL accuracy). Applying the same pipeline to subtraction, multiplication, and the abstractive and extractive tasks from \citet{fv} reveals shared important heads: one head in a middle layer is identified as a main head across all five families, several additional heads recur across arithmetic tasks, and broader head sets are identified for the semantic families (Appendix~\ref{app:generality}).

We next perform a detailed analysis of these heads, through dimensionality reduction and decomposition. We find that the task information in the localized heads is encoded in low-dimensional subspaces across task families. For addition, which we study in depth, these subspaces exhibit an interpretable trigonometric structure. For each of the three main heads in Llama-3-8B-Instruct, we get a 6-dimensional subspace, where 4 dimensions encode periodic functions with periods 2, 5, and 10 (tracking the units digit) and 2 dimensions encode low-frequency functions (tracking the tens digit)~(\S\ref{sec:agg}). This interestingly aligns with recent findings in non-in-context settings where addition is also encoded by trigonometric~\citep{zhou2024pre} or helical functions~\citep{kantamneni2025language}, suggesting a deeper relation between non-ICL and ICL machinery.

Finally, we further study the flow of information across tokens, by deriving a general mathematical relation between ``extractor'' and ``aggregator'' subspaces, building on the second-order logit lens~\citep{gandelsman2025interpretingsecondordereffectsneurons}. This lets us study the task-related information that heads extract from each token. For example, in Llama-3-8B-Instruct we find that task information is mostly extracted from label tokens (\S\ref{sec:individual-extractor}).

In summary, we found that task-specific information in ICL can be localized to a small number of attention heads and low-dimensional subspaces with structured patterns. We also developed a way to trace the task information flowing from the ``extractor'' subspaces to the ``aggregator'' subspaces. Our findings show how even in large models, ICL mechanisms can be localized to specialized activation subspaces from a small number of heads that extract, represent, and aggregate information in interpretable ways. Our key contributions are: (1) a novel method to identify relevant attention heads for ICL; (2) a precise mathematical relation between signals from earlier tokens and the output token; and (3) a focused analysis of ICL addition that reverse-engineers rich latent structures and sophisticated computational strategies in LLMs.

\section{Preliminaries}\label{sec:prelim}
\vspace{-0.3em}
\subsection{Model and Task}\label{sec:model-task}
\vspace{-0.2em}

\textbf{Model.} We focus on Llama-3-8B-Instruct~\citep{grattafiori2024llama3} in the main body, with 32 layers, 32 attention heads per layer, and residual dimension 4096. We denote each head as a tuple (layer index, head index); both indices range from 0 to 31. To broaden our analysis in model size, training type, and model family, we also experiment on Phi-4-14B~\citep{abdin2024phi4} and Qwen-2.5-7B~\citep{qwen2024qwen25} and report results in the appendix. The results are directionally similar across models.

\paragraph{Task.} 
We consider five families of ICL tasks: three arithmetic families (addition, subtraction, and multiplication) and the abstractive and extractive families of \citet{fv}, which include abstractive tasks like product-company (``iPhone 5$\rightarrow$Apple'') and extractive tasks like animal-vs-object (``bear, ukulele, blanket$\rightarrow$bear''); Table~\ref{tab:task-families} in Appendix~\ref{app:generality-tasks} specifies all five families. Our localization pipeline applies to each of them (Appendix~\ref{app:generality}); for mechanistic depth, we center our case study on addition, a structured family of ICL tasks that we call \textit{add-$k$}. For a constant $k$, the add-$k$ task is to add $k$ to a given input integer $x$ to predict $y = x + k$. In an $n$-shot ICL prompt, the model is given $n$ demonstrations of the form  ``$x_i \rightarrow y_i$'' with $y_i = x_i + k$, concatenated using the separator ``$\#$'', followed by a query ``$x_q \rightarrow$'' (see the top row of Figure~\ref{fig:main} for an illustration). A key advantage of this family of tasks is that ICL prompts for different tasks only differ in $k$ (i.e., $y_i-x_i$) but not in the input domain, enabling us to isolate task information from input content so as to dissect the ICL mechanism at a finer granularity. 

This advantage motivates our focus on \textit{add-$k$}: in semantic tasks such as product-company, the input domain itself leaks information about the task~\citep{min2022rethinkingroledemonstrationsmakes}, making it difficult to distinguish whether success stems from extracting the task rule or from leveraging domain-specific associations. For \textit{add-$k$}, the query $x_q$ alone provides no information about $k$, so the model must infer $k$ from the demonstrations, which cleanly isolates the key components of the ICL mechanism. Moreover, addition is well studied in non-in-context settings~\citep{zhou2024pre, kantamneni2025language}, so studying it under ICL fills a gap and allows us to compare our findings with the known non-ICL mechanisms.

We construct data for the \textit{add-$k$} task by varying $x \in [1, 100]$ and $k \in [1, 30]$ (thus $y \in [2, 130]$) since Llama-3 models can reliably solve addition in this range. We consider three types of prompts: (1) \textit{five-shot ICL prompt}, where all five demonstrations satisfy $y_i = x_i+k$ for a fixed $k \in [1, 30]$ (also called \textit{add-k} ICL); (2) \textit{mixed-$k$ ICL prompt}, where the demonstrations are $y_i = x_i+k_i$ for possibly different $k_i$ values; and (3) \textit{zero-shot prompt}, where there are no demonstrations: the prompt is ``$x_q\rightarrow$'' for some $x_q$.
We mostly study five-shot ICL prompts and use them to generate function vectors (\S\ref{sec:ap-and-fv}). We also examine signals extracted from demonstrations in mixed-$k$ ICL prompts, a variant with mistaken demonstrations, in \S\ref{sec:individual-extractor}. We use zero-shot prompts to evaluate heads and function vectors (\S\ref{sec:ap-and-fv}). 

\vspace{-0.2em}
\subsection{Activation Patching and Function Vectors}\label{sec:ap-and-fv}
\vspace{-0.2em}

Next, we briefly review \emph{activation patching}, a common interpretability technique that will be used throughout the paper, and the \emph{function vector}, a construction that we will use to identify important heads for our ICL tasks.

\paragraph{Activation patching.} Activation patching is performed by taking the activations of a model component when the model is run on one prompt, then patching in these activations when the model is run on a different prompt. Patching can either \emph{replace} the model’s base activations or \emph{add to} them; we will primarily consider the latter.

Specifically, if $z_l$ is the original value of the residual stream at layer $l$ at the final token position ``$\to$'', 
we patch in the replacement $z_l + v$, where $v$ is 
constructed from activations on a different input prompt. We choose the layer $l$ at one-third of the network's depth and construct $v$ from ``function vector'' heads, following \citet{fv}, as described next.

\paragraph{Function vectors.} Function vectors $v_k$ are vectors constructed from the outputs of chosen attention heads, designed so that adding~$v_k$ to the residual stream of a zero-shot prompt approximates the effect of the five-shot \textit{add-k} task. For example, $v_k$ might be the average output of one or more attention heads across a set of five-shot \textit{add-$k$} examples. We define the \textit{intervention accuracy} of $v_k$ as the average accuracy obtained on zero-shot prompts when $v_k$ is added to the residual stream across all tasks $k$. Because $k$ is independent of the input queries, this metric captures how effectively $v_k$ encodes task-specific information about $k$.
For comparison, we define \textit{clean accuracy} as the accuracy on five-shot prompts without any intervention, also averaged across all tasks $k$.

In more detail, consider an attention head~$h$: let $h(p)$ denote the output of head~$h$ on prompt~$p$ at the last token position, and define $h_k$ as the average of $h(p)$ across all five-shot \textit{add-k} prompts.\footnote{Similar to \citet{fv}, we approximate this average using $100$ randomly generated five-shot \textit{add-k} prompts, where each prompt contains random demonstration inputs~$x_i$ and the final query input~$x_q$ takes each integer in $[1, 100]$ exactly once. } \citet{fv} identified a subset $\mathcal{H}$ of attention heads (for a different set of tasks) such that the vector $v_k=\sum_{h\in \mathcal{H}} h_k$ has high intervention accuracy---that is, adding it to the residual stream of zero-shot prompts effectively recovers few-shot performance. 

Building on this framework, we consider multiple ways to construct $v_k$: (1) the task-specific mean~$h_k$ over the \textit{add-k} task (as described above); (2) the overall mean~$\overline{h}$ (i.e., average across all $k$), which removes task-specific information about $k$; or (3) the specific value~$h(p)$ on an individual prompt~$p$. Throughout the paper, we call $\overline{h}$ the \emph{mean-ablation} of head~$h$ and call $h_k$ the \emph{head vector} of $h$ with respect to $k$. Beyond these three choices, we sometimes project a head's output onto a lower-dimensional subspace or scale it by a coefficient. Unlike~\citet{fv}, we recover a different set of heads by solving a novel global optimization problem (\S\ref{sec:33head}) and by systematic ablation studies (\S\ref{sec:3head}), achieving higher performance.

\begin{figure*}[t]
    \centering
    \includegraphics[width=0.9\linewidth]{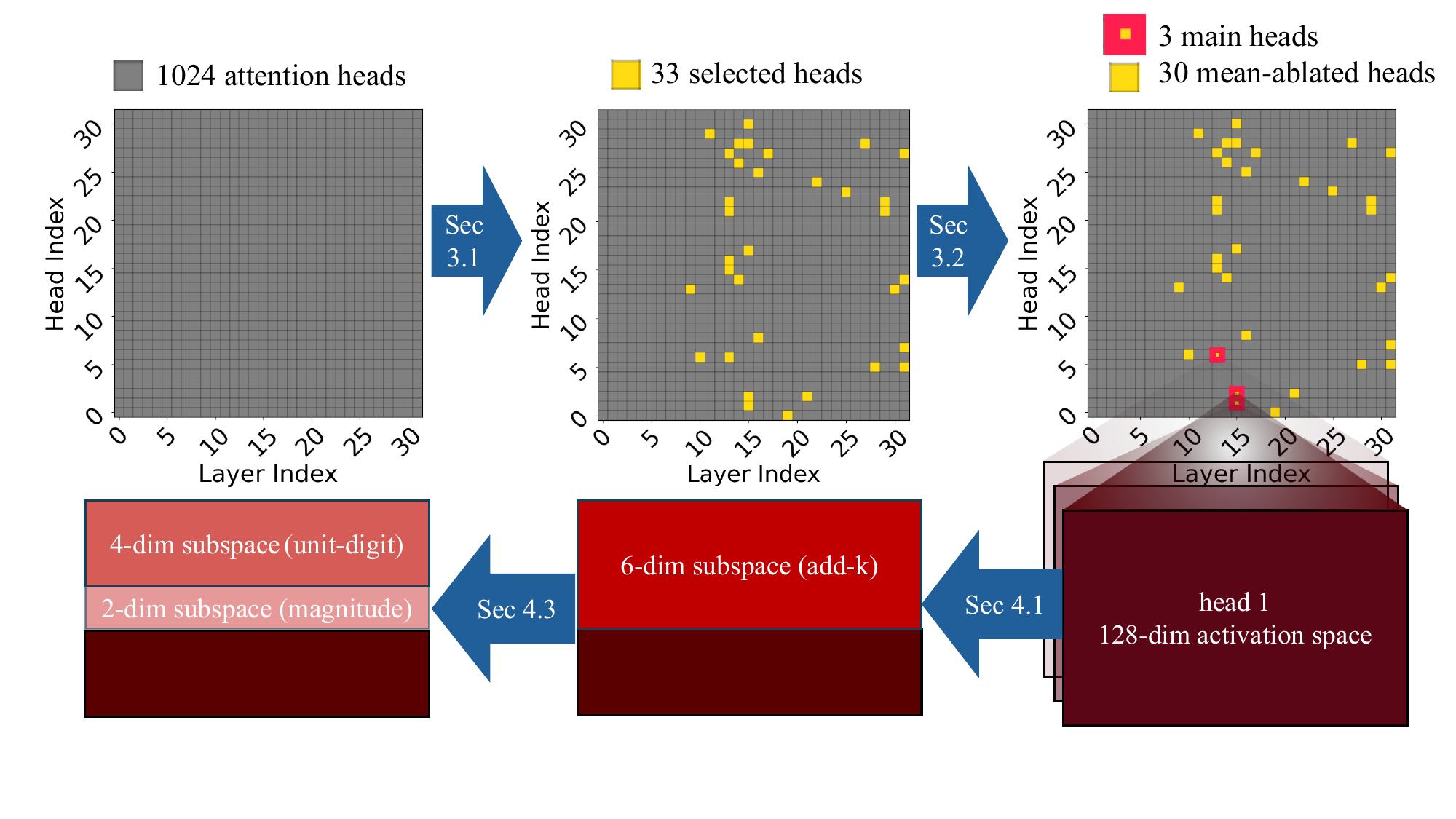}
    \caption{The chain of localization in \S\ref{sec:id} and \S\ref{sec:agg}. We first select $33$ attention heads (out of 1024) via a global optimization method (\S\ref{sec:33head}), then narrow down to 3 main heads while mean-ablating the remaining 30 (\S\ref{sec:3head}). We next study the structure of the representation of each main head by localizing it to a six-dimensional subspace (\S\ref{sec:6dim}), and decompose it into a four-dimensional subspace encoding the units digit and a two-dimensional subspace encoding the coarse magnitude (i.e., the tens digit) (\S\ref{sec:4+2dim}).
    }
    \label{fig:sec3-4}
\end{figure*}

\vspace{-0.3em}
\section{Identifying Three Aggregator Heads}\label{sec:id}
\vspace{-0.3em}
To understand the mechanism behind the \textit{add-k} ICL task, we first need to find out what model components are responsible for performing it.
In this section, we find that three heads do most of the work for \textit{add-k}, by first solving an optimization problem to select $33$ heads out of $1024$ (\S\ref{sec:33head}), and then narrowing down to three main heads via systematic ablations (\S\ref{sec:3head}); this process is illustrated in the first row of Figure~\ref{fig:sec3-4}. \S\ref{sec:necessity} then validates the necessity of the three main heads in the original five-shot setting.

\vspace{-0.2em}
\subsection{Selecting Heads via Sparse Optimization}\label{sec:33head}
\vspace{-0.2em}
We will search for a set of heads that store the information for the \textit{add-k} task, in the sense that their output activations yield good function vectors for \textit{add-k} (\S\ref{sec:ap-and-fv}).

\paragraph{Formulating the sparse optimization.} 
More formally, define $v_k(c) = \sum_{h} c_h \cdot h_k$, the sum of head outputs weighted by $c$, where $h$ goes over all $1024$ heads in the model. (Recall that $h_k$ is the output of head $h$ averaged across a large dataset of five-shot \textit{add-k} prompts.) We will search for a sparse coefficient vector $c \in [0,1]^{1024}$ such that adding $v_k(c)$ to the residual stream of a zero-shot prompt achieves high accuracy on the \textit{add-k} task.

Let $\ell(x_q, y_q; v)$ be the cross-entropy loss between $y_q$ and the model output when intervening with $v$ on the input ``$x_q\to$'' (i.e., replacing $z_l$ with $z_l + v$ in the forward pass, where $z_l$ is the layer-$l$ residual stream at the last token). We optimize $c$ with respect to the loss
\begin{equation}
    \mathcal{L}(c) = \mathbb{E}_{k, x_q}[\ell(x_q, x_q + k; v_k(c))] + \lambda \|c\|_1,
    \label{eq:loss}
\end{equation}
where the regularization term with weight $\lambda$ promotes sparsity.

\textbf{Training details.} 
We randomly select $25$ \textit{add-k} tasks for training and in-distribution testing, using the remaining five for out-of-distribution testing. For each task, we generate $100$ zero-shot prompts ``$x_q\rightarrow$'' with $x_q \in [1,100]$ and target $x_q+k$. We split the $25$ tasks' data into train/val/test sets in proportions $0.7/0.15/0.15$. We use AdamW with learning rate $0.01$, batch size $128$, and regularization $\lambda=0.05$, clipping $c$ to $[0,1]$ after each gradient step. 

\textbf{Results: the selected heads.} We get coefficients that achieve high intervention accuracy. In particular, on the five held-out tasks, the function vector weighted by the final-epoch coefficients achieves an intervention accuracy of $0.812$, close to the clean accuracy of $0.870$.
To identify important heads, we inspect the coefficient values at the end of training for each layer and head index. We find $33$ heads have coefficients $c_h$ greater than $0.2$, most ($21$) of which are one. In contrast, the other ($991$) coefficients are all smaller than $0.03$, most ($889$) of which are zero. We call these $33$ heads \textit{selected heads}, denoted $\mathcal{H}_{\text{sel}}$.

More generally, we do not hand-pick this subset: we sort the trained coefficients (averaged over the last three checkpoints to smooth end-of-training jitter) in decreasing order and cut at the largest gap between consecutive values, keeping the heads above it. This parameter-free rule recovers exactly these $33$ heads for \textit{add-k}, and is the selection rule we apply to every other task family, model, and prompt format we study (results in Appendix~\ref{app:generality}).

\textbf{Comparison to previous approach.} We compare our optimization approach with the previous method from \citet{fv} for identifying important heads, which selects heads based on average indirect effect (AIE). To perform a fair comparison, we construct our function vector by summing the outputs of our selected heads directly without weighting by coefficients, matching their methodology. Using this construction, we achieve an intervention accuracy of $0.798$, close to the clean accuracy of $0.870$, indicating that our $33$ heads capture most of the necessary information for the \textit{add-k} task. In contrast, selecting the top 33 heads according to AIE yields a much lower intervention accuracy of $0.414$.\footnote{\citeauthor{fv}~select roughly the top $2\%$ of heads in their work ($k=20$ for Llama-3-8B-Instruct), which yields an even lower accuracy of $0.124$ in our setting.} 

Intuitively, a key difference is that AIE ranks each head independently over all attention heads, whereas our optimization selects the heads jointly, accounting for their combined effects. In the next subsection, we further narrow down to an even smaller number of effective heads via ablations within this subset.

\vspace{-0.2em}
\subsection{Further Narrowing to Three Heads via Ablations}\label{sec:3head}
\vspace{-0.2em}

\paragraph{Mean-ablating head subsets.}
We suspect that many selected heads may carry signals at the task-family level (such as ensuring the output appears as a number) without encoding the task-specific value of $k$. To test this, we perform \emph{mean-ablations}: replacing a head's task-specific vector $h_k$ with its cross-task mean $\overline{h}$ (\S\ref{sec:ap-and-fv}). For a retained set of heads $\mathcal S\subseteq\mathcal H_{\mathrm{sel}}$, we construct
\begin{equation}
    v_k(\mathcal S)
    =
    \sum_{h\in\mathcal S} h_k
    +
    \sum_{h\in\mathcal H_{\mathrm{sel}}\setminus\mathcal S}
    \overline{h}.
    \label{eq:mean-ablation}
\end{equation}
Mean-ablating a subset of selected heads from \S\ref{sec:33head} gives us a meaningful task-agnostic base against which to measure the task-specific contribution of the complement $\mathcal S$.

\paragraph{Amplifying an individual head's effect for comparison.}
To measure the contribution of an individual selected head $h$, we retain and scale its task-specific vector while mean-ablating all other selected heads:
\begin{equation}
    v_k(h,c)
    =
    c\,h_k
    +
    \sum_{g\in\mathcal H_{\mathrm{sel}}\setminus\{h\}}
    \overline{g},
    \qquad c\geq 0.
    \label{eq:recovery-scan}
\end{equation}
At $c=0$, the candidate head is omitted, yielding a leave-one-out baseline; increasing $c>0$ restores and scales its task-conditioned signal. The resulting accuracy curve therefore measures how much task-specific information that head adds on top of a reasonable base. We also intentionally increase the range of $c$ from $[0,1]$ in equation~\ref{eq:loss} to $\{0,\ldots,20\}$ in equation~\ref{eq:recovery-scan} to make the task-specific effect of each head more distinguishable. 

\paragraph{Refining head selection via two general criteria.}
Using equation~\ref{eq:recovery-scan}, we want to identify two kinds of heads for task-specific information: (1) heads that are important at all, and (2) a core set of the most important heads. 

For (1), we identify heads that have a statistically significant positive effect on accuracy, and denote them as \textit{significant} heads. We perform an ablation comparing $c = 0$ and $c>0$ for each selected head and run a one-sided McNemar test~\citep{mcnemar1947note} of significance. This often yields a much smaller subset than the initial set of selected heads while almost retaining their accuracy.

For (2), we are sometimes interested in finding a small set of heads that still recover a reasonable fraction of the performance. As a heuristic for finding these, we identify significant heads whose best accuracy gain over $c=0$ is at least one quarter of the largest gain achieved by any significant head, and denote them as \textit{main} heads. In practice, setting other thresholds or an eye test may also work, but these criteria get reliable results on our benchmarks (detailed methods and results in Appendices~\ref{app:refine} and~\ref{app:generality}).

\paragraph{Results: narrowing to three heads.}
For \textit{add-k}, these criteria produce $13$ significant heads and, among them, three main heads: $(15,2)$, $(15,1)$, and $(13,6)$. Their accuracy curves are sharply separated from those of the rest (peak accuracies in Table~\ref{tab:scan-peaks}). Keeping only these three task-specific vectors and mean-ablating the other $30$ selected heads (equation~\ref{eq:mean-ablation}) achieves an intervention accuracy of $0.680$, $78\%$ of the clean five-shot accuracy. We therefore focus on these three heads in the following sections.

\begin{table}[t]
\centering
\small
\begin{tabular}{lcc}
\toprule
Head & Optimal scale $c$ & Accuracy \\
\midrule
$(15,2)$ & $6$ & $0.866$ \\
$(15,1)$ & $6$ & $0.817$ \\
$(13,6)$ & $5$ & $0.693$ \\
Any other selected head & per head & $\leq 0.170$ \\
\bottomrule
\end{tabular}
\caption{{Single-head intervention-accuracy peaks: the intervention accuracy of $v_k(h,c)$ from equation~\ref{eq:recovery-scan} at the head's optimal scale $c\in\{0,\ldots,20\}$. The last row reports the maximum over the remaining $30$ selected heads, each at its own optimal scale. For reference, the clean five-shot accuracy is $0.870$.}}
\label{tab:scan-peaks}
\end{table}

\vspace{-0.2em}
\subsection{Validating Necessity of the Three Main Heads}\label{sec:necessity}
\vspace{-0.2em}
The experiments above show that the task-specific components of the three main heads, together with the mean-ablated base, recover much of the \textit{add-k} signal on zero-shot prompts. We next test whether these heads are also necessary in the model's ordinary five-shot computation. Mean-ablating the three heads on randomly generated five-shot prompts sharply reduces accuracy from $0.87$ to $0.43$. In contrast, mean-ablating $20$ random sets of three other selected heads has little effect: $95\%$ of these sets leave accuracy at or above $0.86$. Thus, the three heads do not merely carry useful signals; they are necessary for a large fraction of the model's ordinary five-shot performance.

\vspace{-0.3em}
\section{Characterizing the Aggregator Subspace}\label{sec:agg}
\vspace{-0.3em}
For the model to perform \textit{add-k}, it must infer the task information (the number $k$) from the ICL demonstrations. Our next goal is to understand how task information is represented in the activation space. Since the three aggregator heads we identified carry most of this information, we can focus on analyzing their representation space. 

\begin{figure}[t]
    \centering

    \begin{subfigure}[b]{0.30\linewidth}
        \centering
        \includegraphics[width=\linewidth]{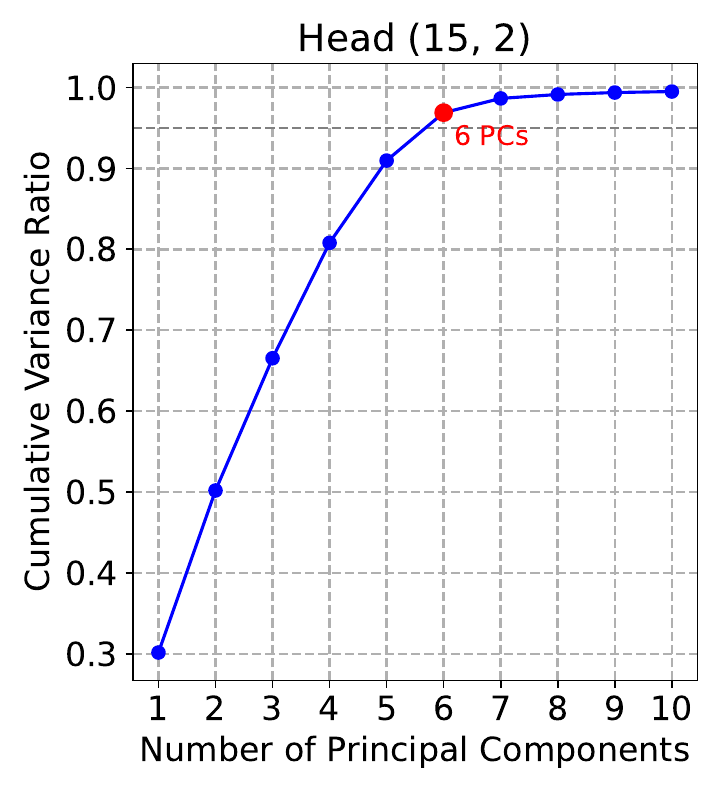}
        \caption{\textbf{Subspace localization.} 6 PCs explain $97\%$ of variance.}
        \label{fig:head1_pca_variance}
    \end{subfigure}\hfill
    \begin{subfigure}[b]{0.68\linewidth}
        \centering
        \includegraphics[width=1.05\linewidth]{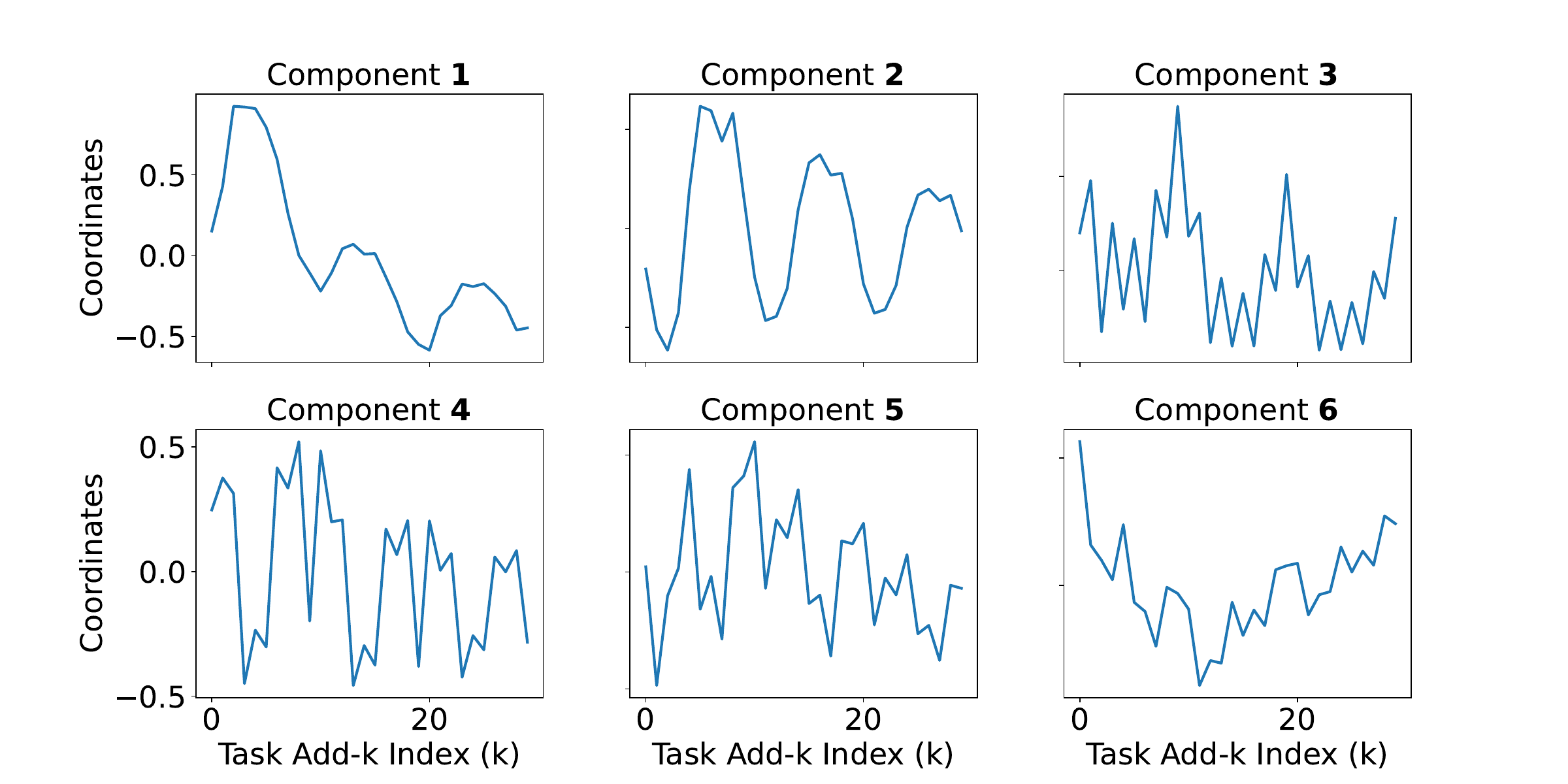}
        \caption{\textbf{Original coordinates.} Head 1's PC coordinates as functions of $k$; first five PCs show partially periodic patterns.}
        \label{fig:head1_pc_coords}
    \end{subfigure}

    \vspace{0.1em}

    \begin{subfigure}[b]{0.54\linewidth}
        \centering
        \includegraphics[width=1\linewidth]{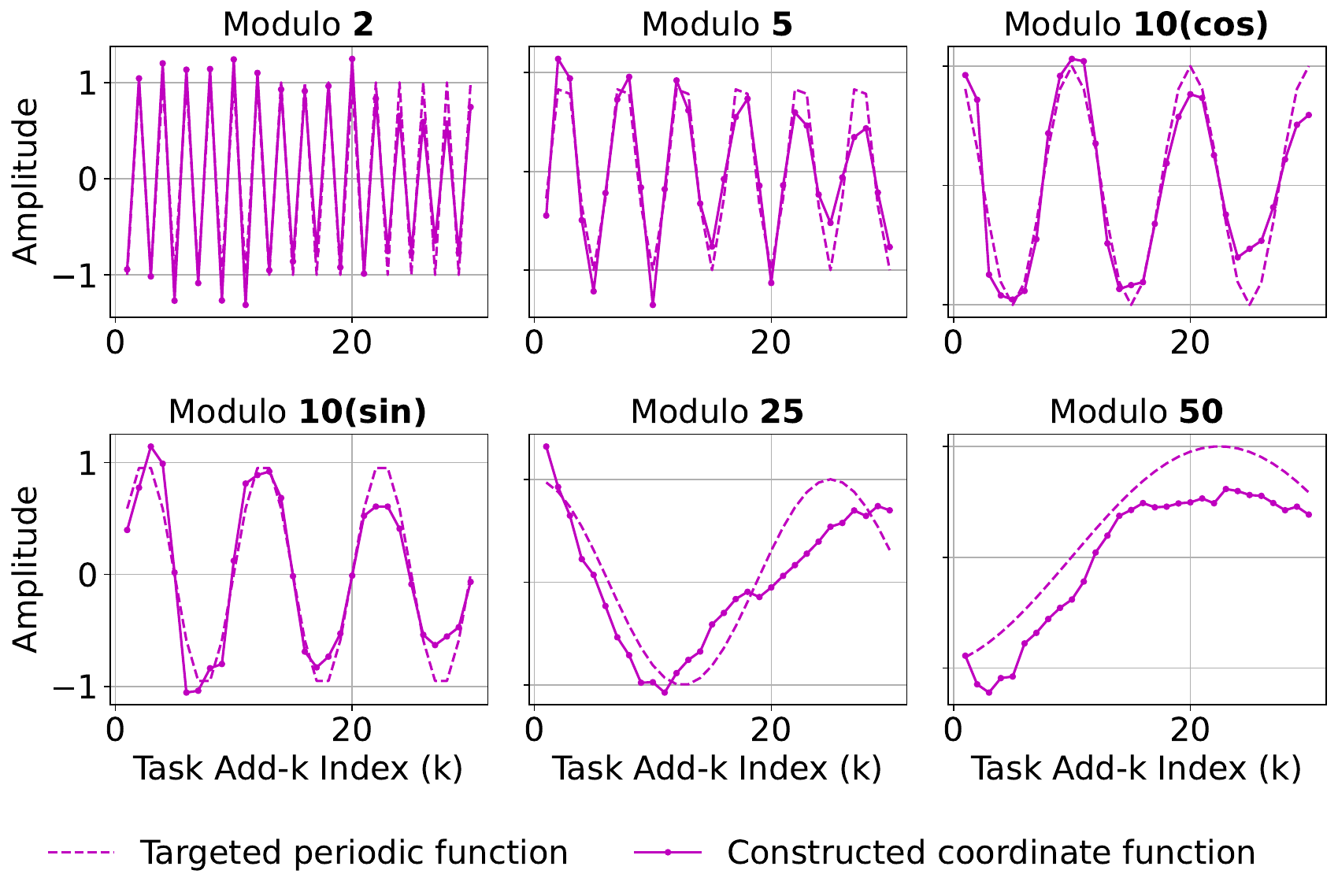}
        \caption{\textbf{Subspace rotation.} Linear transformation of the 6 PCs yields feature directions fitting trigonometric curves at periods $2,5,10,25,50$.}
        \label{fig:head1_trig_fit}
    \end{subfigure}\hfill
    \begin{subfigure}[b]{0.44\linewidth}
        \centering
        \includegraphics[width=1.12\linewidth]{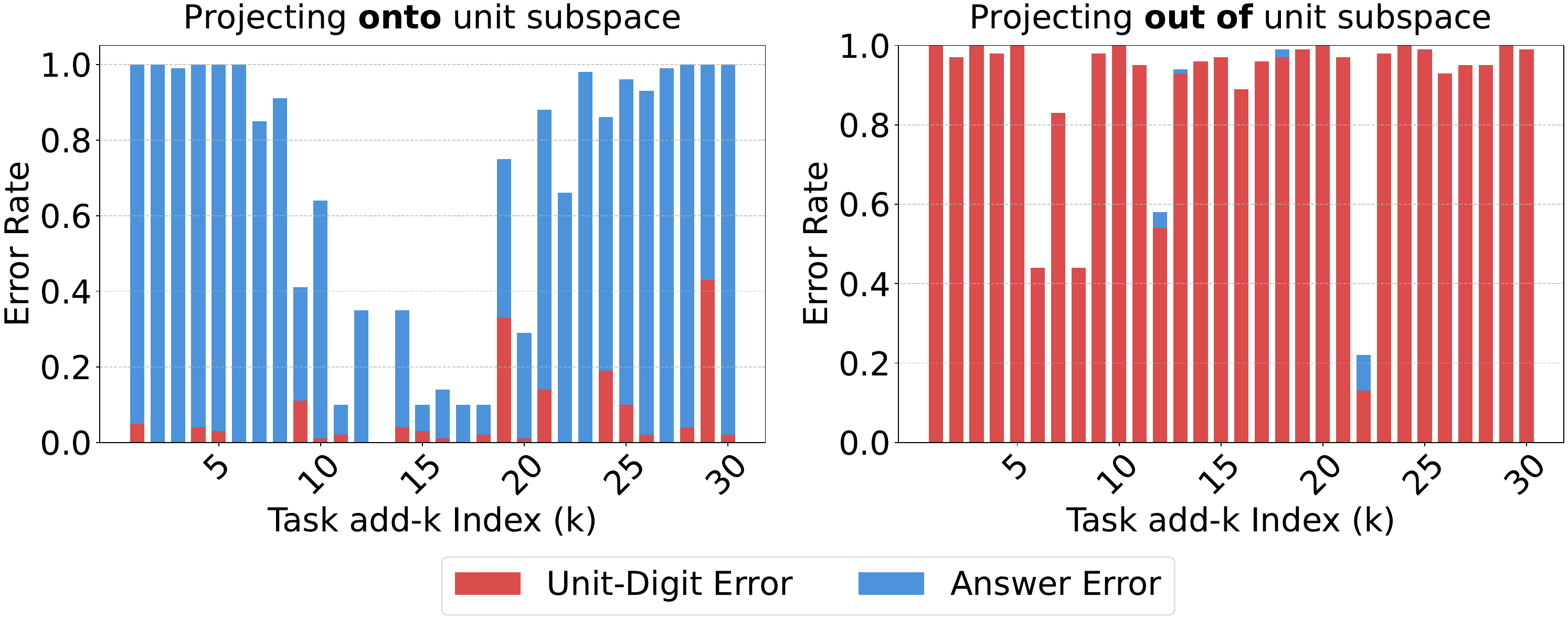}
        \caption{\textbf{Subspace decomposition.} Projecting \emph{onto} the unit subspace preserves unit-digit accuracy while degrading overall accuracy; projecting \emph{out} destroys the unit-digit signal.}
        \label{fig:head1_unit_proj}
    \end{subfigure}

    \caption{Characterizing the aggregator subspace, illustrated on head 1, $(15,2)$ (other heads and models in Appendices~\ref{app:6dim}--\ref{app:4+2dim}):
    (a) PCA localizes task information to a 6-d subspace;
    (b) PC coordinates show periodic dependence on $k$;
    (c) rotated feature directions encode clean trigonometric patterns;
    (d) causal steering validates the unit-digit subspace decomposition.}
    \label{fig:head1_agg_subspace}
\end{figure}

In this section, we dissect their activation spaces in three stages (illustrated in Figure~\ref{fig:head1_agg_subspace}): (1) \textbf{Localize} a six-dimensional task subspace in each head via principal component analysis (PCA) (\S\ref{sec:6dim}); (2) \textbf{Rotate} this subspace into \emph{feature directions} aligned with sinusoidal patterns across $k$ (\S\ref{sec:periodic}); (3) \textbf{Decompose} the six-dimensional space into a four-dimensional \emph{unit-digit} subspace and a two-dimensional \emph{magnitude} subspace that separately encode the units and tens of the task signal $k$ (\S\ref{sec:4+2dim}).

\vspace{-0.3em}
\subsection{Localizing to Six-dimensional Subspace} \label{sec:6dim}
\vspace{-0.2em}
To reduce the $128$-dimensional activation space per head to a more tractable space, we first perform PCA on the $30$ head vectors and find that just six directions explain $97\%$ of the task variance (Figure~\ref{fig:head1_pca_variance}). We then check that the function vectors found earlier remain effective after projecting onto the subspace. Specifically, we take $\mathcal S$ in equation~\ref{eq:mean-ablation} to be the three main heads and replace each of these head vectors $h_k$ with its projection $\tilde{h}_k$ onto the subspace, yielding $\tilde{v}_k=\sum_{h\in\mathcal S} \tilde{h}_k+\sum_{h\in\mathcal H_{\mathrm{sel}}\setminus\mathcal S} \overline{h}$. We find that $\tilde{v}_k$ has intervention accuracy $0.636$, which is close to the intervention accuracy of $0.680$ before projection~(Tables~\ref{tab:nested-headsets} and~\ref{tab:generality-models}). Thus, we confine our study to the concise six-dimensional subspace for each head. Both the low dimensionality and the causal sufficiency of the projection generalize across the task families, models, prompt formats, and shot counts we study (Appendix~\ref{app:generality-subspace}).

\vspace{-0.3em}
\subsection{Identifying Feature Directions Encoding Periodic Patterns}\label{sec:periodic}
\vspace{-0.2em}
To understand how each head's six-dimensional subspace represents task information, we examine head vector coordinates (inner products with principal components (PCs)) as functions of~$k$, revealing partially periodic patterns in the first five PCs (Figure~\ref{fig:head1_pc_coords}). 

This motivates us to linearly transform the six PCs to find directions encoding pure periodic patterns. Mathematically, if we find a linear transformation of the six PC-coordinate functions that fits trigonometric functions, then applying the transformation to the PCs yields six directions whose coordinate functions encode the periodicity. 

To find trigonometric functions to fit, we search over different periods and phases and perform least squares regression. We find six trigonometric functions at periods $2$, $5$, $10$, $10$, $25$, and $50$ expressible as functions of the top 6 PCs with low regression error (Figure~\ref{fig:head1_trig_fit}). We apply the resulting linear transformation to the six PCs to obtain directions encoding these six pure periodic patterns, which we call \textit{feature directions}.

\vspace{-0.3em}
\subsection{Decomposing into Subspaces Encoding Subsignals}\label{sec:4+2dim}
\vspace{-0.2em}
Leveraging the feature directions identified previously, we decompose the head activation subspace into lower-dimensional components that separately encode different subsignals relevant to the task---in this case, the units digit and tens digit.

By construction, the coordinate function of each feature direction (viewed as a linear projection of $h_k$) is a periodic function of $k$. Mathematically, a feature direction with period $T$ carries task information from the head vectors with ``modulo $T$''. Based on this, we hypothesize: (i) the feature direction corresponding to period two, which we call the ``parity direction'', encodes the parity of $k$ in the \textit{add-k} task; (ii) the subspace spanned by the feature directions with periods $2,5,10$, which we call the ``unit subspace'', encodes the units digit of $k$; (iii) the subspace spanned by the directions with periods $25,50$, which we call the ``magnitude subspace'', encodes the coarse magnitude (i.e., the tens digit) of $k$. 

We verify these hypotheses through causal intervention. We establish (1) \textbf{sufficiency} by showing that projecting a head vector \textit{onto} the subspace preserves the relevant task signal; and (2) \textbf{necessity} by showing that projecting a head vector \textit{out of} the subspace (i.e., onto its orthogonal complement) destroys the relevant task signal. We show experimental results for the unit-digit subspace in Figure~\ref{fig:head1_unit_proj} and defer complete results to Appendix~\ref{app:4+2dim}.

\vspace{-0.3em}
\section{Signal Extractors of ICL Demonstrations}\label{sec:extract}
\vspace{-0.3em}
Having localized most of the task signal to three heads and their six-dimensional subspaces, we now analyze \emph{how} the model extracts it from the ICL demonstrations. We find: (1) the label tokens are the main transport channel, carrying $96$--$99\%$ of each main head's task signal (\S\ref{sec:individual-extractor}); and (2) each demonstration $x_i\rightarrow y_i$ contributes a signal $y_i-x_i$ in the subspace.
\vspace{-0.3em}
\subsection{Tracing the Subspace Back to Previous Tokens}
\vspace{-0.2em}
We begin with a mathematical observation that lets us trace the subspace at the final token back to corresponding subspaces at earlier tokens. A head $h$'s output at the last token of a prompt $p$ decomposes linearly over the prompt's tokens:
    $h(p) \;=\; \sum_{t\in p} \alpha_t \, O_h V_h \, z_t,$
where $\alpha_t$ is the attention weight from the last token to each token $t$, satisfying $\sum_{t \in p} \alpha_t=1$, $z_t$ is the residual stream input to the head $h$ at token $t$, and $V_h$ and $O_h$ are the head's value and output matrices. Intuitively, the head's output is an attention-weighted sum of transformed residual streams from previous tokens, so the signal extracted from token $t$ is $O_hV_hz_t$.

Letting $W_h$ denote the projection onto the head's six-dimensional subspace (\S\ref{sec:6dim}), the projected task signal at the final token, $W_h\,h(p)$, thus decomposes into per-token contributions $\alpha_t\,W_hO_hV_h z_t$, each lying in the image of $W_hO_hV_h$.

\vspace{-0.2em}
\subsection{Signal extraction from each demonstration}\label{sec:individual-extractor}
\vspace{-0.2em}
To understand how demonstrations contribute to the model's generation at the last token, we identify which tokens contribute most, then examine what information they provide. By the analysis above, each token's task-signal contribution through head $h$ is $\alpha_t\,W_hO_hV_h z_t$, which factors into (1) \textbf{extracted information:} $W_hO_hV_h z_t$, the residual stream input projected into the relevant subspace; and (2) \textbf{aggregation weight:} the attention weight $\alpha_t$. On a random mixed-$k$ ICL prompt (with possibly conflicting $k_i$ values), both the norm of the extracted information and the aggregation weight peak at the label tokens $y_i$ (Appendix~\ref{app: individual-extractor}).

To quantify this on standard five-shot prompts, we project each token's contribution $\alpha_t\,O_hV_h z_t$ and the total output $h(p)$ onto the head vector $h_k$ of the prompt's task; the ratios are signed per-token shares that sum to one by linearity. Averaged over the $25$ training \textit{add-k} tasks, the label tokens contribute $96$--$99\%$ of each main head's task signal (Figure~\ref{fig:label_share} in Appendix~\ref{app: individual-extractor}), while input, separator, and query tokens contribute almost nothing.

Finally, we ask \emph{what} information these tokens carry: for each task $k \in \{1,\ldots,30\}$, we measure the inner product $\langle W_hO_hV_hz_t,\, \tilde{h}_k/\|\tilde{h}_k\| \rangle$ between the extracted information and the unit-normalized projected head vector of \S\ref{sec:6dim}. At each label token $y_i$ of a random mixed-$k$ prompt, this inner product peaks at $k = y_i - x_i$ (Figure~\ref{fig:extractor_15_2}, Appendix~\ref{app: individual-extractor}): the model extracts $y_i-x_i$ from each demonstration $x_i\rightarrow y_i$.

\vspace{-0.3em}
\section{Related Work}
\vspace{-0.3em}
Our work builds on interpretability research that aims to uncover circuits and internal computations of language models.
Many studies focus on synthetic tasks or small models~\citep{nanda2023progress, bietti2023birthtransformermemoryviewpoint, reddy2023mechanisticbasisdatadependence, singh2024needsrightinductionhead}. For large pretrained LLMs, some provide coarser-grained analyses~\citep{fv, ih, tv}, while others target particular model families and task classes for finer-grained insights~\citep{hanna2023doesgpt2computegreaterthan, feng2023language, wu2023interpretability, zhou2024pre, panickssery2024steeringllama2contrastive}. Our work follows the latter, providing a deeper account of ICL mechanisms than prior studies.

\paragraph{Interpreting arithmetic tasks.} 
A recent line of work examines how LLMs perform arithmetic \citep{stolfo-etal-2023-mechanistic, hanna2023doesgpt2computegreaterthan, nikankin2024arithmeticalgorithmslanguagemodels,Maltoni_2024}, and in particular addition \citep{nanda2023progress,zhong2023clockpizzastoriesmechanistic, zhou2024pre,kantamneni2025language}. \citet{zhou2024pre} and \citet{kantamneni2025language} identify Fourier and helical representations for direct addition. \citet{zhang2024interpreting} find direct arithmetic uses fewer than $5\%$ of attention heads focused on operands and operators. \citet{engels2025not} identify circular features for calendar-based modular arithmetic, while \citet{levy2025language} find digit-wise circular representations of numbers. These works study how models represent numbers or execute a given operation; we instead study how a task rule is inferred from demonstrations and represented within attention heads, revealing related periodic structure in the task representation. Concurrent work identifies partial-sum representations and linear function vectors in three-operand arithmetic ICL \citep{polyakov2025interpretability}.

\paragraph{Interpreting in-context learning.} Researchers have constructed detailed models of ICL for small transformers in supervised learning~\citep{garg2022can, akyurek2023learningalgorithmincontextlearning, zhang2023trainedtransformerslearnlinear, pmlr-v202-li23l, wu2024pretrainingtasksneededincontext}, as well as more complex settings~\citep{vonoswald2023transformerslearnincontextgradient, bai2023transformersstatisticiansprovableincontext, bietti2023birthtransformermemoryviewpoint, reddy2023mechanisticbasisdatadependence, guo2023transformerslearnincontextsimple, nichani2024transformerslearncausalstructure}.
For large pretrained models, ICL performance has been attributed to either induction heads~\citep{ih,singh2024needsrightinductionhead, crosbie2025inductionheadsessentialmechanism, bansal-etal-2023-rethinking} or function vector (FV) heads~\citep{fv, tv}.
Broader work traces how ICL information is encoded, aggregated,
retrieved, or filtered across the prompt
\citep{cho2025revisiting,bakalova2025contextualize,
cho2026mechanism}. Related work localizes task information with SAE
features \citep{kharlapenko2025scaling}, while concurrent and more
recent studies analyze attention heads through query-state geometry
and task-recognition versus task-learning roles, respectively
\citep{yang2025unifying,yang2026localizing}. Whereas these works
focus on prompt-level information flow, feature circuits, or query
transformations, \citet{yin2025attentionheadsmatterincontext} find
that few-shot ICL performance depends primarily on FV heads.
Motivated by this, we study FV heads in detail, jointly identifying
compact task-carrying head sets across task families---rather than
ranking heads independently as in \citet{fv}---and characterizing
task-conditioned low-dimensional subspaces within individual head
outputs. Concurrent work studies function induction in off-by-one
addition \citep{ye2026function}, while subsequent work approximates
FVs as combinations of example-level sub-FVs
\citep{wang2026fewshot}.

\paragraph{Causal analysis.} Our methodological approach follows recent work on causal effects of model components via interventions on internal states~\citep{vig2020investigating, NEURIPS2021_4f5c422f}, drawing inspiration from causal mediation analysis~\citep{fv}, activation patching~\citep{meng2022locating}, and causal scrubbing~\citep{chan2022causal}.  

\vspace{-0.3em}
\section{Discussion}
\vspace{-0.3em}
We provided a detailed mechanistic analysis of in-context learning for addition tasks. We found that a small number of attention heads operating in low-dimensional subspaces can extract, represent, and aggregate ICL task information in structured and interpretable ways. Our analysis proceeded in five steps: (1) sparse optimization to identify important heads (\S\ref{sec:33head}); (2) ablation to localize to a smaller subset (\S\ref{sec:3head}); (3) PCA to find low-dimensional subspaces (\S\ref{sec:6dim}); (4) rotation and decomposition revealing periodic unit-digit and magnitude subspaces (\S\ref{sec:periodic}, \S\ref{sec:4+2dim}); and (5) algebraic tracing connecting aggregator subspaces at the final token to extractor subspaces at earlier $y_i$ tokens (\S\ref{sec:extract}). Most steps scale easily; future work could automate the qualitative subspace examination using AI.

Beyond automation, our analysis raises two further directions. First, \emph{within} a task family, how do the identified heads and their subspaces function individually and together? Second, \emph{across} task families, are the important heads and subspaces shared or transferable? Appendix~\ref{app:generality-tasks} offers initial evidence of head-level sharing---one head, $(15,1)$, is a main head in every family we study---but subspace-level sharing and transfer remain open.

\section*{Acknowledgments}
We thank Lisa Dunlap, Alex Pan, Yaowen Ye, Ryan Wang, Prasann Singhal, Hanlin Zhu, Tony Lian, Jiaxin Ge, Shu Liu, Anya Ji, Junyi Zhang, and Baifeng Shi for their helpful feedback and suggestions on this work and earlier drafts of the manuscript. XH was supported in part by the Eric and Wendy Schmidt Fund for Strategic Innovation. KY is supported by the Vitalik Buterin Ph.D. Fellowship in AI Existential Safety. MJ is supported by the European Union (ERC-2022-SYG-OCEAN-101071601). LC is supported by a Miller Research Fellowship. JS is supported by Open Philanthropy, Simons Foundation, and NSF.

\bibliography{custom}
\bibliographystyle{colm2026_conference}

\newpage
\appendix

\section{Details of the Head-Set Refinement Criteria}\label{app:refine}

This appendix gives the full details of the two head-set refinement criteria stated in \S\ref{sec:3head} --- significant heads and main heads --- both built on the recovery scan of equation~\ref{eq:recovery-scan}. Table~\ref{tab:nested-headsets} summarizes the nested head sets they produce for Llama-3 \textit{add-k}, together with the AIE baseline of \citet{fv}. We reuse both criteria for every task family, model, and prompt format we study (Appendix~\ref{app:generality}).

\paragraph{Significant heads: statistically supporting recovery.}
For each selected head and each positive coefficient $c$ in the recovery scan, we compare per-example correctness against $c=0$ on the same prompts. Let $b$ be the number of examples fixed by the intervention and $d$ the number broken. We use a one-sided exact McNemar test---equivalently, an exact binomial test on the $b+d$ discordant examples~\citep{mcnemar1947note}. We apply a Bonferroni correction over the positive coefficients within each head, then control the false discovery rate across heads using the Benjamini--Hochberg procedure at $q=0.05$~\citep{benjamini1995controlling}. We call a head \emph{significant} if at least one positive coefficient has $b>d$ and survives both statistical corrections.

The paired comparison accounts for variation in prompt difficulty, while the two corrections protect against selecting a head merely because we scan many coefficients and many candidates. On Llama-3 \textit{add-k}, this procedure narrows $33$ selected heads to $13$ significant heads, while intervention accuracy decreases only from $0.798$ to $0.774$. The compression is more dramatic for Qwen-2.5-7B \textit{add-k}: $230$ selected heads narrow to $18$, with accuracy $0.358$ versus $0.368$ for the full selected set (full results in Appendix~\ref{app:generality-models}).

\paragraph{Main heads: materially dominant recovery.}
The paired tests above are designed to detect statistically reliable effects using per-example outcomes and may therefore retain heads with small overall gains. For a compact mechanistic core, we further compare each significant head's aggregated recovery gain
\[
    G_h
    =
    \max_{c\in\{0,\ldots,20\}}A_h(c)-A_h(0),
\]
where $A_h(c)$ denotes the resulting intervention accuracy of $v_k(h,c)$ from equation~\ref{eq:recovery-scan} on zero-shot prompts.
Formally, let $G^\star$ be the largest gain $G_h$ among the significant heads. For any task family with clean accuracy at least $0.25$, we call a significant head \emph{main} if $G_h\geq G^\star/4$. On \textit{add-k}, this rule recovers exactly the three heads of \S\ref{sec:3head}.

Beyond addition, we find one head, $(15,1)$, in Llama-3-8B is a main head for every task family we study. Addition, subtraction, and multiplication are localized to $3$, $4$, and $5$ main heads, respectively, with substantial reuse of the three main heads for addition. Phi-4 mirrors this structure: one head, $(18,12)$, appears in every family's main set, and its subtraction main heads are exactly its addition main heads plus one extra head --- the analogue of subtraction main heads $=$ addition trio $+\,(15,17)$ in Llama-3 (Appendices~\ref{app:generality-tasks}, \ref{app:generality-models}). The selection is also robust along the prompt-format and shot-count axes: the three \textit{add-k} main heads remain main heads under every prompt format and at every $n\in\{1,\dots,10\}$ (Appendices~\ref{app:generality-fmt}, \ref{app:generality-shots}). Moreover, in every task family, model, prompt format, and shot count we consider, the main heads achieve higher intervention accuracy than the top-$k$ head sets selected by \citet{fv}'s AIE ranking while using far fewer heads (Appendix~\ref{app:generality-aie}).

\paragraph{Comparison of the two refinements.}
The two criteria serve different purposes: significant-head selection retains all statistically supported contributors and aims to preserve as much of the selected set’s recovery as possible, whereas main-head selection further isolates, within the significant set, a small number of materially dominant heads for subsequent mechanistic analysis.

\begin{table}[t]
\centering
\small
\begin{tabular}{p{1.08in}p{2.05in}cc}
\toprule
Head set & Selection criterion & \# Heads & Accuracy [95\% CI] \\
\midrule
Selected
& Identified by joint sparse optimization
& 33 & 0.798 [0.760, 0.832] \\
Significant
& Statistically significant to fix more examples than it breaks
& 13 & 0.774 [0.736, 0.810] \\
Main
& Accuracy gain of at least $1/4$ of the best accuracy gain of significant heads
& 3 & 0.680 [0.640, 0.720] \\
\midrule
AIE top-$33$
& Ranked by average indirect effect
& 33 & 0.414 [0.372, 0.456] \\
\bottomrule
\end{tabular}
\caption{{Nested head sets for Llama-3-8B-Instruct \textit{add-k}, plus the AIE baseline of \citet{fv} at the matched budget. Selected-set accuracy uses all $33$ task-conditioned heads; the significant and main rows keep the listed set task-conditioned and mean-ablate the remaining selected heads. Brackets are $95\%$ bootstrap confidence intervals. For reference, the clean five-shot ICL accuracy is $0.870$ [$0.838$, $0.898$] and the corrupted accuracy (zero-shot, every selected head mean-ablated) is $0.037$ [$0.030$, $0.044$].}}
\label{tab:nested-headsets}
\end{table}

\section{Generality of the Localization Pipeline}\label{sec:generality}\label{app:generality}

This appendix applies the full pipeline of \S\ref{sec:id} --- head selection via sparse optimization, significant-head refinement, main-head selection (criteria detailed in Appendix~\ref{app:refine}), and the subspace projection of \S\ref{sec:6dim} --- to other task families (Table~\ref{tab:generality-tasks}), models (Table~\ref{tab:generality-models}), prompt formats (Table~\ref{tab:generality-formats}), and numbers of demonstrations (Table~\ref{tab:generality-shots}). In every variant that trains a coefficient matrix, the selected set is determined by the largest-gap rule of \S\ref{sec:33head}; the $n$-shot variants reuse the canonical five-shot selection.

\paragraph{Setup.} For every variant, the family's tasks are split into training tasks and $5$ held-out tasks (per-family task counts in Table~\ref{tab:task-families}); coefficient-matrix training and all head selection see only the training tasks, and every clean and intervention accuracy is measured on the held-out tasks ($100$ queries each, i.e., $500$ evaluations per number). Each entry reports the mean with a $95\%$ bootstrap confidence interval ($2{,}000$ resamples) beneath it. The rows are: \textbf{clean} ($n$-shot accuracy without intervention); \textbf{corrupted} (zero-shot accuracy with every selected head mean-ablated; this baseline comes from the recovery scan and is measured on the training-task scan prompts); the \textbf{intervention accuracies} of the selected, significant, and main head sets (zero-shot prompts; the function vector is the unit-coefficient sum of the row's head set, with the remaining selected heads mean-ablated), annotated with (heads used\,/\,total heads); \textbf{projected main heads} (each main head first projected onto its top principal components at $95\%$ explained variance, annotated with PC dimensions kept\,/\,total head-output dimensions $= 128\times$\,\#main heads); and the \textbf{AIE baseline} (function vector from the top-$k$ heads ranked by a paper-faithful re-implementation of \citet{fv}'s average indirect effect, at the small top-$k$ budget of their paper: $k=20$ for Llama-3-8B and Phi-4, $k=16$ for Qwen-2.5-7B --- $\approx2\%$ of heads for Llama-3 and Qwen). Non-arrow prompt formats use per-format retrained coefficient matrices; the $n$-shot variants reuse the canonical five-shot matrix and only re-run the recovery scan ($20$ queries per training task for $n\neq5$). All accuracies use the exact-token-match scorer. In the tables, intervention accuracies and head budgets of our head sets are typeset in \textcolor{red}{red} and the AIE baseline in \textcolor{blue}{blue}.

\subsection{Task families}\label{app:generality-tasks}

Table~\ref{tab:task-families} specifies the five task families. Each arithmetic task fixes a constant $k$ and applies a fixed rule to an input integer $x\in[1,100]$, with $100$ datapoints per task: the addition family is \textit{add-$k$} with $k\in[1,30]$ (\S\ref{sec:model-task}), the subtraction family is \textit{subtract-$k$} ($y=x-k$) with $k\in[0,30]$ (including the identity task $k=0$), and the multiplication family is \textit{multiply-$k$} ($y=kx$) with $k\in[1,30]$ (including the identity task $k=1$). The abstractive and extractive families use the task datasets of \citet{fv}. An abstractive task maps its input to a semantically associated output that need not appear in the input: antonym and synonym; English--French and English--Spanish translation; country-capital, country-currency, landmark-country, park-country, and national-parks; person-occupation, person-sport, and person-instrument; product-company; present-past; singular-plural; sentiment; next-item and previous-item; next-capital-letter; word-length; and six capitalization/lowercasing manipulations ($26$ tasks in total). An extractive task outputs a span of its input. Most are word-list tasks: given a list of words, choose the first, middle, or last item; the alphabetically first or last item; or the item of the first-named category (adjective-vs-verb, verb-vs-adjective, animal-vs-object, color-vs-animal, concept-vs-object, object-vs-concept, and fruit-vs-animal --- e.g., the animal in animal-vs-object); each of these $12$ rules is a separate task for lists of $3$ and of $5$ words. The remaining three tasks are CoNLL-2003 named-entity extraction: given a news sentence or fragment, output the (possibly multi-word) person, location, or organization it mentions ($27$ tasks in total).

\begin{table}[h]
\centering
\small
\setlength{\tabcolsep}{4pt}
\begin{tabular}{llcl}
\toprule
Task family & Task family feature & \# Tasks & Example task: input$\rightarrow$label \\
\midrule
Addition & $y=x+k$,\ $k\in[1,30]$ & $30$ & add-$7$: $3\rightarrow10$ \\
Subtraction & $y=x-k$,\ $k\in[0,30]$ & $31$ & subtract-$7$: $3\rightarrow-4$ \\
Multiplication & $y=kx$,\ $k\in[1,30]$ & $30$ & multiply-$4$: $3\rightarrow12$ \\
Abstractive~\citep{fv} & semantic association & $26$ & product-company: iPhone 5$\rightarrow$Apple \\
Extractive~\citep{fv} & copy an input span & $27$ & \begin{tabular}[t]{@{}l@{}}animal-vs-object:\\bear, ukulele, blanket$\rightarrow$bear\end{tabular} \\
\bottomrule
\end{tabular}
\caption{The five task families. Arithmetic inputs range over $x\in[1,100]$ ($100$ datapoints per task); the abstractive and extractive task datasets come from \citet{fv} and vary in size. Within each family, the split into training tasks and $5$ held-out tasks is fixed once and reused across all models, prompt formats, and shot counts; every clean and intervention accuracy is measured on the held-out tasks. Example datapoints are verbatim dataset entries.}
\label{tab:task-families}
\end{table}

Beyond addition, we run the pipeline on subtraction, multiplication, and the extractive and abstractive families of \citet{fv}, all on Llama-3-8B-Instruct (Table~\ref{tab:generality-tasks}). The localization structure is shared but family-dependent: one head, $(15,1)$, is a main head in every family; the subtraction main heads are exactly the addition trio plus $(15,17)$; multiplication is led by $(13,6)$ (with $(15,1)$, $(15,17)$, $(15,28)$, and $(13,21)$); and the semantic families are genuinely more distributed --- $11$ (extractive) and $10$ (abstractive) main heads with a broader layer spread --- while still led by $(15,1)$.

\begin{table}[h]
\centering
\small
\setlength{\tabcolsep}{5pt}
\begin{tabular}{lcccc}
\toprule
 & Subtraction & Multiplication & Extractive & Abstractive \\
\midrule
Clean accuracy & 0.526 & 0.348 & 0.810 & 0.422 \\
 & \scriptsize[0.484, 0.572] & \scriptsize[0.308, 0.390] & \scriptsize[0.776, 0.846] & \scriptsize[0.378, 0.466] \\
\addlinespace
Corrupted accuracy & 0.020 & 0.030 & 0.309 & 0.023 \\
 & \scriptsize[0.015, 0.025] & \scriptsize[0.024, 0.037] & \scriptsize[0.289, 0.328] & \scriptsize[0.017, 0.030] \\
\addlinespace
Selected heads &  \textcolor{red}{0.330} (\textcolor{red}{28}/1024) &  \textcolor{red}{0.226} (\textcolor{red}{37}/1024) &  \textcolor{red}{0.506} (\textcolor{red}{15}/1024) &  \textcolor{red}{0.300} (\textcolor{red}{23}/1024) \\
 & \scriptsize[0.290, 0.372] & \scriptsize[0.190, 0.264] & \scriptsize[0.460, 0.548] & \scriptsize[0.260, 0.342] \\
\addlinespace
Significant heads &  \textcolor{red}{0.326} (\textcolor{red}{19}/1024) &  \textcolor{red}{0.134} (\textcolor{red}{13}/1024) &  \textcolor{red}{0.596} (\textcolor{red}{13}/1024) &  \textcolor{red}{0.290} (\textcolor{red}{19}/1024) \\
 & \scriptsize[0.286, 0.368] & \scriptsize[0.104, 0.162] & \scriptsize[0.554, 0.640] & \scriptsize[0.250, 0.330] \\
\addlinespace
Main heads &  \textcolor{red}{0.290} (\textcolor{red}{4}/1024) &  \textcolor{red}{0.080} (\textcolor{red}{5}/1024) &  \textcolor{red}{0.564} (\textcolor{red}{11}/1024) &  \textcolor{red}{0.248} (\textcolor{red}{10}/1024) \\
 & \scriptsize[0.252, 0.330] & \scriptsize[0.058, 0.104] & \scriptsize[0.522, 0.608] & \scriptsize[0.210, 0.286] \\
\addlinespace
Projected main heads &  \textcolor{red}{0.282} (\textcolor{red}{28}/512) &  \textcolor{red}{0.076} (\textcolor{red}{40}/640) &  \textcolor{red}{0.538} (\textcolor{red}{109}/1408) &  \textcolor{red}{0.218} (\textcolor{red}{130}/1280) \\
 & \scriptsize[0.244, 0.320] & \scriptsize[0.054, 0.100] & \scriptsize[0.494, 0.582] & \scriptsize[0.182, 0.256] \\
\addlinespace
AIE top-$k$ heads &  \textcolor{blue}{0.002} (\textcolor{blue}{$k{=}20$}) &  \textcolor{blue}{0.000} (\textcolor{blue}{$k{=}20$}) &  \textcolor{blue}{0.448} (\textcolor{blue}{$k{=}20$}) &  \textcolor{blue}{0.168} (\textcolor{blue}{$k{=}20$}) \\
 & \scriptsize[0.000, 0.006] & \scriptsize[0.000, 0.000] & \scriptsize[0.406, 0.494] & \scriptsize[0.136, 0.200] \\
\bottomrule
\end{tabular}
\caption{Generalization across task families (Llama-3-8B-Instruct, arrow format, five-shot). Rows as defined in the setup paragraph; brackets are $95\%$ bootstrap confidence intervals. Parentheses report the budget of each intervention: (number of heads used\,/\,total heads in the model) for the selected, significant, and main rows; (PC dimensions kept\,/\,total dimensions of the main heads' outputs, $128$ per head) for the projected row; and the top-$k$ head budget for the AIE row. The extractive corrupted floor reflects the multiple-choice nature of that family (chance $\approx0.31$).}
\label{tab:generality-tasks}
\end{table}

\subsection{Models}\label{app:generality-models}

Table~\ref{tab:generality-models} applies the pipeline to \textit{add-k} on Phi-4-14B and Qwen-2.5-7B. Phi-4 localizes to three main heads $(18,12)$, $(18,14)$, $(17,23)$ in layers $17$--$18$ (of $40$ layers and $1600$ heads) and mirrors the Llama-3 family structure: $(18,12)$ is a main head for every Phi-4 task family, and its subtraction main heads are exactly its addition main heads plus $(17,22)$. Table~\ref{tab:generality-tasks-phi4} reports the Phi-4 task-family results, the analogue of Table~\ref{tab:generality-tasks}; the per-family main-head sets are subtraction $\{(18,12),(18,14),(17,22),(17,23)\}$, multiplication $\{(18,12),(11,9),(18,14),(24,2)\}$, extractive $\{(24,31),(19,4),(18,12),(21,25),(17,22),(17,23)\}$, and abstractive ($14$ heads, again including $(18,12)$). Note that the Phi-4 subtraction interventions from our head sets sit \emph{above} the clean accuracy. Qwen-2.5-7B's main heads are exactly the layer-$21$ quartet $(21,0)$, $(21,5)$, $(21,4)$, $(21,2)$. Qwen's unit-coefficient composites carry a metric caveat: its tokenizer splits numbers into digits, which the exact-token-match scorer punishes; its trained-coefficient function vector reaches $0.460$.

\begin{table}[h]
\centering
\small
\setlength{\tabcolsep}{6pt}
\begin{tabular}{lccc}
\toprule
 & Llama-3-8B-Instruct & Phi-4-14B & Qwen-2.5-7B \\
\midrule
Clean accuracy & 0.870 & 0.928 & 0.896 \\
 & \scriptsize[0.838, 0.898] & \scriptsize[0.904, 0.950] & \scriptsize[0.870, 0.922] \\
\addlinespace
Corrupted accuracy & 0.037 & 0.031 & 0.032 \\
 & \scriptsize[0.030, 0.044] & \scriptsize[0.024, 0.038] & \scriptsize[0.026, 0.039] \\
\addlinespace
Selected heads &  \textcolor{red}{0.798} (\textcolor{red}{33}/1024) &  \textcolor{red}{0.908} (\textcolor{red}{23}/1600) &  \textcolor{red}{0.368} (\textcolor{red}{230}/784) \\
 & \scriptsize[0.760, 0.832] & \scriptsize[0.882, 0.932] & \scriptsize[0.326, 0.410] \\
\addlinespace
Significant heads &  \textcolor{red}{0.774} (\textcolor{red}{13}/1024) &  \textcolor{red}{0.910} (\textcolor{red}{13}/1600) &  \textcolor{red}{0.358} (\textcolor{red}{18}/784) \\
 & \scriptsize[0.736, 0.810] & \scriptsize[0.884, 0.934] & \scriptsize[0.314, 0.400] \\
\addlinespace
Main heads &  \textcolor{red}{0.680} (\textcolor{red}{3}/1024) &  \textcolor{red}{0.876} (\textcolor{red}{3}/1600) &  \textcolor{red}{0.270} (\textcolor{red}{4}/784) \\
 & \scriptsize[0.640, 0.720] & \scriptsize[0.848, 0.904] & \scriptsize[0.232, 0.310] \\
\addlinespace
Projected main heads &  \textcolor{red}{0.636} (\textcolor{red}{18}/384) &  \textcolor{red}{0.870} (\textcolor{red}{22}/384) &  \textcolor{red}{0.252} (\textcolor{red}{28}/512) \\
 & \scriptsize[0.592, 0.678] & \scriptsize[0.842, 0.898] & \scriptsize[0.216, 0.290] \\
\addlinespace
AIE top-$k$ heads &  \textcolor{blue}{0.124} (\textcolor{blue}{$k{=}20$}) &  \textcolor{blue}{0.728} (\textcolor{blue}{$k{=}20$}) &  \textcolor{blue}{0.136} (\textcolor{blue}{$k{=}16$}) \\
 & \scriptsize[0.096, 0.152] & \scriptsize[0.690, 0.766] & \scriptsize[0.104, 0.168] \\
\bottomrule
\end{tabular}
\caption{Generalization across models (\textit{add-k}, arrow format, five-shot). Rows as defined in the setup paragraph; brackets are $95\%$ bootstrap confidence intervals. Parentheses report the budget of each intervention: (number of heads used\,/\,total heads in the model) for the selected, significant, and main rows; (PC dimensions kept\,/\,total dimensions of the main heads' outputs, $128$ per head) for the projected row; and the top-$k$ head budget for the AIE row.}
\label{tab:generality-models}
\end{table}

\begin{table}[h]
\centering
\small
\setlength{\tabcolsep}{5pt}
\begin{tabular}{lcccc}
\toprule
 & Subtraction & Multiplication & Extractive & Abstractive \\
\midrule
Clean accuracy & 0.658 & 0.426 & 0.768 & 0.442 \\
 & \scriptsize[0.618, 0.700] & \scriptsize[0.382, 0.468] & \scriptsize[0.732, 0.808] & \scriptsize[0.398, 0.486] \\
\addlinespace
Corrupted accuracy & 0.036 & 0.030 & 0.261 & 0.033 \\
 & \scriptsize[0.029, 0.043] & \scriptsize[0.024, 0.036] & \scriptsize[0.243, 0.279] & \scriptsize[0.026, 0.040] \\
\addlinespace
Selected heads & \textcolor{red}{0.754} (\textcolor{red}{37}/1600) & \textcolor{red}{0.438} (\textcolor{red}{36}/1600) & \textcolor{red}{0.412} (\textcolor{red}{18}/1600) & \textcolor{red}{0.326} (\textcolor{red}{31}/1600) \\
 & \scriptsize[0.718, 0.792] & \scriptsize[0.392, 0.482] & \scriptsize[0.370, 0.456] & \scriptsize[0.286, 0.366] \\
\addlinespace
Significant heads & \textcolor{red}{0.776} (\textcolor{red}{13}/1600) & \textcolor{red}{0.438} (\textcolor{red}{24}/1600) & \textcolor{red}{0.416} (\textcolor{red}{16}/1600) & \textcolor{red}{0.324} (\textcolor{red}{29}/1600) \\
 & \scriptsize[0.740, 0.814] & \scriptsize[0.392, 0.482] & \scriptsize[0.372, 0.460] & \scriptsize[0.284, 0.364] \\
\addlinespace
Main heads & \textcolor{red}{0.754} (\textcolor{red}{4}/1600) & \textcolor{red}{0.128} (\textcolor{red}{4}/1600) & \textcolor{red}{0.402} (\textcolor{red}{6}/1600) & \textcolor{red}{0.234} (\textcolor{red}{14}/1600) \\
 & \scriptsize[0.718, 0.792] & \scriptsize[0.102, 0.158] & \scriptsize[0.360, 0.444] & \scriptsize[0.198, 0.272] \\
\addlinespace
Projected main heads & \textcolor{red}{0.754} (\textcolor{red}{30}/512) & \textcolor{red}{0.120} (\textcolor{red}{38}/512) & \textcolor{red}{0.390} (\textcolor{red}{63}/768) & \textcolor{red}{0.182} (\textcolor{red}{187}/1792) \\
 & \scriptsize[0.718, 0.792] & \scriptsize[0.094, 0.148] & \scriptsize[0.348, 0.432] & \scriptsize[0.148, 0.216] \\
\addlinespace
AIE top-$k$ heads & \textcolor{blue}{0.038} (\textcolor{blue}{$k{=}20$}) & \textcolor{blue}{0.004} (\textcolor{blue}{$k{=}20$}) & \textcolor{blue}{0.344} (\textcolor{blue}{$k{=}20$}) & \textcolor{blue}{0.180} (\textcolor{blue}{$k{=}20$}) \\
 & \scriptsize[0.022, 0.056] & \scriptsize[0.000, 0.010] & \scriptsize[0.302, 0.388] & \scriptsize[0.146, 0.214] \\
\bottomrule
\end{tabular}
\caption{Generalization across task families for Phi-4-14B (arrow format, five-shot). Rows as defined in the setup paragraph; brackets are $95\%$ bootstrap confidence intervals. Parentheses report the budget of each intervention: (number of heads used\,/\,total heads in the model) for the selected, significant, and main rows; (PC dimensions kept\,/\,total dimensions of the main heads' outputs, $128$ per head) for the projected row; and the top-$k$ head budget for the AIE row.}
\label{tab:generality-tasks-phi4}
\end{table}

\subsection{Prompt formats}\label{app:generality-fmt}

Table~\ref{tab:generality-formats} varies the prompt format on Llama-3-8B \textit{add-k}, with a separately retrained coefficient matrix per format. Clean accuracy varies substantially across formats and the selected sets differ ($24$--$46$ heads), but the localization is structurally stable: \textbf{every format recovers the three main heads $(15,2)$, $(15,1)$, $(13,6)$} (formats qa and ab additionally recover $(15,17)$, giving four main heads there).

\begin{table}[h]
\centering
\setlength{\tabcolsep}{3pt}
\resizebox{\textwidth}{!}{%
\begin{tabular}{lcccccc}
\toprule
 & arrow (orig.) & qa & io & ab & textlabel & fx \\
\midrule
Clean accuracy & 0.870 & 0.418 & 0.532 & 0.700 & 0.832 & 0.348 \\
 & \scriptsize[0.838, 0.898] & \scriptsize[0.372, 0.462] & \scriptsize[0.490, 0.574] & \scriptsize[0.660, 0.738] & \scriptsize[0.798, 0.864] & \scriptsize[0.306, 0.390] \\
\addlinespace
Corrupted accuracy & 0.037 & 0.018 & 0.034 & 0.030 & 0.018 & 0.032 \\
 & \scriptsize[0.030, 0.044] & \scriptsize[0.013, 0.023] & \scriptsize[0.026, 0.041] & \scriptsize[0.024, 0.037] & \scriptsize[0.013, 0.024] & \scriptsize[0.025, 0.039] \\
\addlinespace
Selected heads &  \textcolor{red}{0.798} (\textcolor{red}{33}/1024) &  \textcolor{red}{0.498} (\textcolor{red}{28}/1024) &  \textcolor{red}{0.470} (\textcolor{red}{46}/1024) &  \textcolor{red}{0.736} (\textcolor{red}{38}/1024) &  \textcolor{red}{0.582} (\textcolor{red}{24}/1024) &  \textcolor{red}{0.270} (\textcolor{red}{32}/1024) \\
 & \scriptsize[0.760, 0.832] & \scriptsize[0.454, 0.542] & \scriptsize[0.426, 0.512] & \scriptsize[0.696, 0.772] & \scriptsize[0.540, 0.624] & \scriptsize[0.230, 0.310] \\
\addlinespace
Significant heads &  \textcolor{red}{0.774} (\textcolor{red}{13}/1024) &  \textcolor{red}{0.492} (\textcolor{red}{26}/1024) &  \textcolor{red}{0.440} (\textcolor{red}{9}/1024) &  \textcolor{red}{0.726} (\textcolor{red}{15}/1024) &  \textcolor{red}{0.558} (\textcolor{red}{21}/1024) &  \textcolor{red}{0.270} (\textcolor{red}{11}/1024) \\
 & \scriptsize[0.736, 0.810] & \scriptsize[0.448, 0.536] & \scriptsize[0.396, 0.482] & \scriptsize[0.686, 0.762] & \scriptsize[0.516, 0.598] & \scriptsize[0.230, 0.308] \\
\addlinespace
Main heads &  \textcolor{red}{0.680} (\textcolor{red}{3}/1024) &  \textcolor{red}{0.432} (\textcolor{red}{4}/1024) &  \textcolor{red}{0.334} (\textcolor{red}{3}/1024) &  \textcolor{red}{0.650} (\textcolor{red}{4}/1024) &  \textcolor{red}{0.496} (\textcolor{red}{3}/1024) &  \textcolor{red}{0.190} (\textcolor{red}{3}/1024) \\
 & \scriptsize[0.640, 0.720] & \scriptsize[0.388, 0.476] & \scriptsize[0.294, 0.374] & \scriptsize[0.608, 0.690] & \scriptsize[0.452, 0.538] & \scriptsize[0.156, 0.224] \\
\addlinespace
Projected main heads &  \textcolor{red}{0.636} (\textcolor{red}{18}/384) &  \textcolor{red}{0.410} (\textcolor{red}{24}/512) &  \textcolor{red}{0.306} (\textcolor{red}{18}/384) &  \textcolor{red}{0.614} (\textcolor{red}{24}/512) &  \textcolor{red}{0.464} (\textcolor{red}{18}/384) &  \textcolor{red}{0.184} (\textcolor{red}{18}/384) \\
 & \scriptsize[0.592, 0.678] & \scriptsize[0.366, 0.456] & \scriptsize[0.266, 0.348] & \scriptsize[0.570, 0.656] & \scriptsize[0.418, 0.506] & \scriptsize[0.150, 0.218] \\
\addlinespace
AIE top-$k$ heads (\textcolor{blue}{$k{=}20$}) &  \textcolor{blue}{0.124} &  \textcolor{blue}{0.126} &  \textcolor{blue}{0.078} &  \textcolor{blue}{0.196} &  \textcolor{blue}{0.104} &  \textcolor{blue}{0.064} \\
 & \scriptsize[0.096, 0.152] & \scriptsize[0.096, 0.156] & \scriptsize[0.056, 0.102] & \scriptsize[0.160, 0.230] & \scriptsize[0.078, 0.130] & \scriptsize[0.042, 0.086] \\
\bottomrule
\end{tabular}}
\caption{Generalization across prompt formats (Llama-3-8B-Instruct \textit{add-k}, five-shot; per-format retrained coefficient matrices). Rows as defined in the setup paragraph; brackets are $95\%$ bootstrap confidence intervals. Parentheses report the budget of each intervention: (number of heads used\,/\,total heads in the model) for the selected, significant, and main rows; (PC dimensions kept\,/\,total dimensions of the main heads' outputs, $128$ per head) for the projected row; and the top-$k$ head budget for the AIE row.}
\label{tab:generality-formats}
\end{table}

\subsection{Number of demonstrations}\label{app:generality-shots}

Table~\ref{tab:generality-shots} varies the number of demonstrations $n$ from $1$ to $10$ while reusing the canonical five-shot coefficient matrix. Clean accuracy rises from $0.256$ to $0.944$ (saturating around $n\approx7$), while the intervention accuracies plateau from $n\approx4$. The localization is again structurally stable: \textbf{every $n$ recovers the same three main heads}. The AIE baseline, by contrast, depends strongly on the demonstration budget (it ranks heads by corrupting demonstration labels): at $n{=}1$ its top-$33$ list omits $(15,2)$ and $(13,6)$ entirely, and its top-$20$ accuracy climbs from $0.014$ to $\sim0.3$ only at large $n$.

\begin{table}[h]
\centering
\setlength{\tabcolsep}{2.5pt}
\resizebox{\textwidth}{!}{%
\begin{tabular}{lcccccccccc}
\toprule
$n$ & 1 & 2 & 3 & 4 & 5 (orig.) & 6 & 7 & 8 & 9 & 10 \\
\midrule
Clean accuracy & 0.256 & 0.502 & 0.700 & 0.816 & 0.870 & 0.898 & 0.932 & 0.944 & 0.940 & 0.944 \\
 & \scriptsize[0.216, 0.294] & \scriptsize[0.458, 0.544] & \scriptsize[0.658, 0.740] & \scriptsize[0.780, 0.850] & \scriptsize[0.838, 0.898] & \scriptsize[0.870, 0.924] & \scriptsize[0.908, 0.954] & \scriptsize[0.924, 0.962] & \scriptsize[0.918, 0.960] & \scriptsize[0.924, 0.962] \\
\addlinespace
Corrupted accuracy & 0.046 & 0.034 & 0.028 & 0.032 & 0.037 & 0.032 & 0.034 & 0.032 & 0.032 & 0.032 \\
 & \scriptsize[0.030, 0.066] & \scriptsize[0.020, 0.050] & \scriptsize[0.014, 0.044] & \scriptsize[0.018, 0.048] & \scriptsize[0.030, 0.044] & \scriptsize[0.018, 0.048] & \scriptsize[0.020, 0.050] & \scriptsize[0.018, 0.048] & \scriptsize[0.018, 0.048] & \scriptsize[0.018, 0.048] \\
\addlinespace
Selected heads (\textcolor{red}{33}/1024) &  \textcolor{red}{0.188} &  \textcolor{red}{0.586} &  \textcolor{red}{0.752} &  \textcolor{red}{0.782} &  \textcolor{red}{0.798} &  \textcolor{red}{0.774} &  \textcolor{red}{0.772} &  \textcolor{red}{0.770} &  \textcolor{red}{0.772} &  \textcolor{red}{0.754} \\
 & \scriptsize[0.154, 0.222] & \scriptsize[0.544, 0.630] & \scriptsize[0.714, 0.790] & \scriptsize[0.746, 0.818] & \scriptsize[0.760, 0.832] & \scriptsize[0.736, 0.808] & \scriptsize[0.736, 0.808] & \scriptsize[0.734, 0.806] & \scriptsize[0.736, 0.806] & \scriptsize[0.716, 0.790] \\
\addlinespace
Significant heads &  \textcolor{red}{0.184} (\textcolor{red}{6}) &  \textcolor{red}{0.534} (\textcolor{red}{8}) &  \textcolor{red}{0.700} (\textcolor{red}{10}) &  \textcolor{red}{0.776} (\textcolor{red}{10}) &  \textcolor{red}{0.774} (\textcolor{red}{13}) &  \textcolor{red}{0.756} (\textcolor{red}{11}) &  \textcolor{red}{0.746} (\textcolor{red}{10}) &  \textcolor{red}{0.750} (\textcolor{red}{10}) &  \textcolor{red}{0.748} (\textcolor{red}{10}) &  \textcolor{red}{0.740} (\textcolor{red}{10}) \\
 & \scriptsize[0.150, 0.218] & \scriptsize[0.490, 0.578] & \scriptsize[0.658, 0.740] & \scriptsize[0.738, 0.812] & \scriptsize[0.736, 0.810] & \scriptsize[0.716, 0.792] & \scriptsize[0.706, 0.782] & \scriptsize[0.712, 0.786] & \scriptsize[0.710, 0.784] & \scriptsize[0.702, 0.778] \\
\addlinespace
Main heads (\textcolor{red}{3}/1024) &  \textcolor{red}{0.156} &  \textcolor{red}{0.452} &  \textcolor{red}{0.626} &  \textcolor{red}{0.674} &  \textcolor{red}{0.680} &  \textcolor{red}{0.664} &  \textcolor{red}{0.652} &  \textcolor{red}{0.652} &  \textcolor{red}{0.644} &  \textcolor{red}{0.628} \\
 & \scriptsize[0.122, 0.186] & \scriptsize[0.408, 0.496] & \scriptsize[0.582, 0.668] & \scriptsize[0.632, 0.716] & \scriptsize[0.640, 0.720] & \scriptsize[0.622, 0.704] & \scriptsize[0.608, 0.694] & \scriptsize[0.608, 0.692] & \scriptsize[0.600, 0.684] & \scriptsize[0.584, 0.670] \\
\addlinespace
Projected main heads (\textcolor{red}{18}/384) &  \textcolor{red}{0.156} &  \textcolor{red}{0.410} &  \textcolor{red}{0.592} &  \textcolor{red}{0.638} &  \textcolor{red}{0.636} &  \textcolor{red}{0.632} &  \textcolor{red}{0.618} &  \textcolor{red}{0.616} &  \textcolor{red}{0.620} &  \textcolor{red}{0.620} \\
 & \scriptsize[0.122, 0.186] & \scriptsize[0.366, 0.452] & \scriptsize[0.548, 0.636] & \scriptsize[0.594, 0.680] & \scriptsize[0.592, 0.678] & \scriptsize[0.588, 0.674] & \scriptsize[0.572, 0.660] & \scriptsize[0.570, 0.658] & \scriptsize[0.574, 0.662] & \scriptsize[0.574, 0.662] \\
\addlinespace
AIE top-$k$ heads (\textcolor{blue}{$k{=}20$}) &  \textcolor{blue}{0.014} &  \textcolor{blue}{0.042} &  \textcolor{blue}{0.072} &  \textcolor{blue}{0.074} &  \textcolor{blue}{0.124} &  \textcolor{blue}{0.312} &  \textcolor{blue}{0.186} &  \textcolor{blue}{0.318} &  \textcolor{blue}{0.172} &  \textcolor{blue}{0.316} \\
 & \scriptsize[0.004, 0.026] & \scriptsize[0.026, 0.060] & \scriptsize[0.050, 0.096] & \scriptsize[0.052, 0.098] & \scriptsize[0.096, 0.152] & \scriptsize[0.270, 0.350] & \scriptsize[0.152, 0.220] & \scriptsize[0.276, 0.356] & \scriptsize[0.138, 0.204] & \scriptsize[0.274, 0.354] \\
\bottomrule
\end{tabular}}
\caption{Generalization across the number of demonstrations $n$ (Llama-3-8B-Instruct \textit{add-k}, arrow format; the five-shot coefficient matrix is reused, so the selected set is fixed at $33$ heads). Significant-head counts are per-$n$ (out of $1024$). Recovery scans for $n\neq5$ use $20$ queries per training task. Brackets are $95\%$ bootstrap confidence intervals. Parenthetical row-label annotations report the budget of each intervention: (number of heads used\,/\,total heads in the model) for the selected and main rows; (PC dimensions kept\,/\,total dimensions of the main heads' outputs, $128$ per head) for the projected row; and the top-$k$ head budget for the AIE row.}
\label{tab:generality-shots}
\end{table}

\subsection{Comparison with the AIE baseline}\label{app:generality-aie}

The AIE rows above evaluate \citet{fv}'s head selection at the small top-$k$ budget of their paper. The comparison is not driven by the budget: at matched $k$ equal to our selected-set size, the AIE top-$k$ set also trails our selected set in every variant --- on canonical \textit{add-k}, $0.414$ [$0.372$, $0.456$] at $k{=}33$ versus our $0.798$ [$0.760$, $0.832$]. In \textbf{every} task family, model, prompt format, and shot count above, our $3$--$14$ main heads achieve higher intervention accuracy than the AIE top-$k$ set while using fewer heads; the pattern also reproduces on the Phi-4 non-addition families (Table~\ref{tab:generality-tasks-phi4}). We note one nuance in fairness to AIE: its per-head \emph{ranking} is informative --- it places our main heads at or near the top of the per-head list in essentially every variant --- but its top-$k$ \emph{sets} underperform because the tail of the ranking does not sum into an effective function vector.

\subsection{Subspace dimensionality across variants}\label{app:generality-subspace}

The six-dimensional finding of \S\ref{sec:6dim} generalizes as low-dimensional structure with family-dependent dimensionality. The number of principal components needed to reach $95\%$ of task variance for main heads is $6$ per head on Llama-3 addition (every format and every $n$), $7$ on subtraction, $6$--$9$ on multiplication, $7$--$8$ on Phi-4 addition, and $5$--$8$ on Qwen-2.5 addition, rising to $6$--$11$ (extractive) and $11$--$15$ (abstractive) for the semantic families. Causal sufficiency is universal: as the projected rows of Tables~\ref{tab:generality-tasks}--\ref{tab:generality-shots} show, replacing every main head's vector by its projection onto these subspaces costs at most $0.052$ accuracy in any displayed cell ($0.058$ across the full grid including significant sets).

\clearpage

\section{Supplement for \S\ref{sec:6dim}}\label{app:6dim}
\begin{figure}[htbp]
    \centering

    \begin{subfigure}{0.8\linewidth}
        \includegraphics[width=\linewidth]{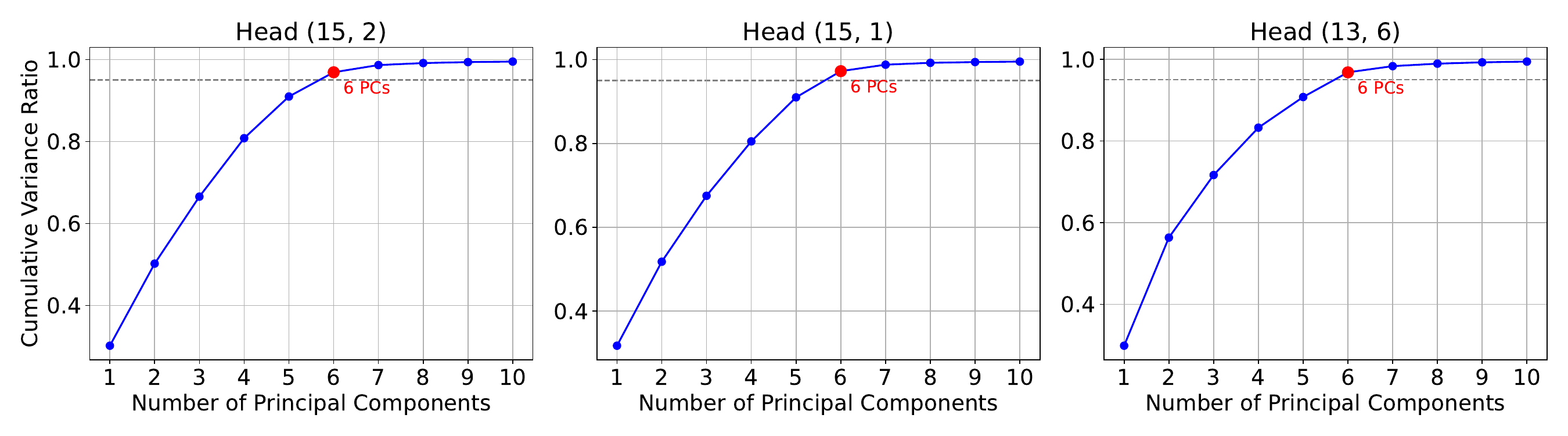}
        \caption{Llama-3-8B-Instruct}
        \label{fig:pca_variance}
    \end{subfigure}\hfill
    \begin{subfigure}{0.8\linewidth}
        \includegraphics[width=\linewidth]{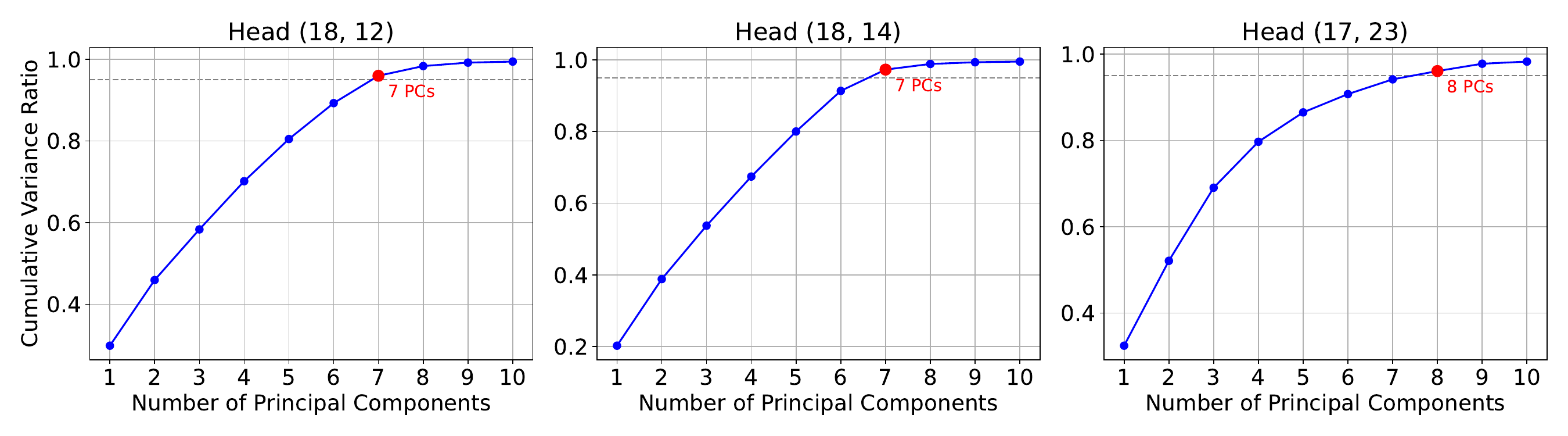}
        \caption{Phi-4-14B}
        \label{fig:pca_variance_phi4}
    \end{subfigure}\hfill
    \begin{subfigure}{0.8\linewidth}
        \includegraphics[width=\linewidth]{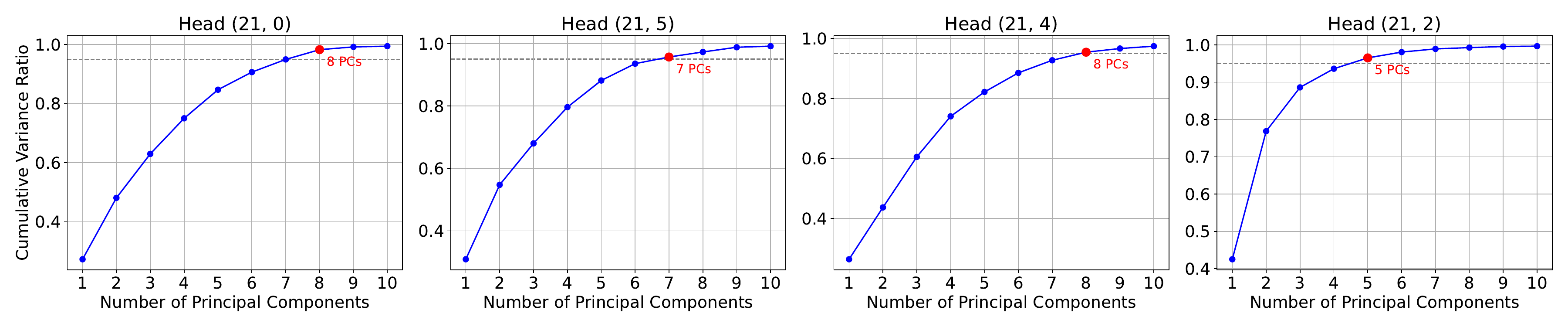}
        \caption{Qwen-2.5-7B}
        \label{fig:pca_variance_qwen}
    \end{subfigure}

    \caption{Cumulative explained variance ratio vs.\ number of PCs for each main head. The first six PCs capture $97\%$ of the task variance for the Llama-3-8B-Instruct main heads; the Phi-4 main heads reach $95\%$ at $7$--$8$ PCs and the Qwen-2.5-7B main heads at $5$--$8$ PCs.}
    \label{fig:pca_variance_all}
\end{figure}

We perform PCA on the $30$ head vectors of each main head and plot the cumulative explained variance ratio against the number of PCs (Figure~\ref{fig:pca_variance_all}). Six directions explain $97\%$ of the task variance for every Llama-3-8B-Instruct main head (Figure~\ref{fig:pca_variance}); the Phi-4 main heads reach $95\%$ at $7$--$8$ PCs and the Qwen-2.5-7B main heads at $5$--$8$ PCs (Figures~\ref{fig:pca_variance_phi4} and \ref{fig:pca_variance_qwen}).

\clearpage

\section{Supplement for \S\ref{sec:periodic}}
We first present additional figures for Llama-3-8B-Instruct, followed by results for Phi-4 and Qwen-2.5-7B. The periodic structure generalizes: all Phi-4 main heads fit the six trigonometric patterns well, while for Qwen-2.5 the periodic (unit-digit) patterns are fit well by heads $(21,0)$ and $(21,4)$ but poorly by $(21,2)$ and $(21,5)$, whose coordinates instead follow the two low-frequency (magnitude) directions and behave distinctly across the intervals $[1,10]$, $[11,20]$, and $[21,30]$. We conjecture this difference arises from tokenization: Qwen encodes numbers digit by digit, leading to discontinuities across each 10-interval, whereas Llama-3 and Phi-4 represent every number in our range as a single token.

\subsection{Additional figures for \S\ref{sec:periodic}}\label{app:periodic}
\begin{figure}[htbp]
    \centering
    \begin{subfigure}{0.8\linewidth}
        \includegraphics[width=\linewidth]{figures_new/pca_2x3_15_2_llama3.pdf}
        \caption{Head $(15,2)$}
        \label{fig:periodic}
    \end{subfigure}\hfill
    \begin{subfigure}{0.8\linewidth}
        \includegraphics[width=\linewidth]{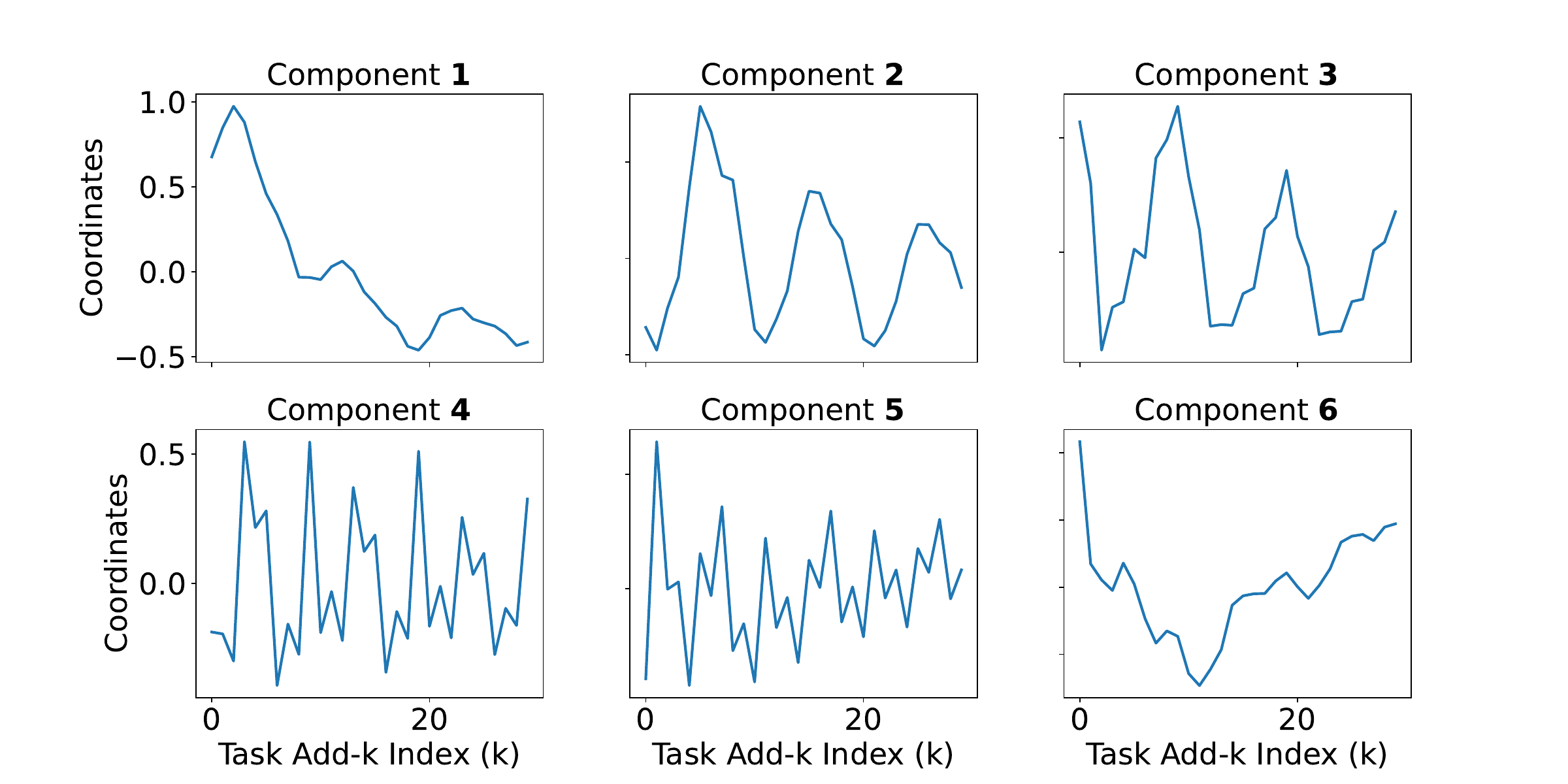}
        \caption{Head $(15,1)$}
        \label{fig:periodic_(15, 1)}
    \end{subfigure}\hfill
    \begin{subfigure}{0.8\linewidth}
        \includegraphics[width=\linewidth]{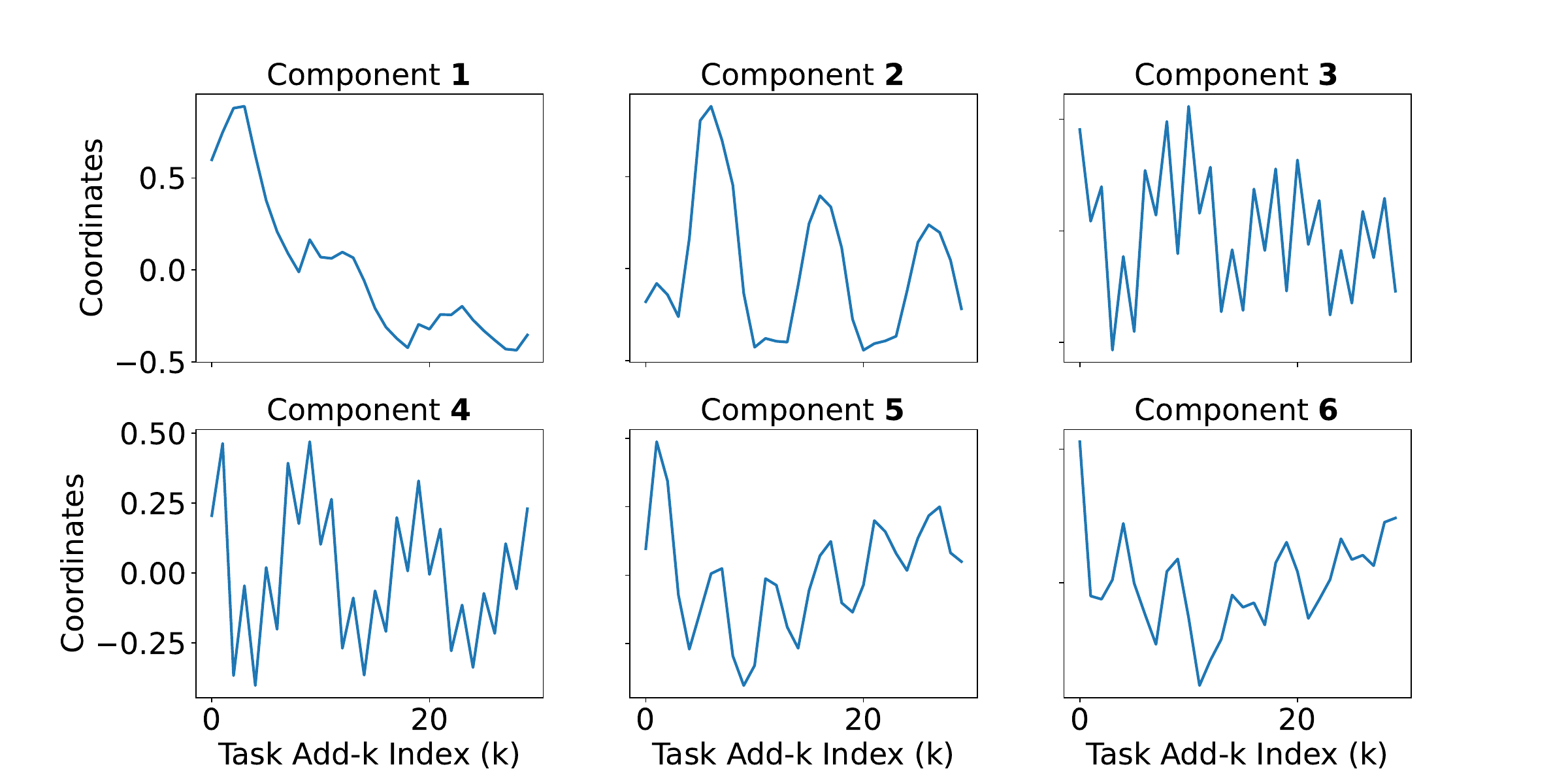}
        \caption{Head $(13,6)$}
        \label{fig:periodic_(13, 6)}
    \end{subfigure}
    \caption{Coordinates of three heads’ vectors (inner products with PCs) for the first six PCs across different add-$k$ tasks on Llama-3-8B-Instruct. Periodic patterns are visible in the first few PCs of each head.}
    \label{fig:periodic_group}
\end{figure}

\begin{figure}[htbp]
    \centering
    \begin{subfigure}{0.7\linewidth}
        \includegraphics[width=\linewidth]{figures_new/fitted_15_2_llama3.pdf}
        \caption{Head $(15,2)$}
        \label{fig:fitted}
    \end{subfigure}\hfill
    \begin{subfigure}{0.7\linewidth}
        \includegraphics[width=\linewidth]{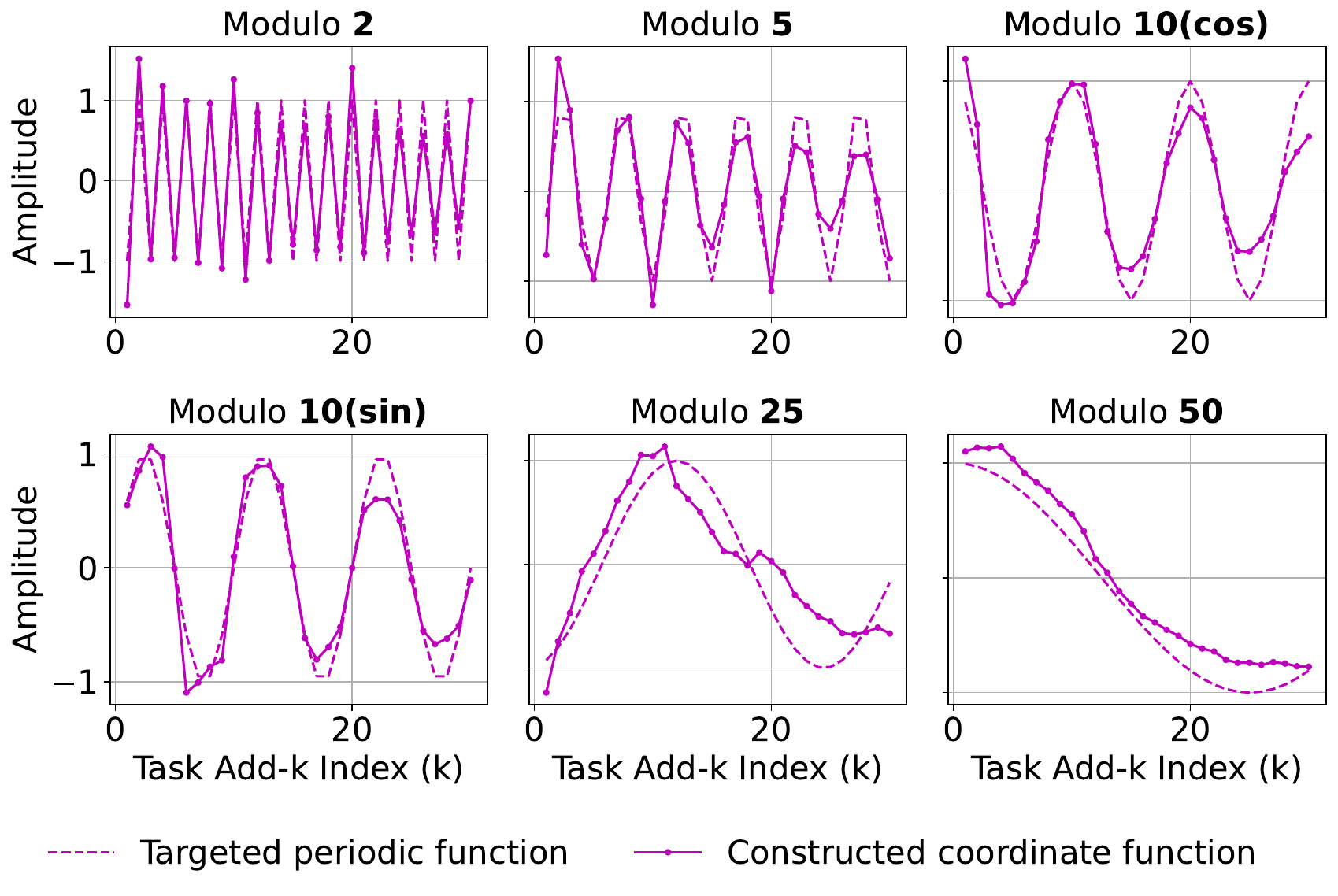}
        \caption{Head $(15,1)$}
        \label{fig:fitted_separated_head2}
    \end{subfigure}\hfill
    \begin{subfigure}{0.7\linewidth}
        \includegraphics[width=\linewidth]{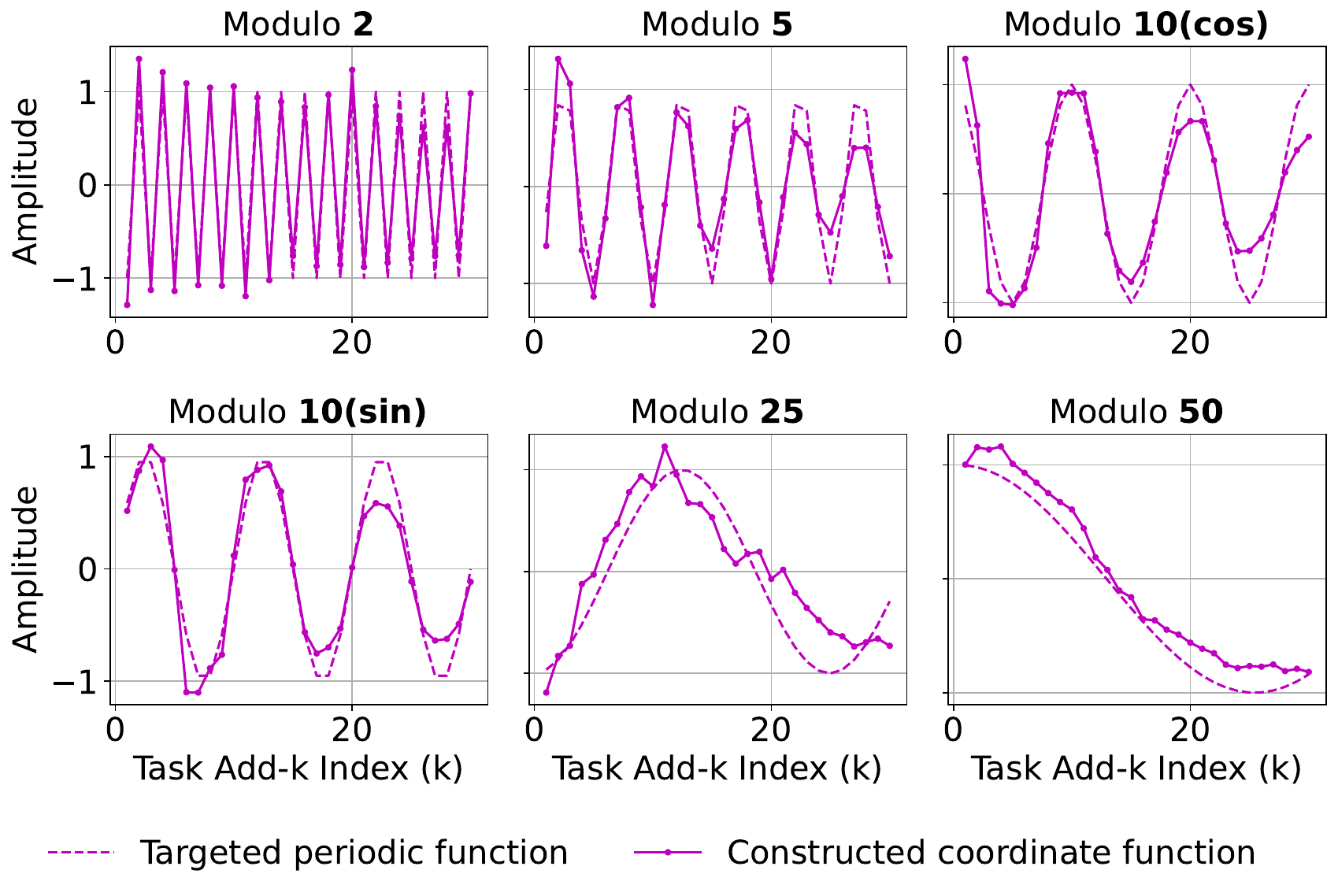}
        \caption{Head $(13,6)$}
        \label{fig:fitted_separated_head3}
    \end{subfigure}
    \caption{Coordinate functions of three heads can fit six trigonometric functions with periods $2, 5, 10, 10, 25$, and $50$ on Llama-3-8B-Instruct.}
    \label{fig:fitted_group}
\end{figure}

In \S\ref{sec:periodic}, we found six trigonometric functions that can be linearly fitted by the coordinate functions. Here we first supplement the plot from which we observe the periodic pattern of the coordinate functions for the three heads (Figure~\ref{fig:periodic_group}) and then show the fitting functions for the three heads (Figure~\ref{fig:fitted_group}).

\clearpage

\subsection{Phi-4-14B for \S\ref{sec:periodic}}\label{app:periodic_phi4}
\begin{figure}[htbp]
    \centering
    \begin{subfigure}{0.8\linewidth}
        \includegraphics[width=\linewidth]{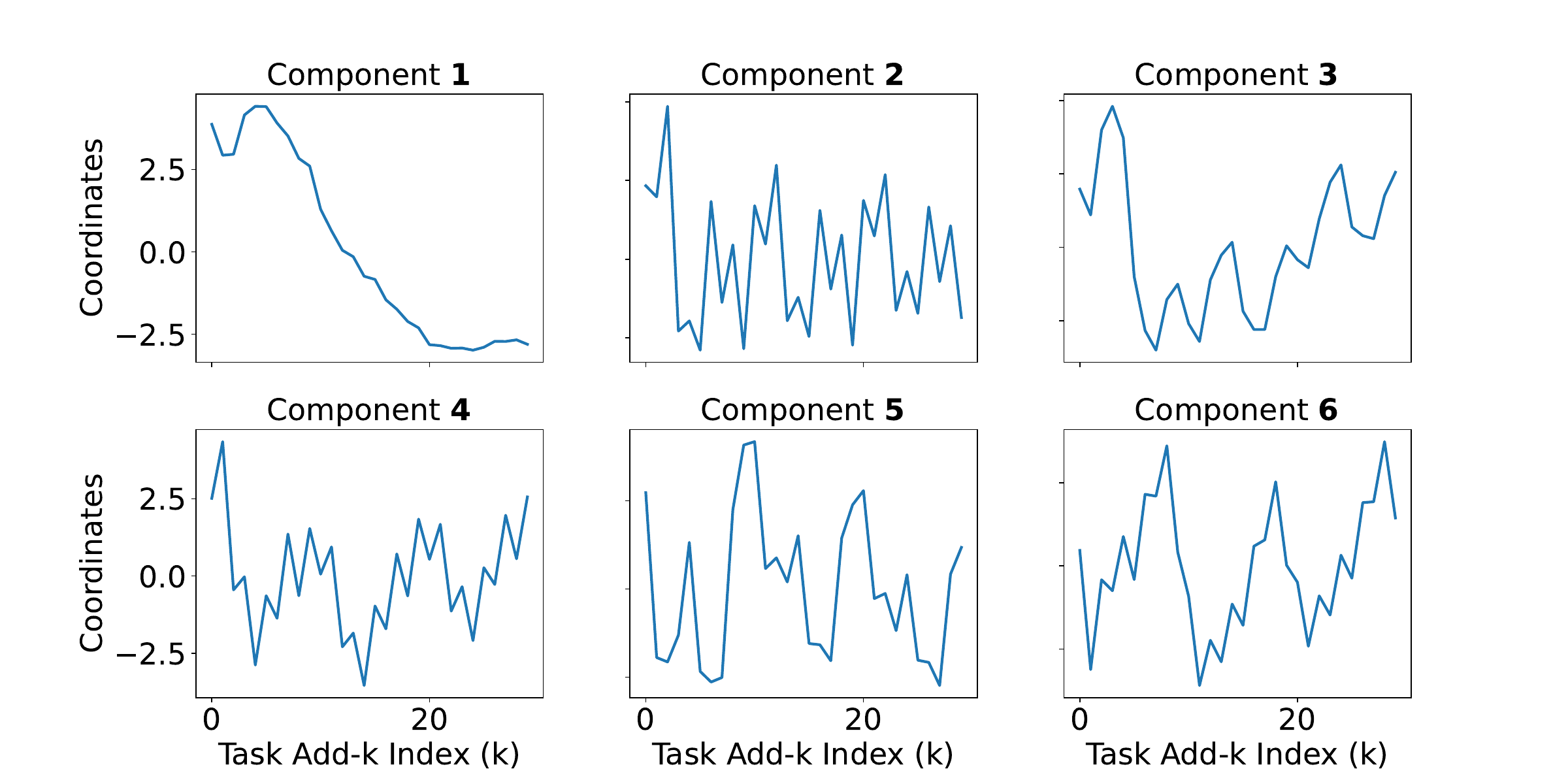}
        \caption{Head $(18,12)$}
    \end{subfigure}\hfill
    \begin{subfigure}{0.8\linewidth}
        \includegraphics[width=\linewidth]{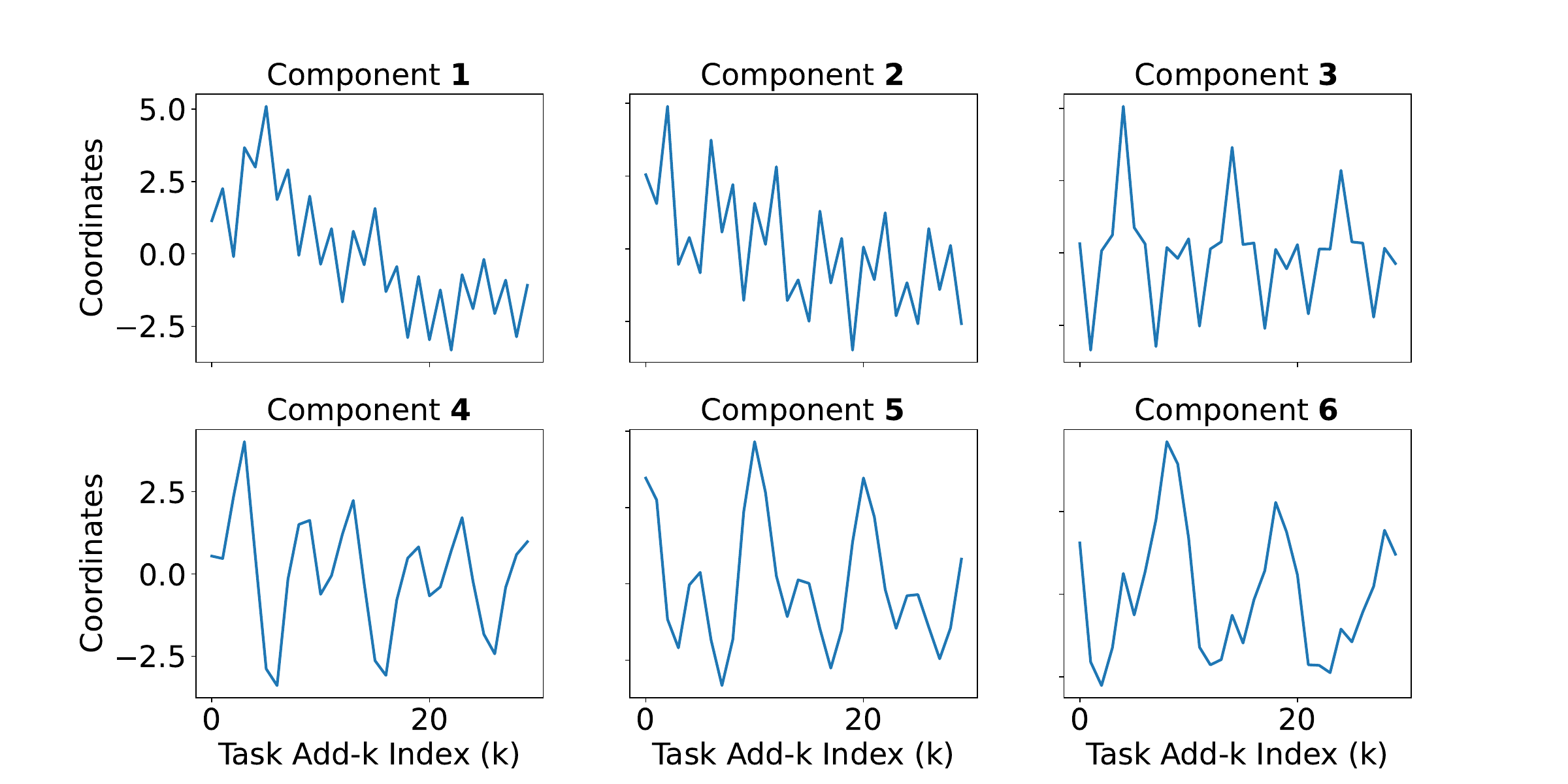}
        \caption{Head $(18,14)$}
    \end{subfigure}\hfill
    \begin{subfigure}{0.8\linewidth}
        \includegraphics[width=\linewidth]{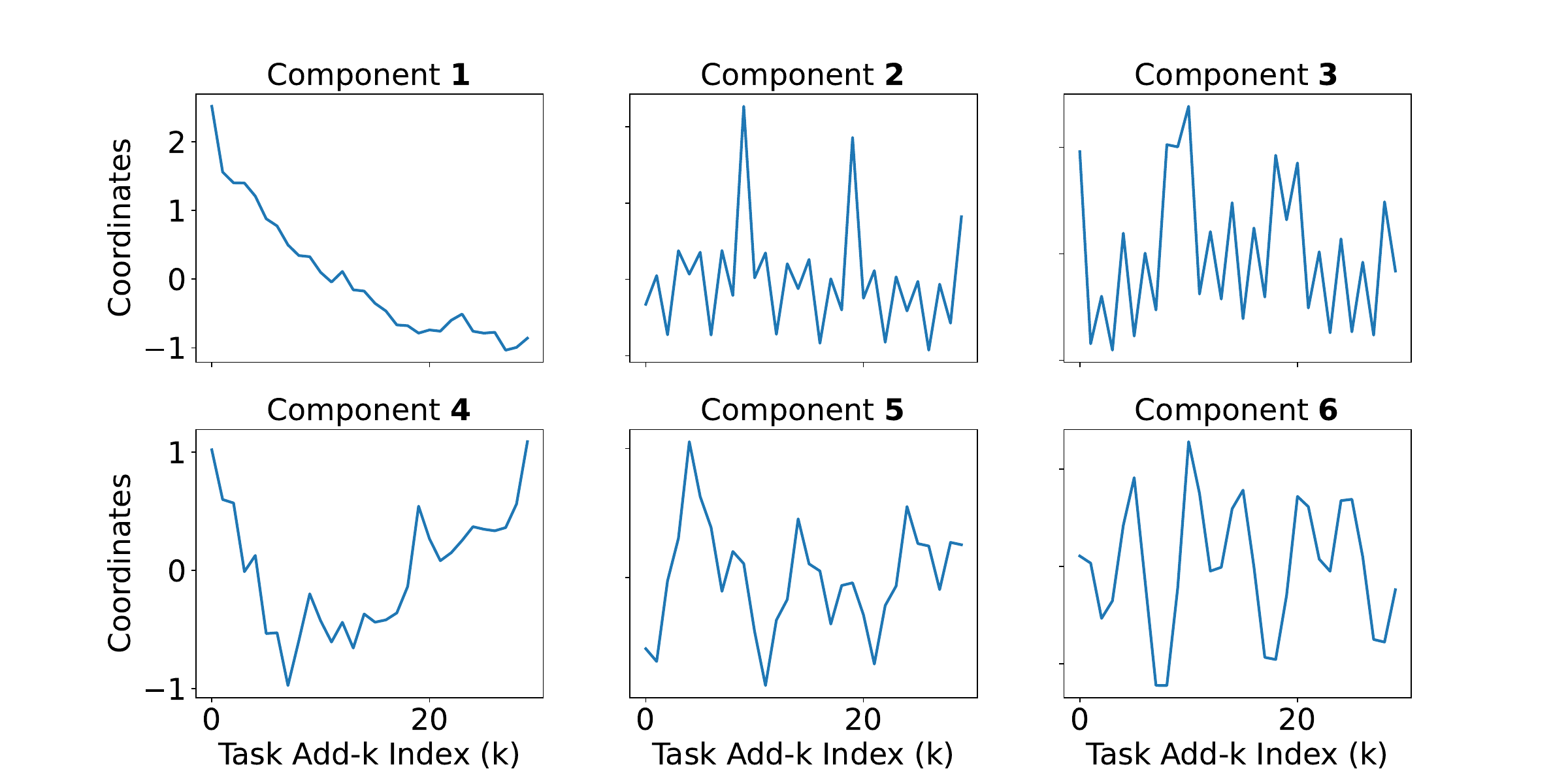}
        \caption{Head $(17,23)$}
    \end{subfigure}
    \caption{Coordinates of the three Phi-4 main heads' vectors (inner products with PCs) for the first six PCs across different add-$k$ tasks. Periodic patterns are visible in the first few PCs of each head.}
    \label{fig:periodic_phi4}
\end{figure}

\begin{figure}[htbp]
    \centering
    \begin{subfigure}{0.7\linewidth}
        \includegraphics[width=\linewidth]{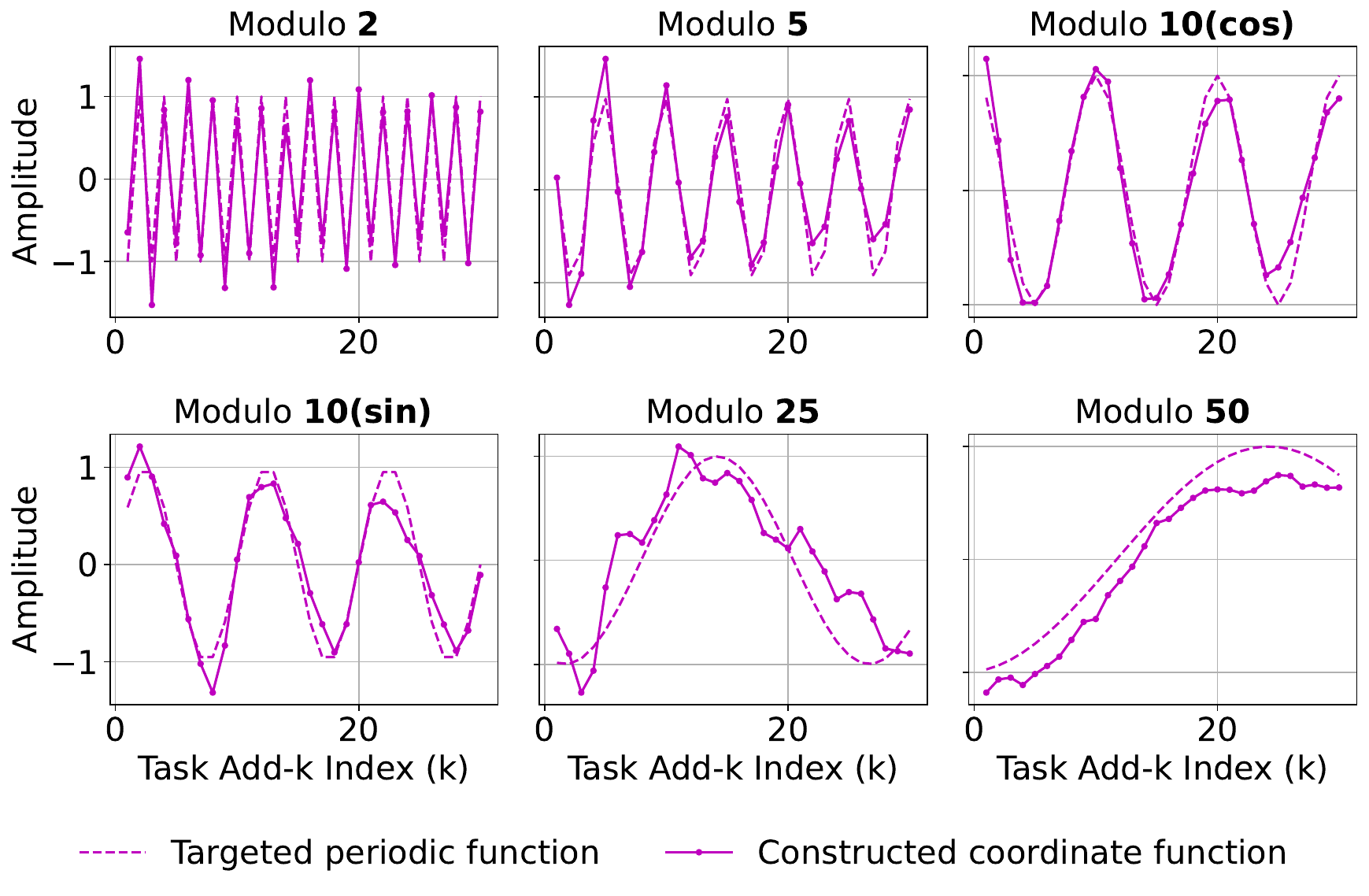}
        \caption{Head $(18,12)$}
    \end{subfigure}\hfill
    \begin{subfigure}{0.7\linewidth}
        \includegraphics[width=\linewidth]{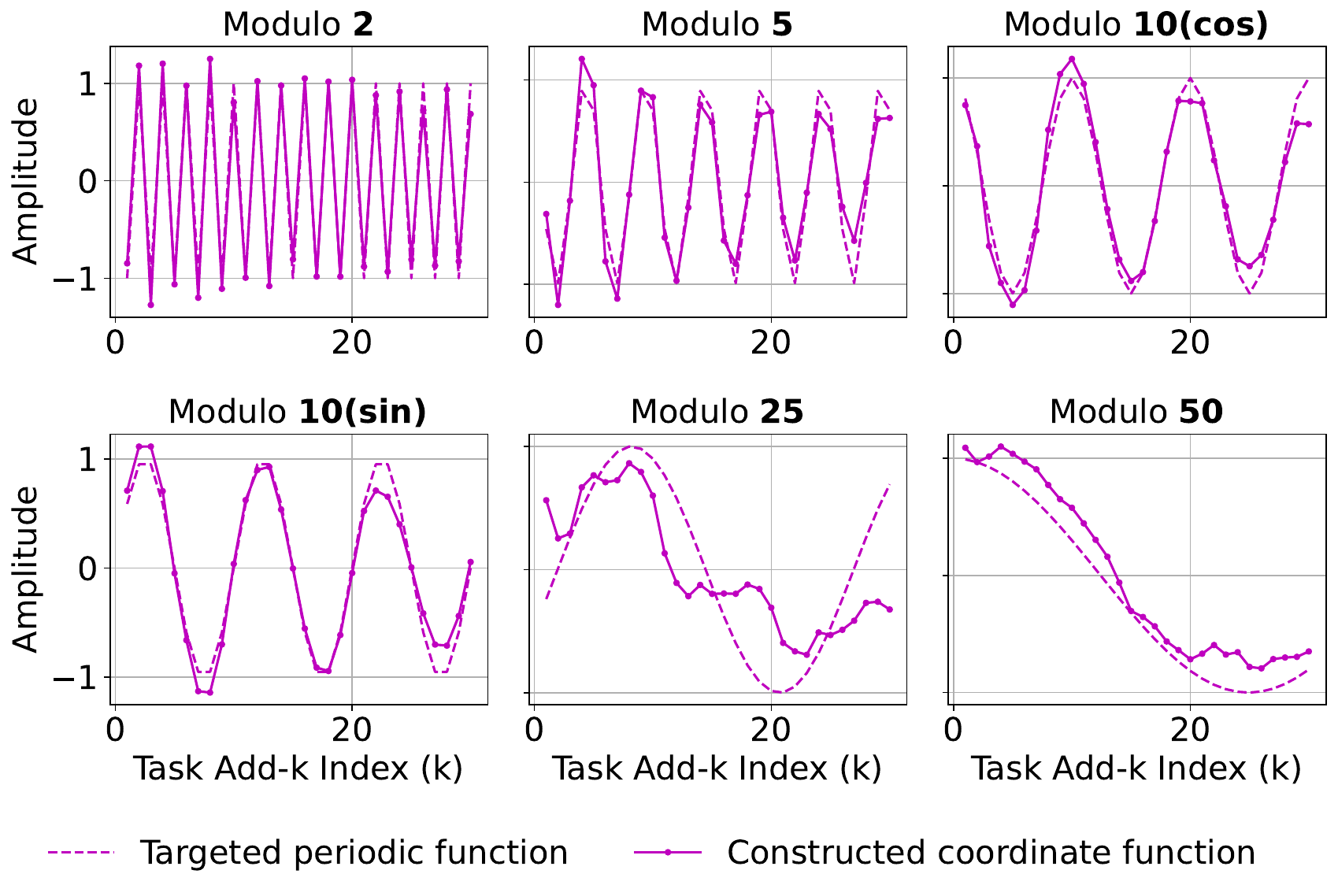}
        \caption{Head $(18,14)$}
    \end{subfigure}\hfill
    \begin{subfigure}{0.7\linewidth}
        \includegraphics[width=\linewidth]{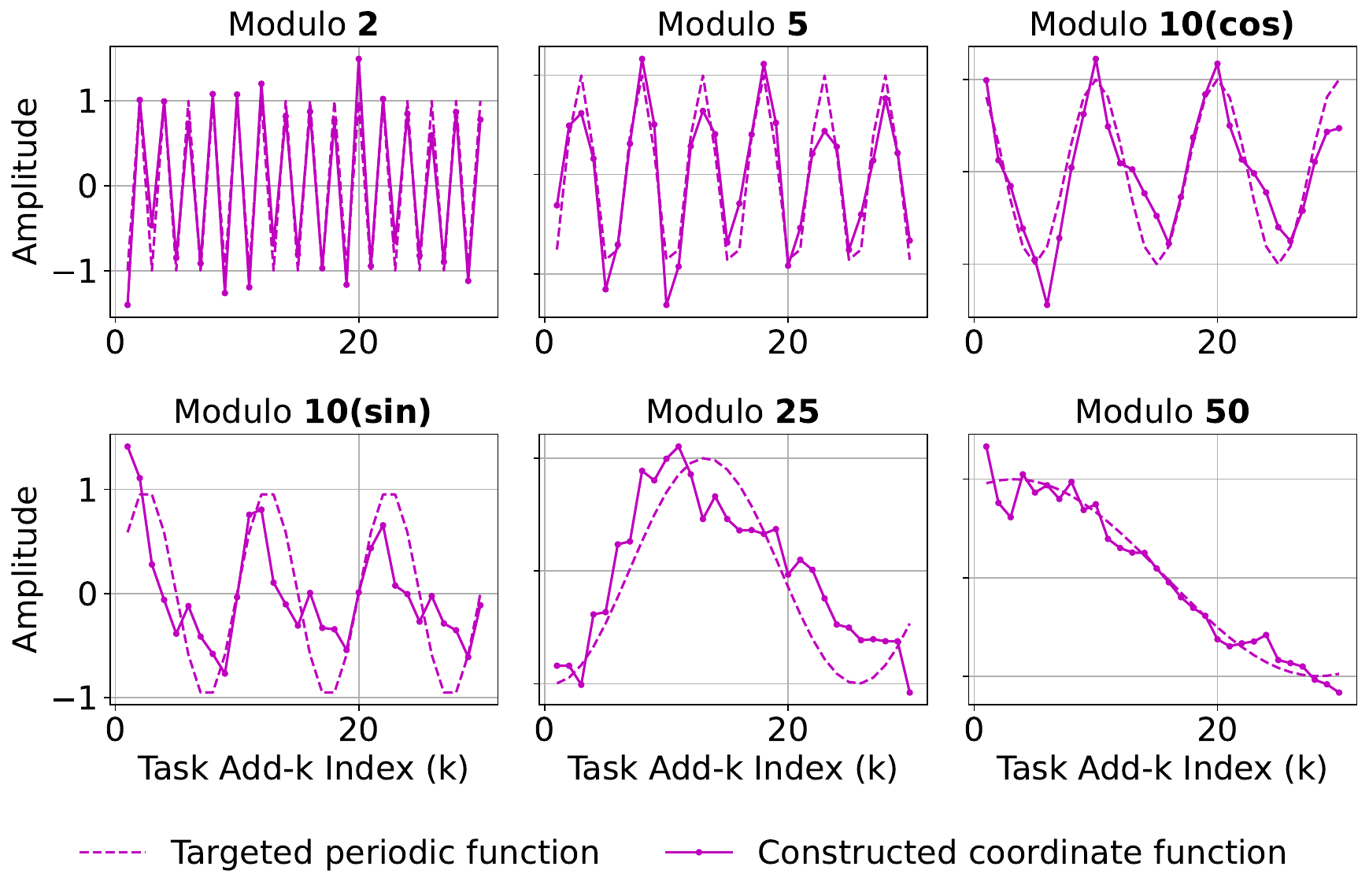}
        \caption{Head $(17,23)$}
    \end{subfigure}
    \caption{Coordinate functions of the three Phi-4 main heads fit six trigonometric functions with periods $2, 5, 10, 10, 25$, and $50$.}
    \label{fig:fitted_phi4}
\end{figure}

The Phi-4 main heads show similar periodic structure to Llama-3-8B-Instruct: their PC coordinates are periodic in $k$ (Figure~\ref{fig:periodic_phi4}) and fit the six trigonometric patterns well (Figure~\ref{fig:fitted_phi4}).

\clearpage
\subsection{Qwen-2.5-7B for \S\ref{sec:periodic}}\label{app:periodic_qwen}
\begin{figure}[htbp]
    \centering
    \begin{subfigure}{0.49\linewidth}
        \includegraphics[width=\linewidth]{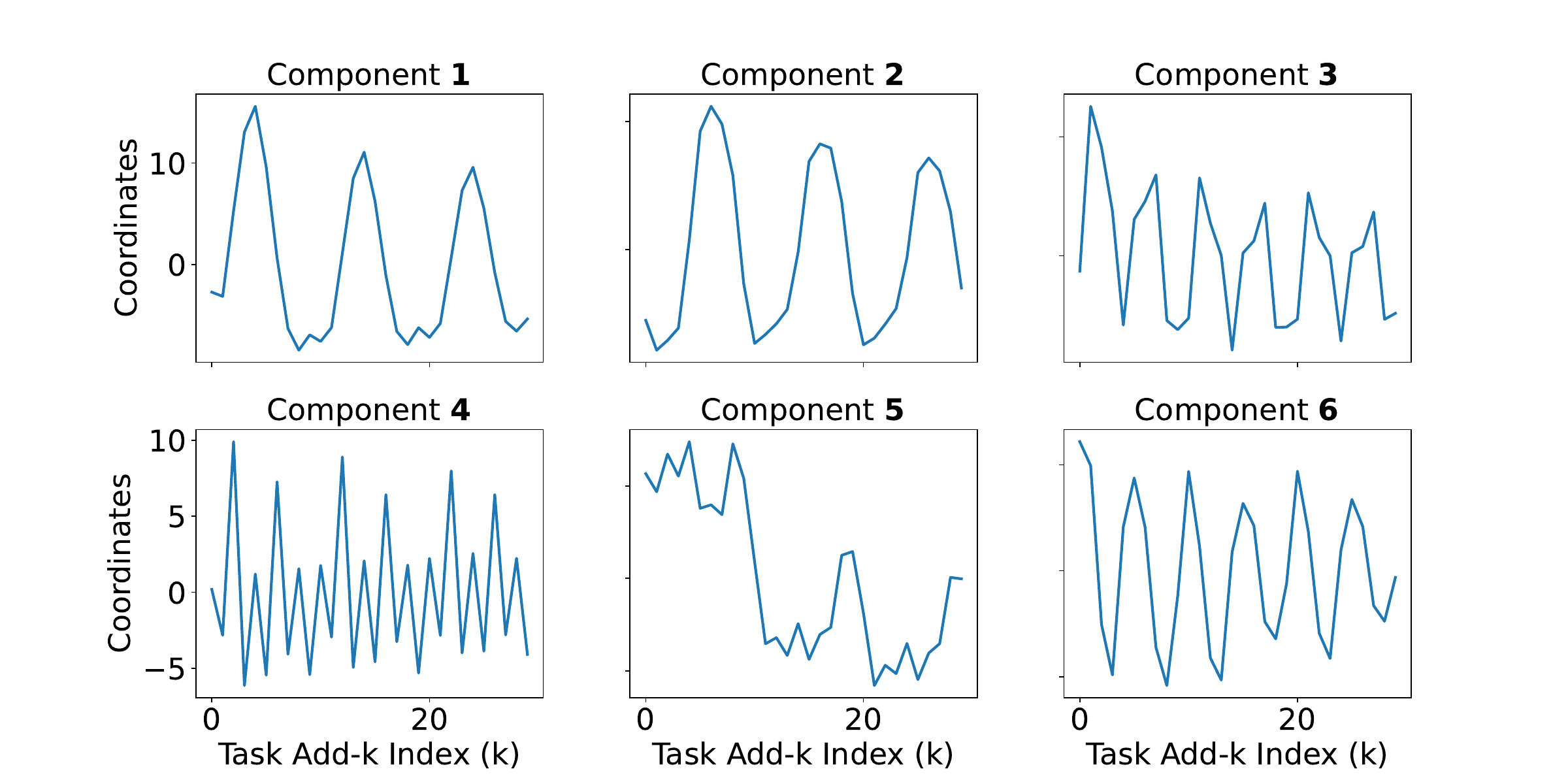}
        \caption{Head $(21,0)$}
    \end{subfigure}\hfill
    \begin{subfigure}{0.49\linewidth}
        \includegraphics[width=\linewidth]{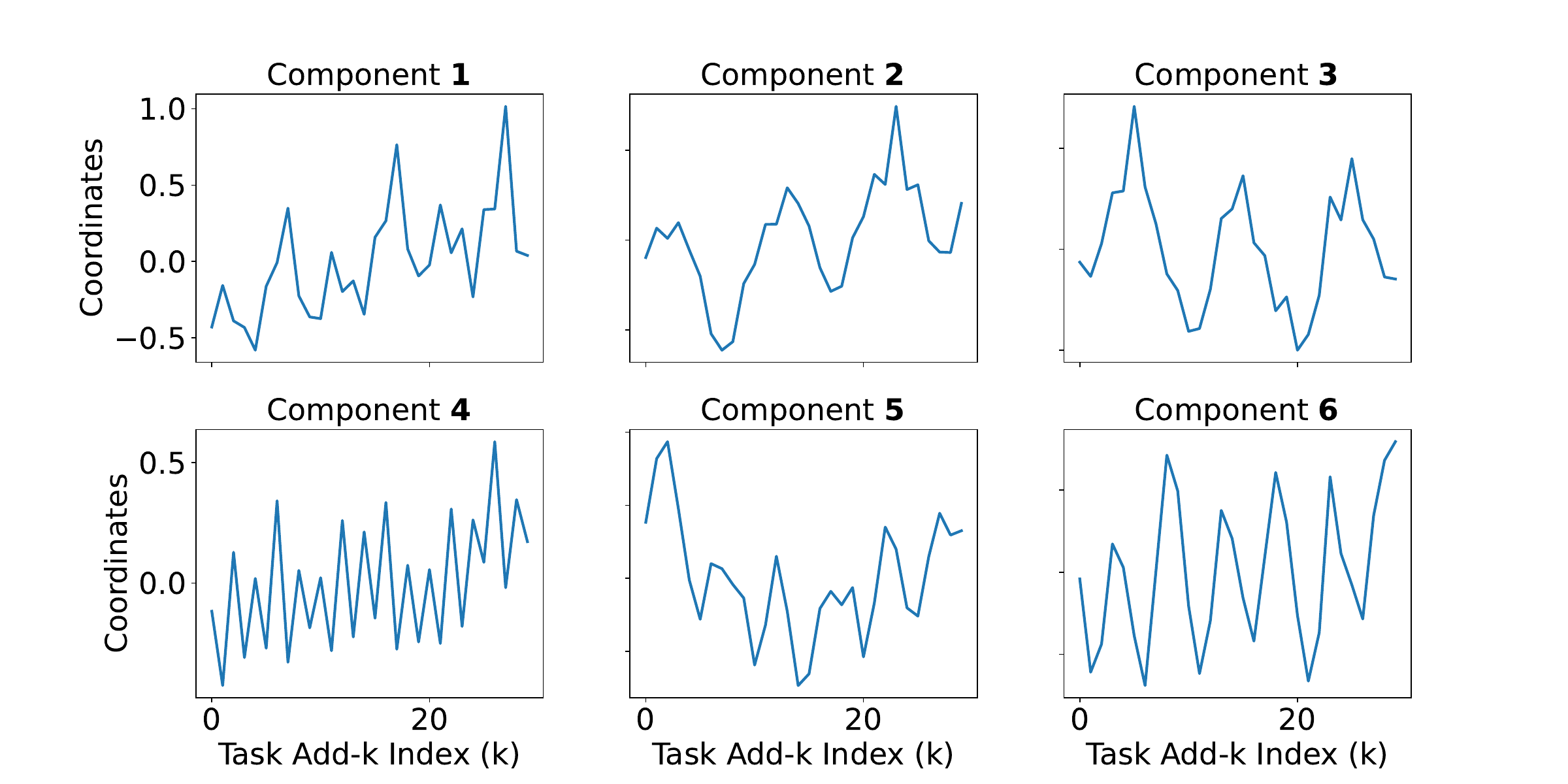}
        \caption{Head $(21,4)$}
    \end{subfigure}

    \begin{subfigure}{0.49\linewidth}
        \includegraphics[width=\linewidth]{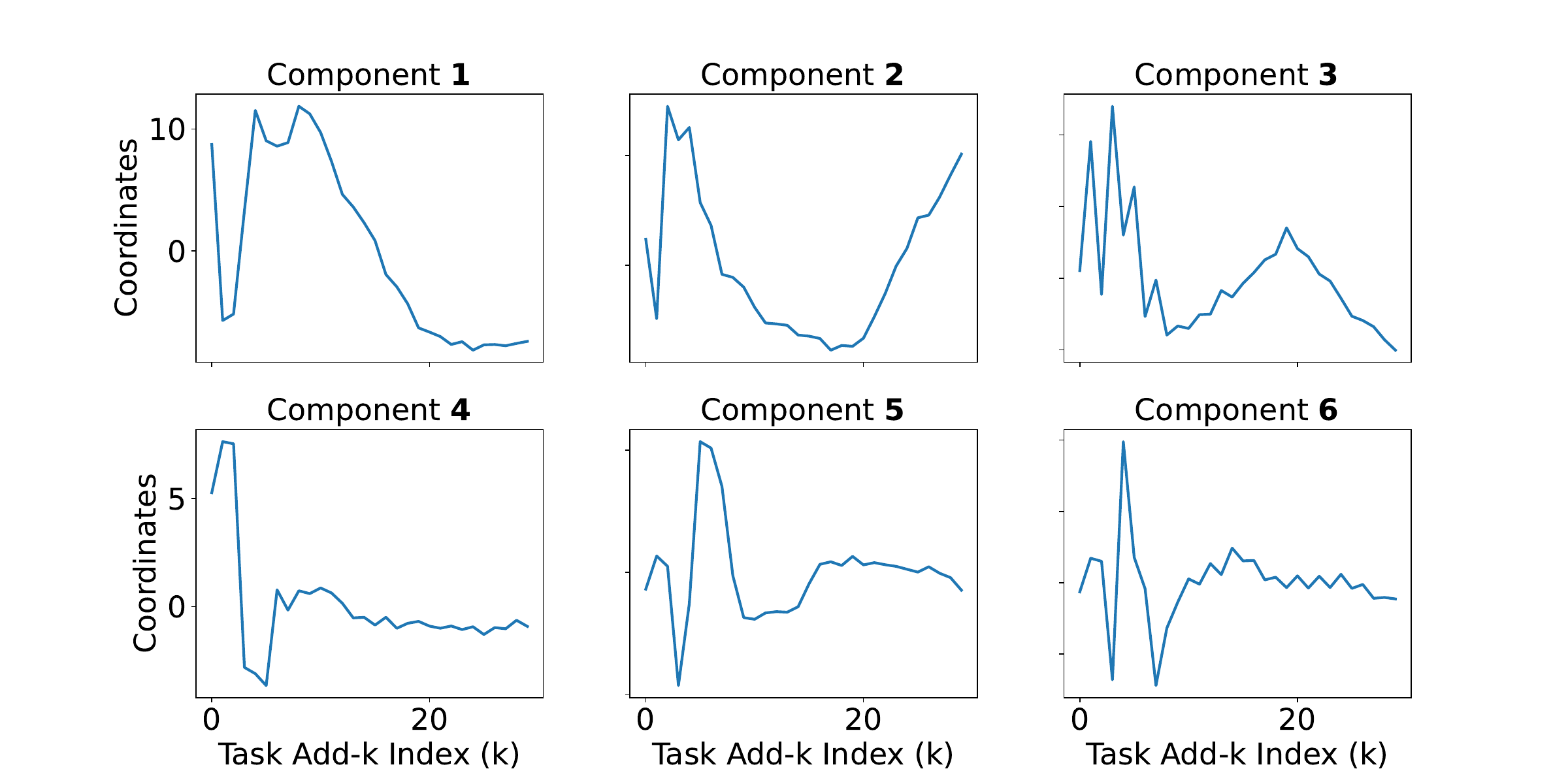}
        \caption{Head $(21,2)$}
    \end{subfigure}\hfill
    \begin{subfigure}{0.49\linewidth}
        \includegraphics[width=\linewidth]{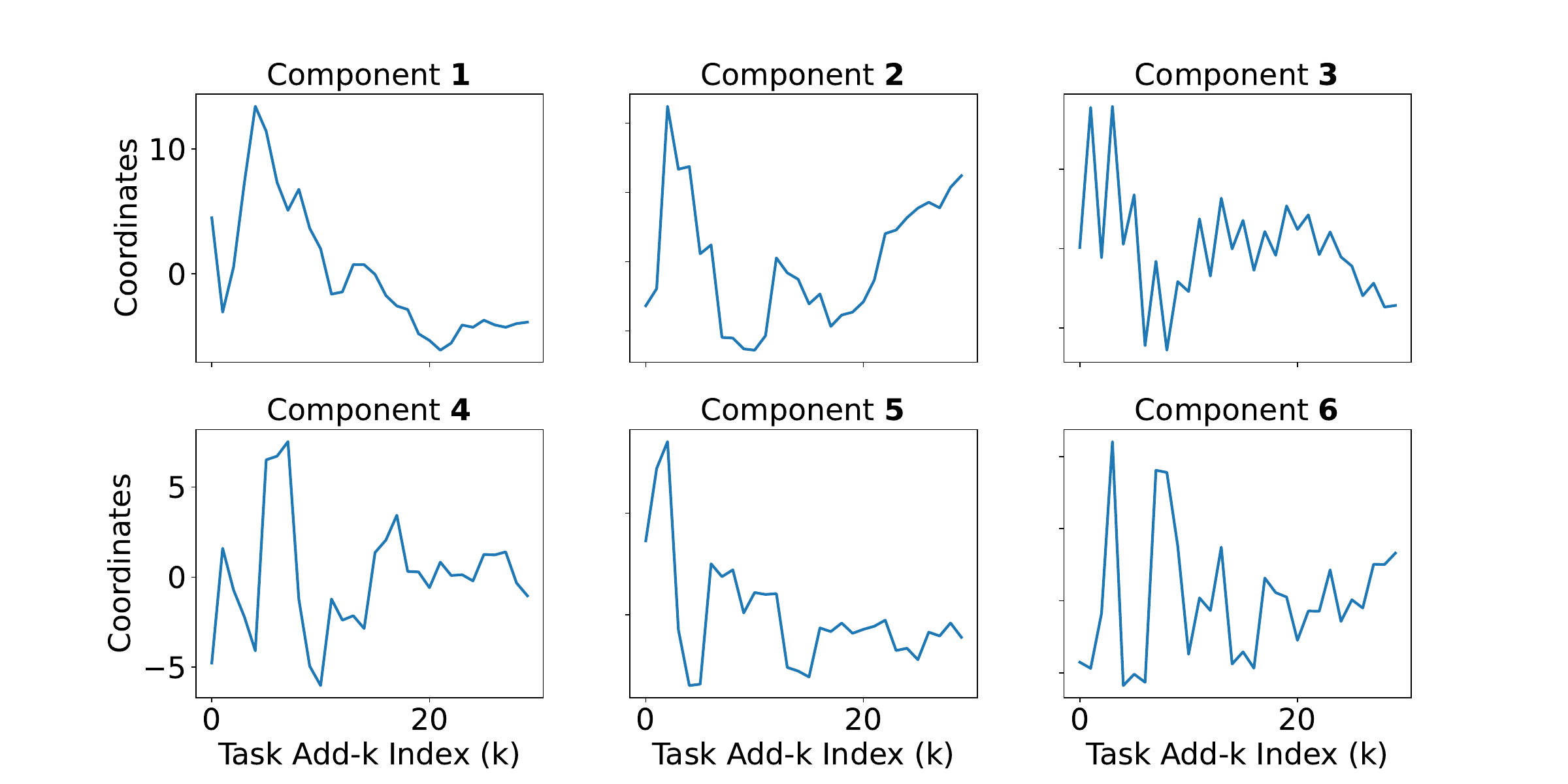}
        \caption{Head $(21,5)$}
    \end{subfigure}
    \caption{Coordinates of the four Qwen-2.5-7B main heads' vectors (inner products with PCs) for the first six PCs across different add-$k$ tasks. Heads $(21,0)$ and $(21,4)$ reveal periodic patterns, whereas heads $(21,2)$ and $(21,5)$ show discontinuous behaviors across the intervals $[1,10]$, $[11,20]$, and $[21,30]$.}
    \label{fig:periodic_qwen}
\end{figure}

\begin{figure}[htbp]
    \centering
    \begin{subfigure}{0.49\linewidth}
        \includegraphics[width=\linewidth]{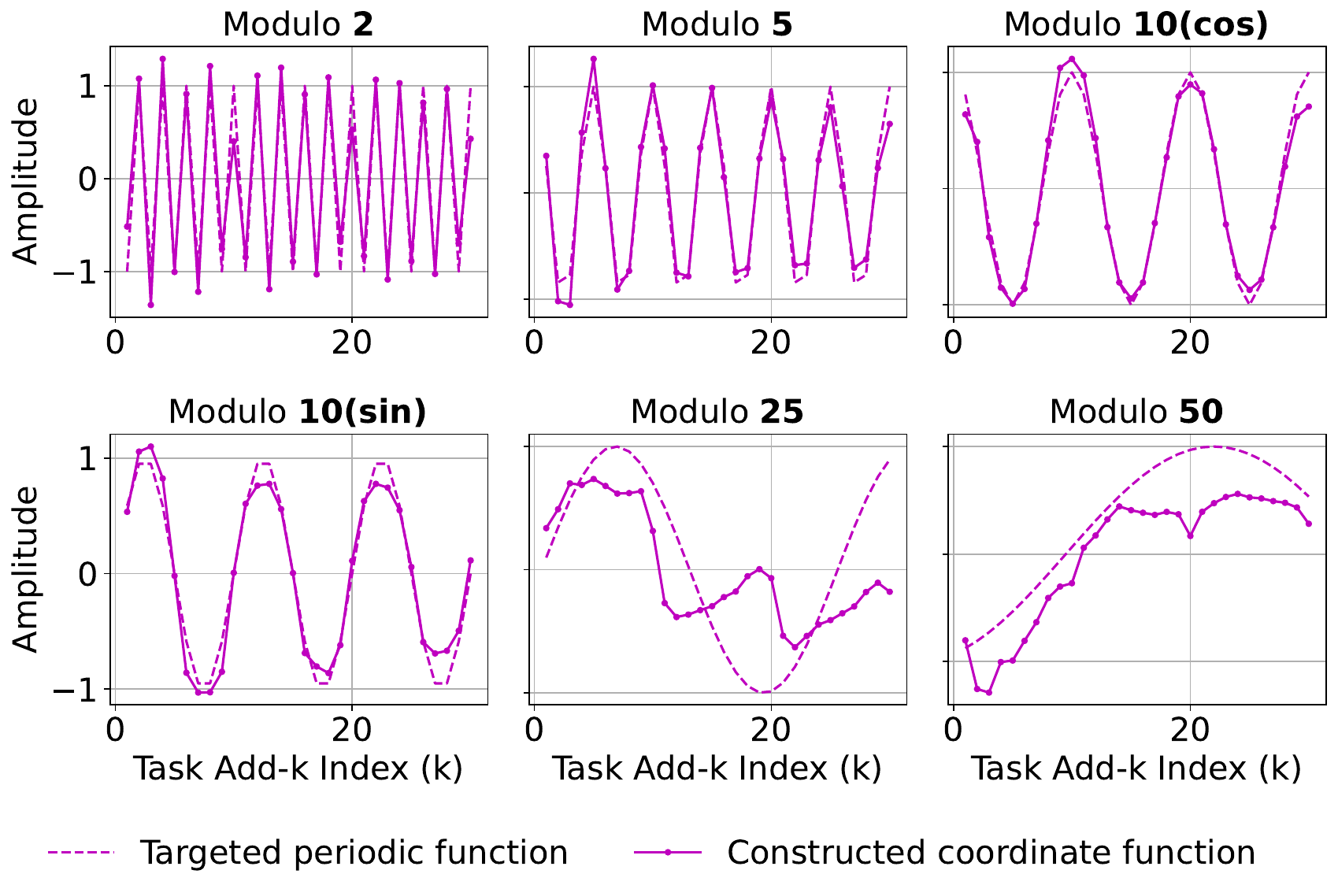}
        \caption{Head $(21,0)$}
    \end{subfigure}\hfill
    \begin{subfigure}{0.49\linewidth}
        \includegraphics[width=\linewidth]{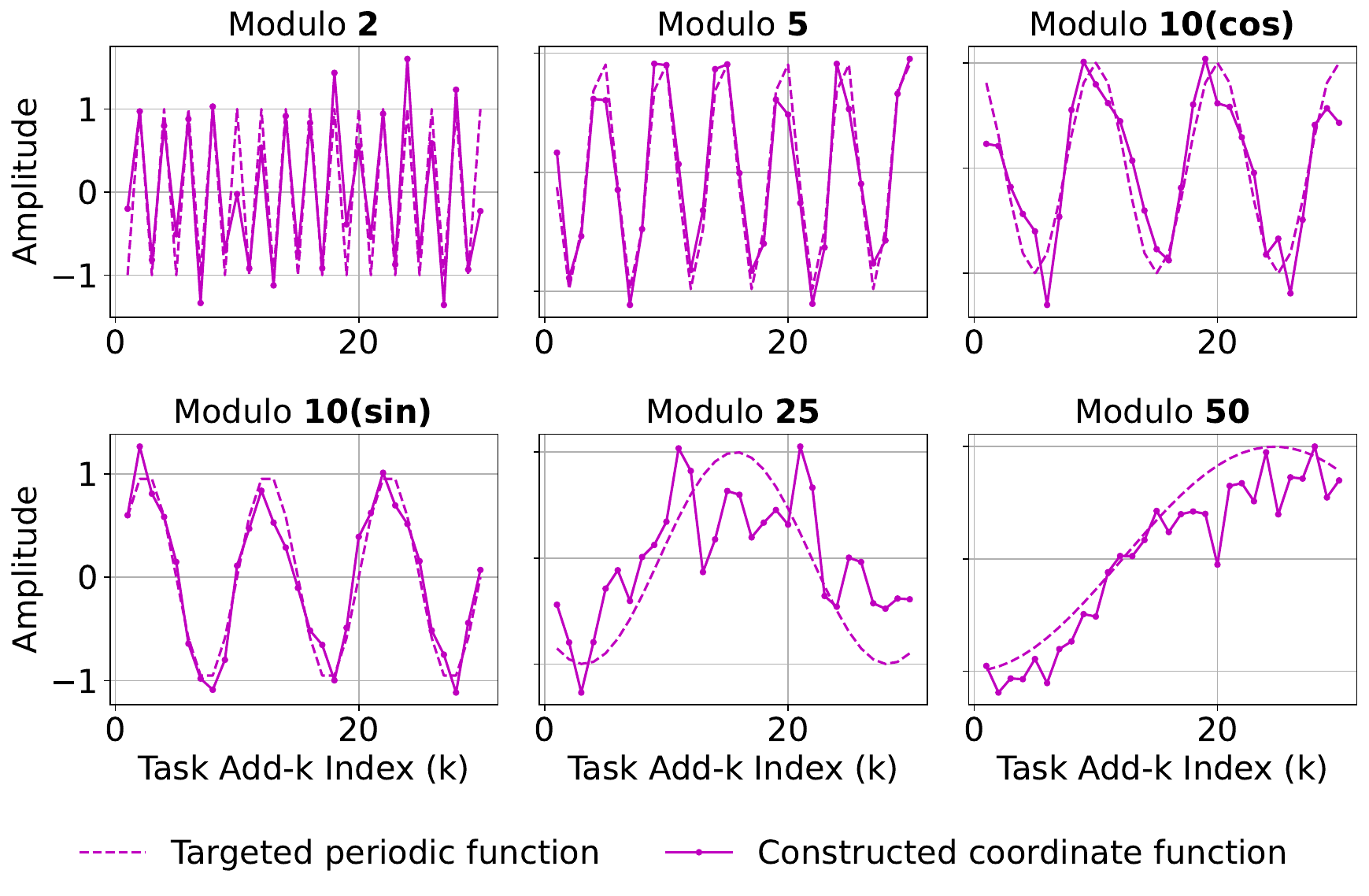}
        \caption{Head $(21,4)$}
    \end{subfigure}

    \begin{subfigure}{0.49\linewidth}
        \includegraphics[width=\linewidth]{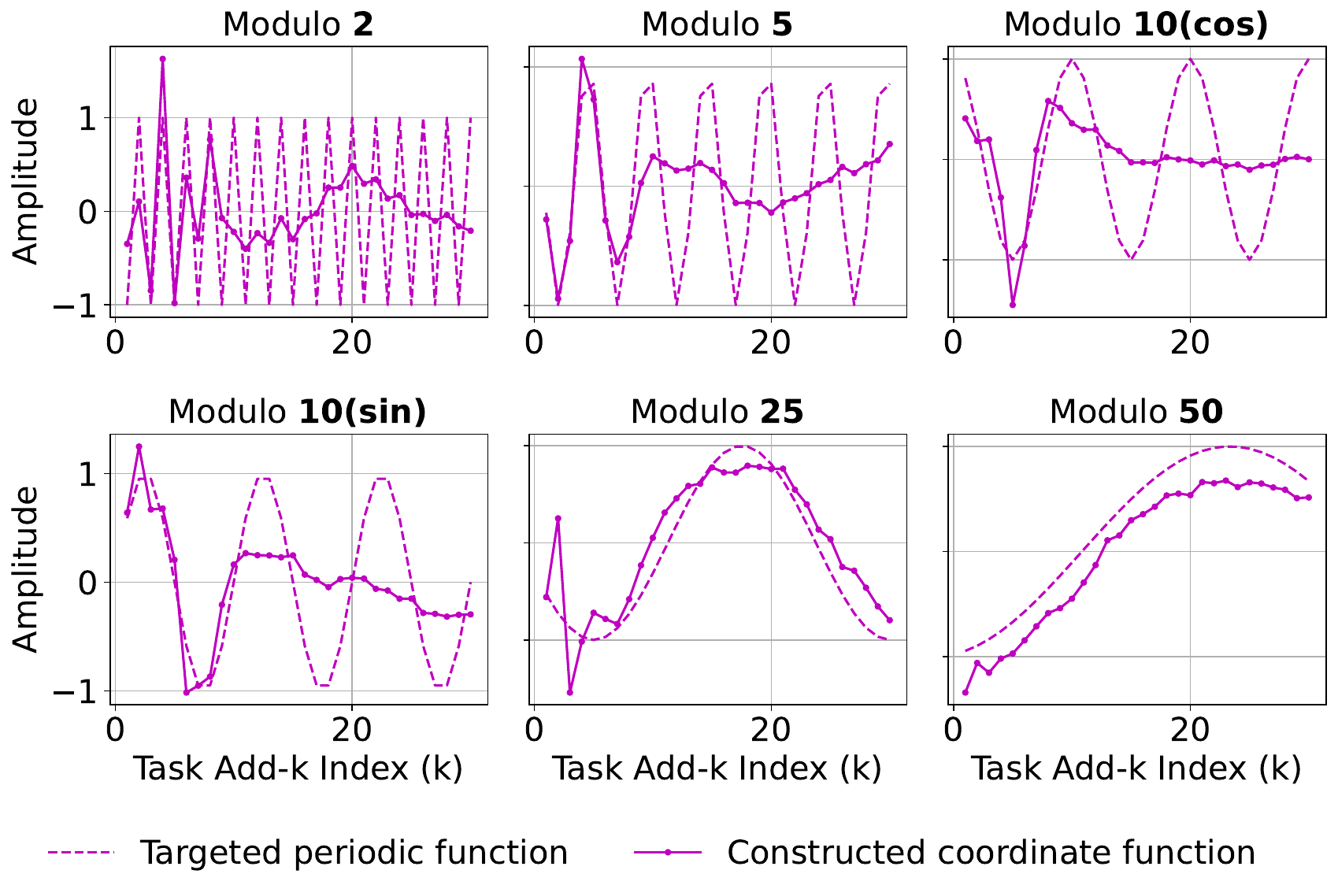}
        \caption{Head $(21,2)$}
    \end{subfigure}\hfill
    \begin{subfigure}{0.49\linewidth}
        \includegraphics[width=\linewidth]{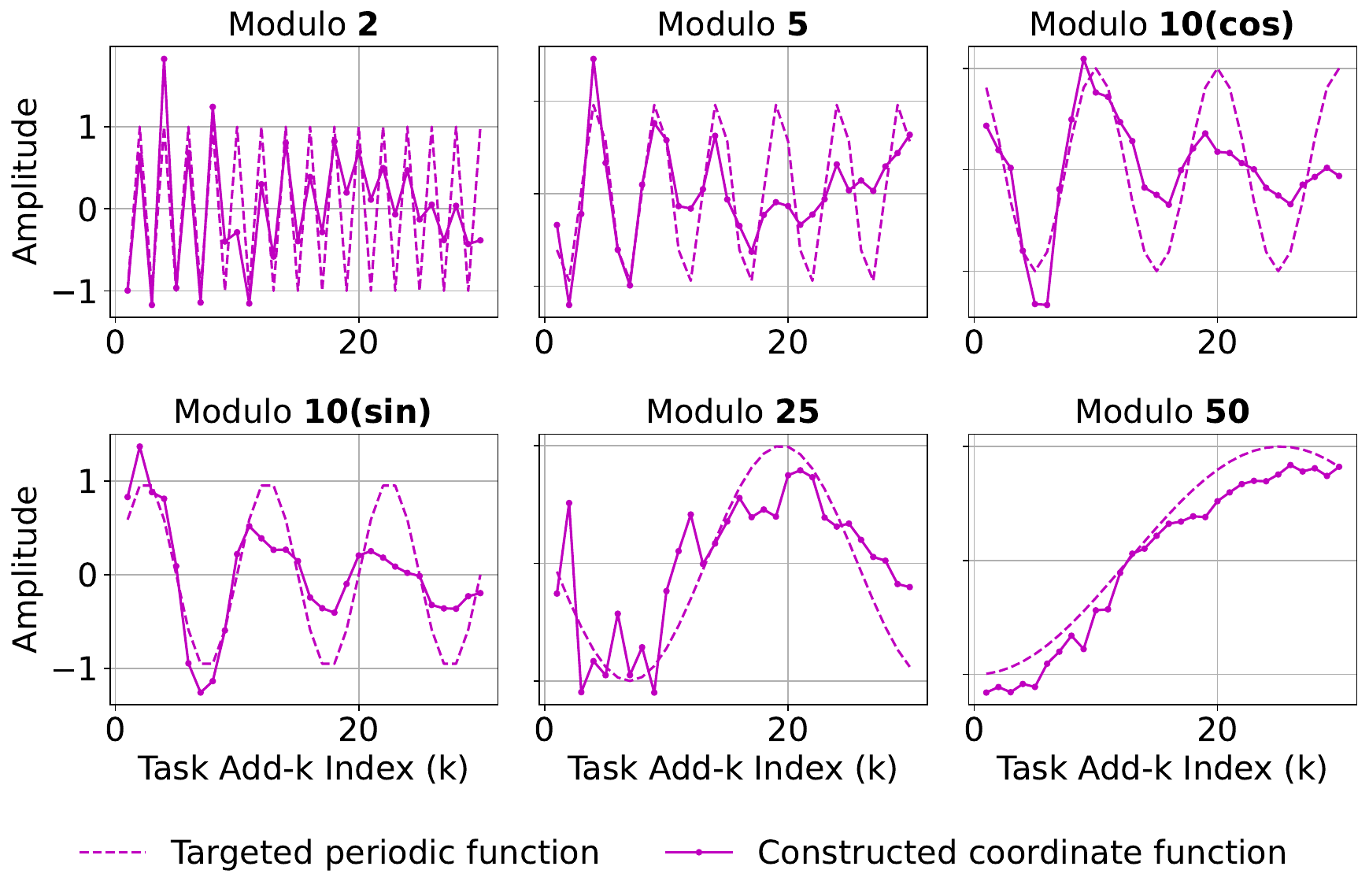}
        \caption{Head $(21,5)$}
    \end{subfigure}
    \caption{Six trigonometric functions with periods $2, 5, 10, 10, 25$, and $50$ fitted by the coordinate functions of the four Qwen-2.5-7B main heads. Heads $(21,0)$ and $(21,4)$ fit the periodic functions well, while heads $(21,2)$ and $(21,5)$ fit only the low-frequency directions, consistent with their non-periodic coordinate behaviors.}
    \label{fig:fitted_qwen}
\end{figure}

Heads $(21,0)$ and $(21,4)$ show periodic patterns similar to the Llama-3 heads, while heads $(21,2)$ and $(21,5)$ do not encode the unit-digit periodicity; their coordinates instead follow the low-frequency magnitude directions and behave distinctly across the intervals $[1,10]$, $[11,20]$, and $[21,30]$ (Figures~\ref{fig:periodic_qwen} and \ref{fig:fitted_qwen}). We conjecture this difference arises from tokenization: Qwen encodes numbers digit by digit, creating discontinuities across each 10-interval.

\clearpage

\section{Supplement for \S\ref{sec:4+2dim}}\label{app:4+2dim}
Recall that the three hypotheses in \S\ref{sec:4+2dim} are as follows:
\begin{enumerate}
    \item [(i)] the feature direction corresponding to period two, which we call the ``parity direction'', encodes the parity of $k$ in the \textit{add-$k$} task; 
    \item [(ii)] the subspace spanned by the feature directions with periods $2,5,10$, which we call the ``unit subspace'', encodes the units digit of $k$;
    \item[(iii)] the subspace spanned by the directions with periods $25,50$, which we call the ``magnitude subspace'', encodes the coarse magnitude (i.e., the tens digit) of $k$. 
\end{enumerate}

\begin{figure}[htbp]
    \centering
    \includegraphics[width=0.97\linewidth]{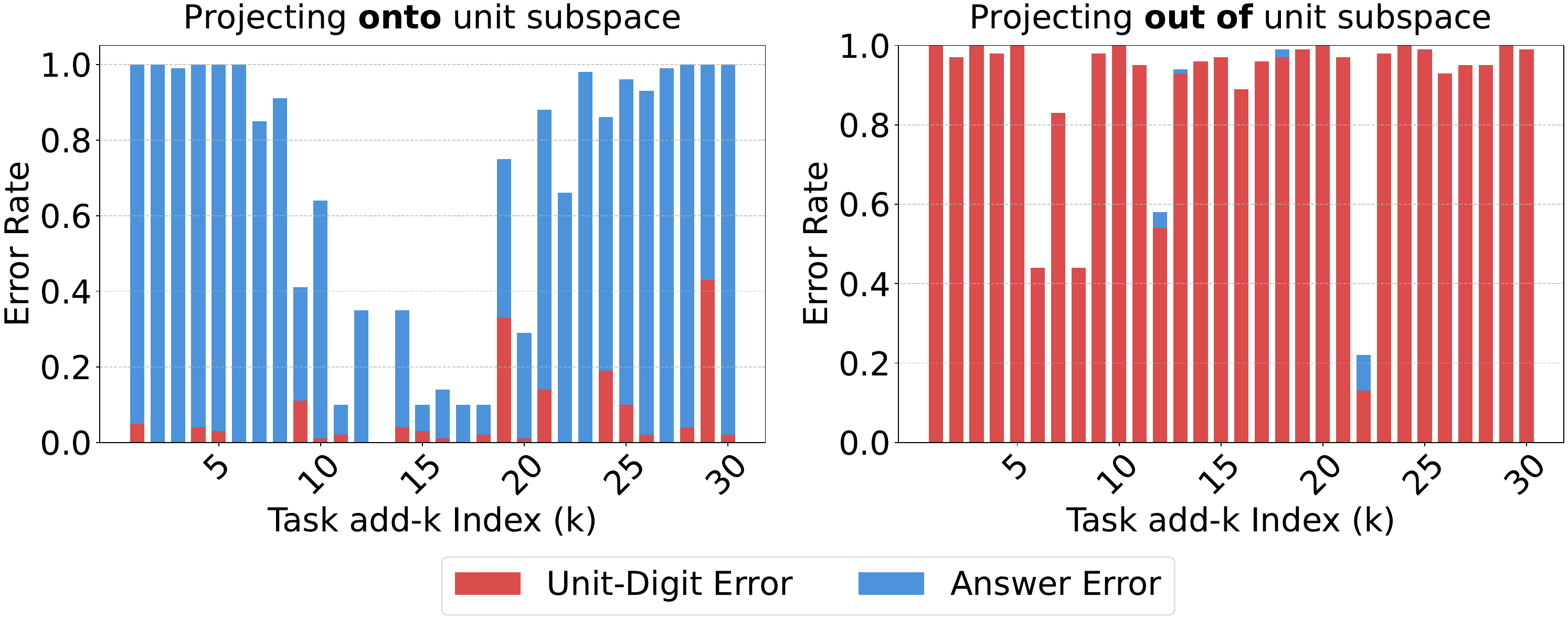}
    \caption{The error rates for the \textcolor{red}{units digit} and the \textcolor{blue}{final answer} across tasks when projecting head 1's vectors \textbf{onto} (left) and \textbf{out of} (right) the ``unit subspace''. Projecting onto the unit subspace yields a low unit-digit error rate even when the final-answer error remains high, while projecting out leads to high unit-digit error rates that almost fully account for final-answer errors. This confirms that the unit subspace encodes the unit-digit signal.}
    \label{fig:onto_out_unit}
\end{figure}

\begin{figure}[htbp]
  \centering

  \begin{subfigure}{0.48\linewidth}
    \includegraphics[width=\linewidth]{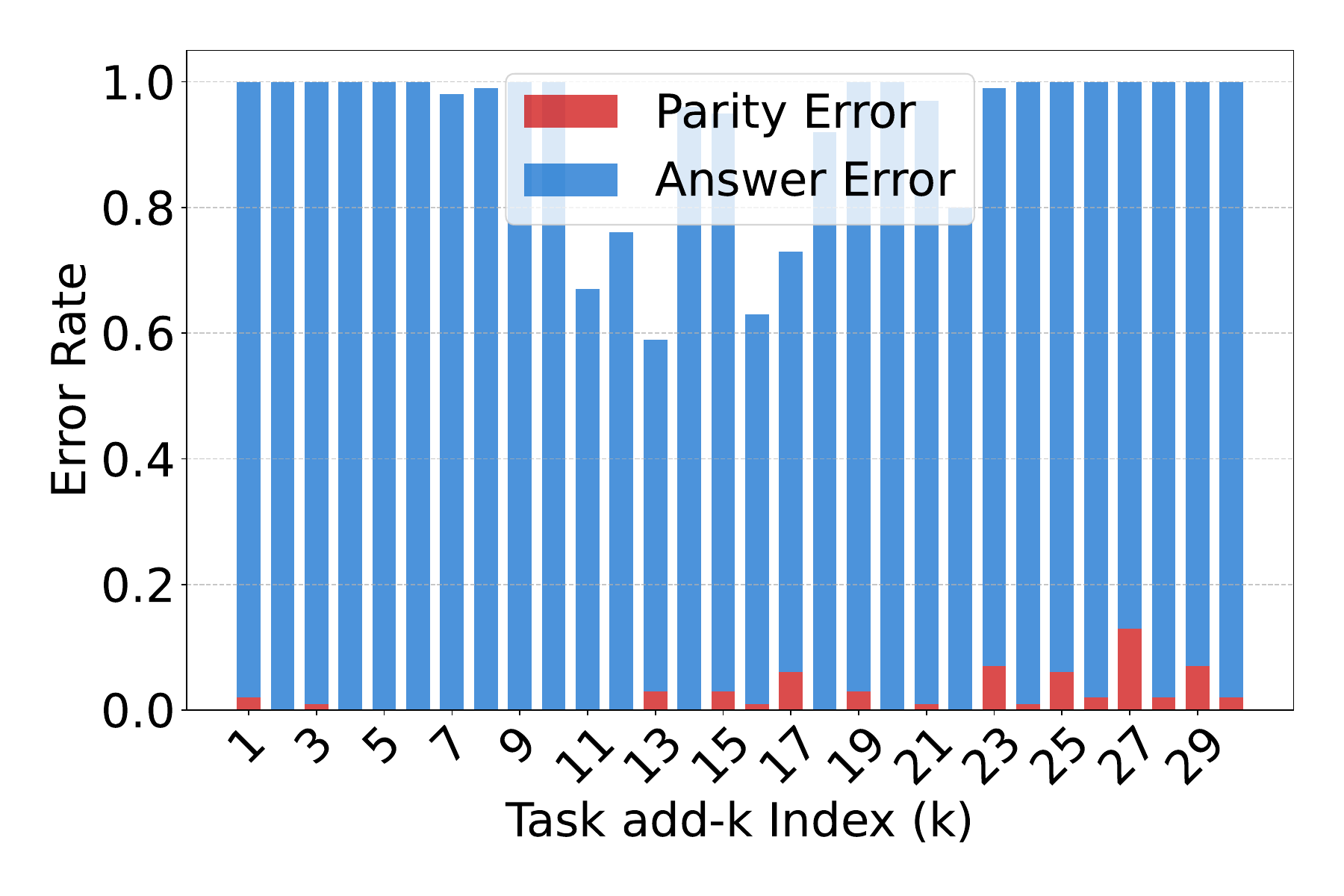}
    \caption{Onto parity direction}
  \end{subfigure}\hfill
  \begin{subfigure}{0.48\linewidth}
    \includegraphics[width=\linewidth]{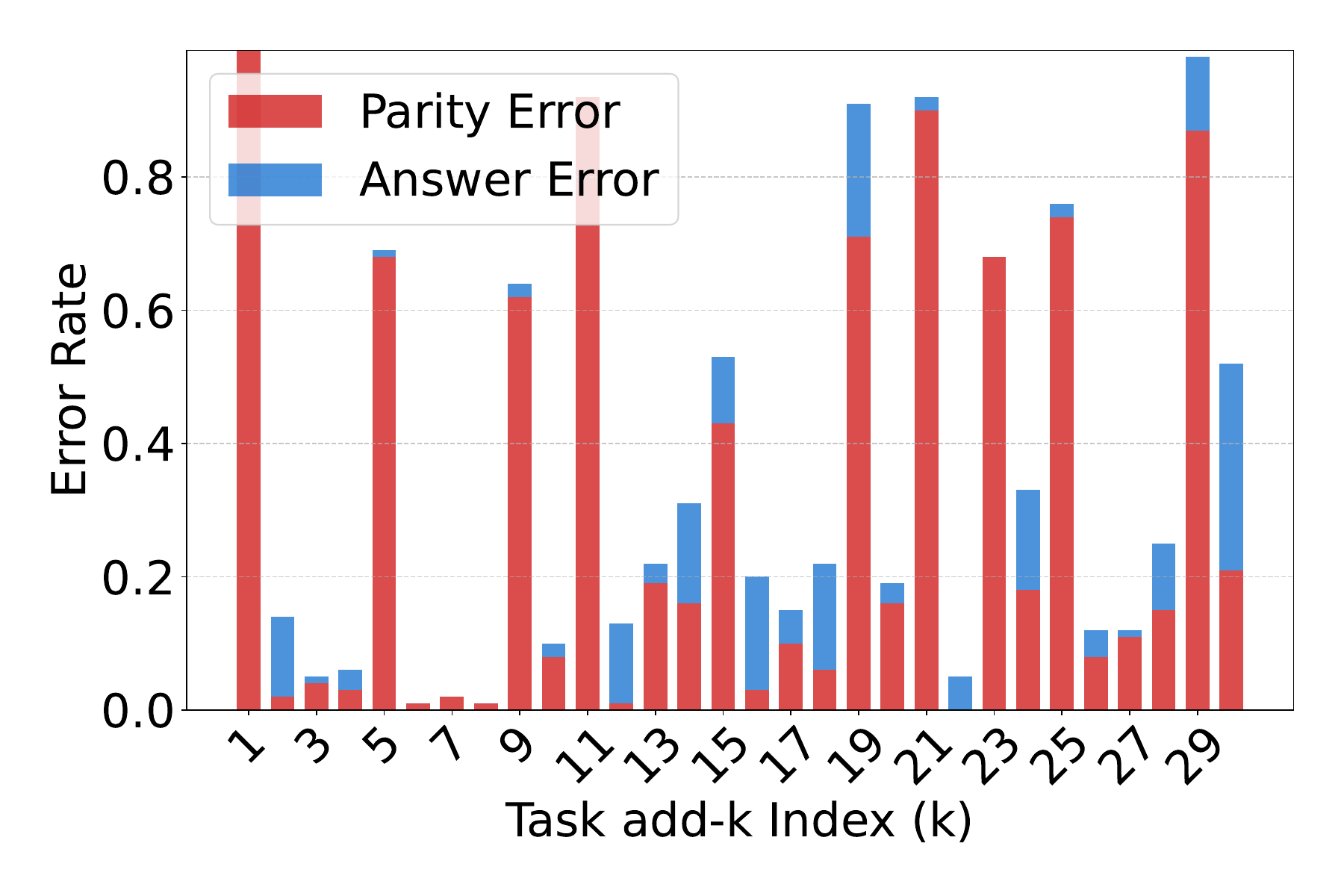}
    \caption{Out of parity direction}
  \end{subfigure}

  \vspace{1em}
  \begin{subfigure}{0.48\linewidth}
    \includegraphics[width=\linewidth]{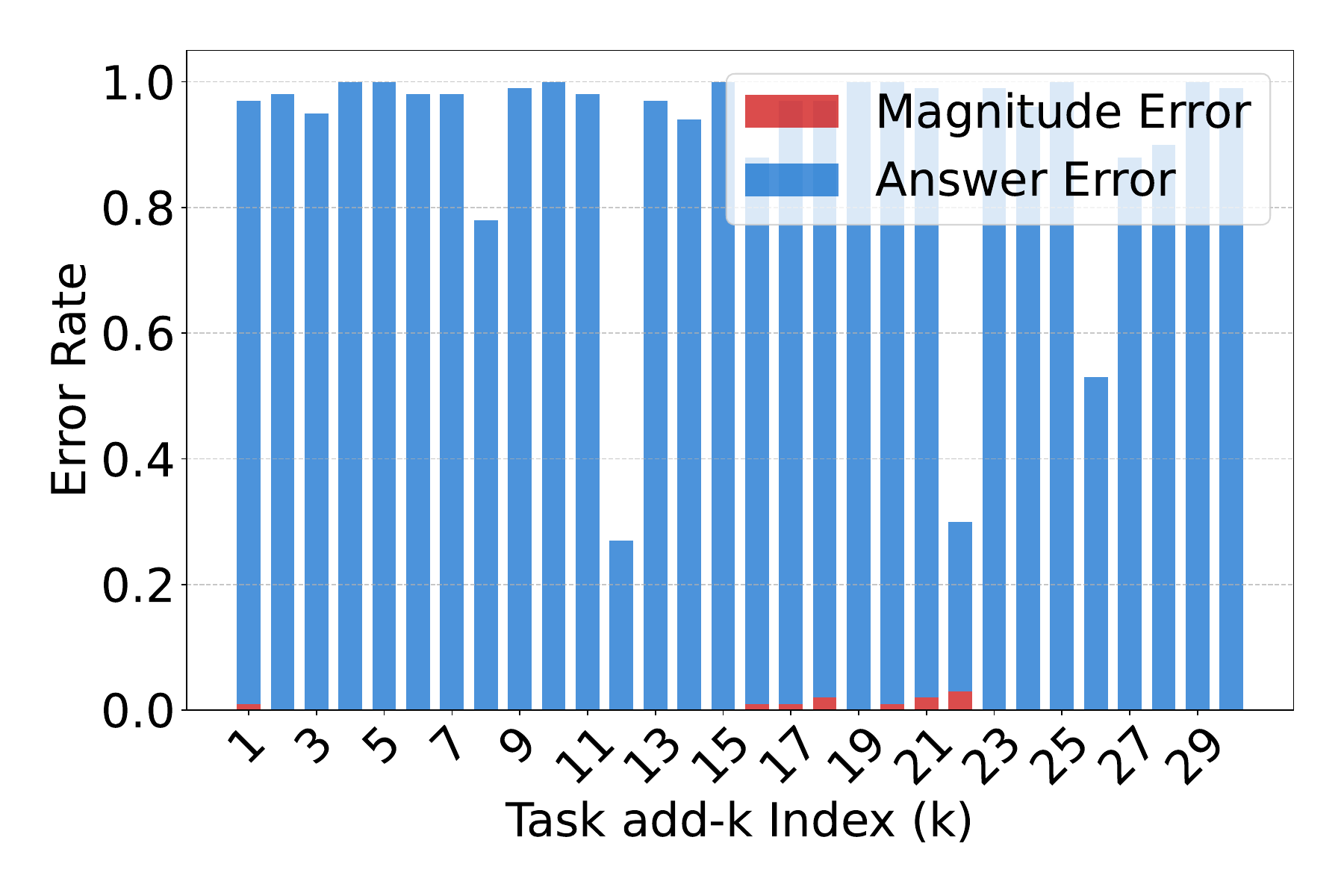}
    \caption{Onto magnitude subspace}
  \end{subfigure}\hfill
  \begin{subfigure}{0.48\linewidth}
    \includegraphics[width=\linewidth]{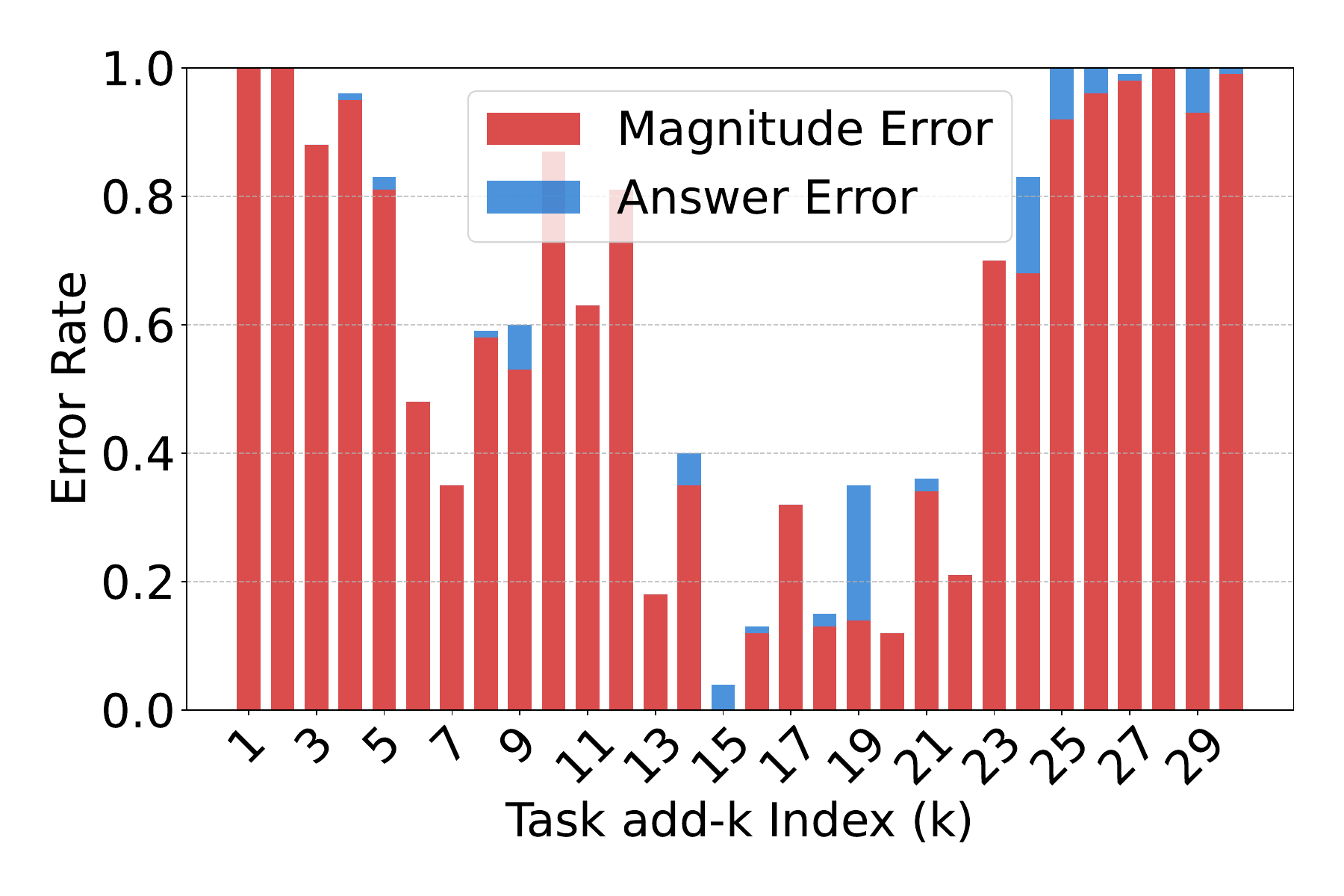}
    \caption{Out of magnitude subspace}
  \end{subfigure}

  \caption{Validation of hypotheses (i) and (iii) for head 1.  
  Each row shows results for the parity and magnitude subspaces (left: projection \textbf{onto}; right: projection \textbf{out of}).  
  The unit-digit hypothesis (ii) is omitted here since it is already shown in Figure~\ref{fig:onto_out_unit}.}
  \label{fig:head1_combined}
\end{figure}

\begin{figure}[htbp]
    \centering
    \begin{subfigure}{0.48\linewidth}
        \includegraphics[width=\linewidth]{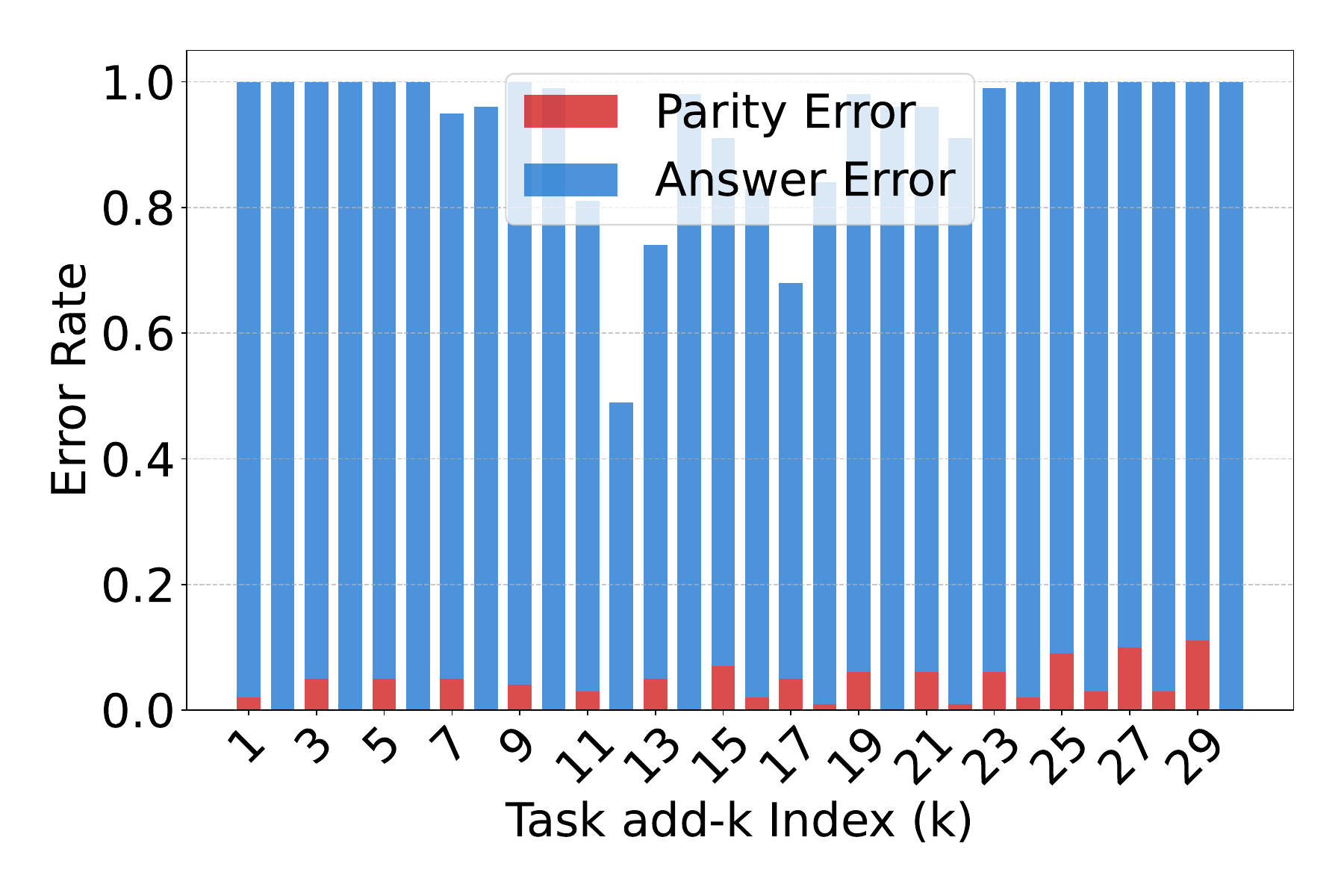}
        \caption{Head 2 onto parity direction}
    \end{subfigure}
    \hfill
    \begin{subfigure}{0.48\linewidth}
        \includegraphics[width=\linewidth]{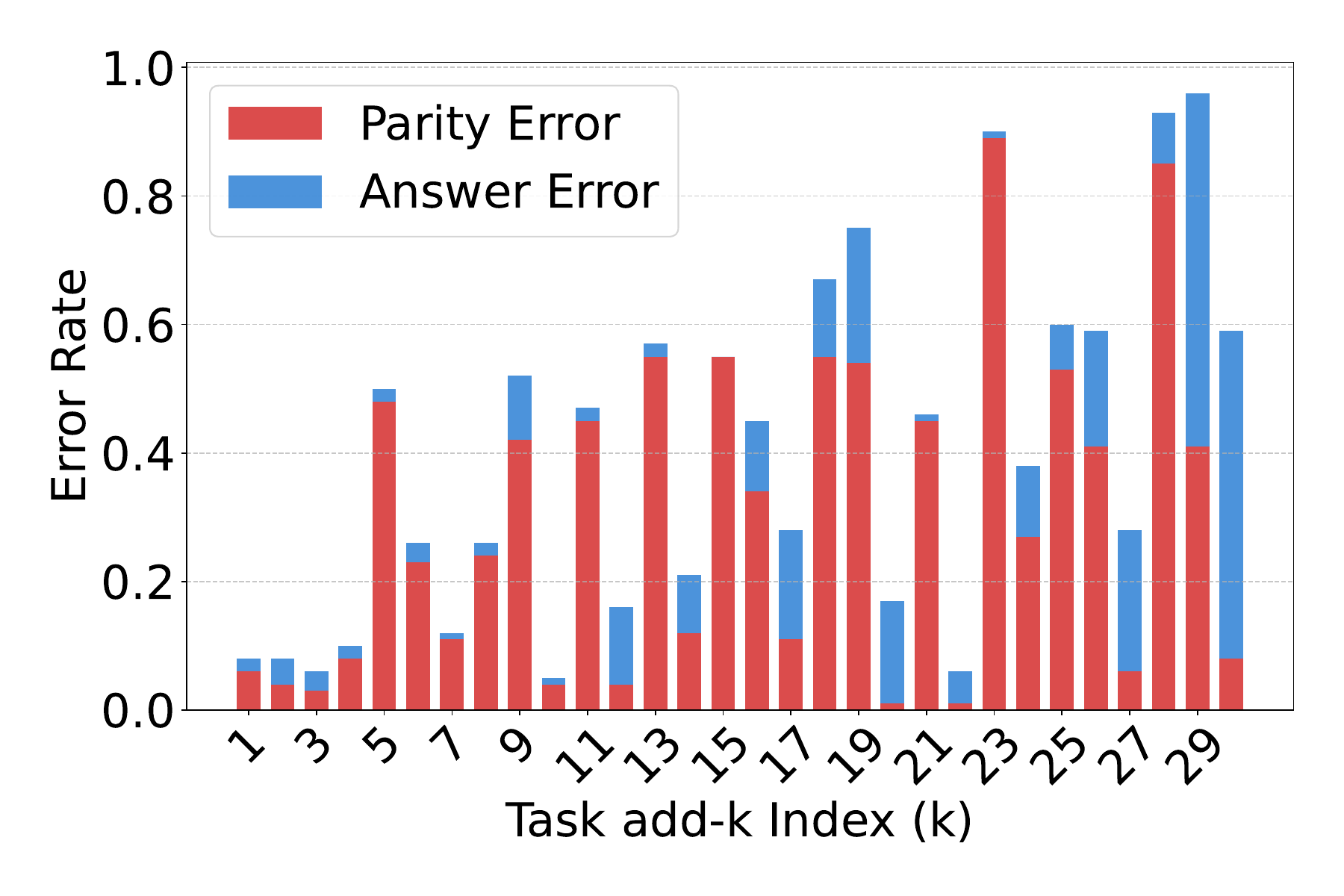}
        \caption{Head 2 out of parity direction}
    \end{subfigure}

    \vspace{1em}
    \begin{subfigure}{0.48\linewidth}
        \includegraphics[width=\linewidth]{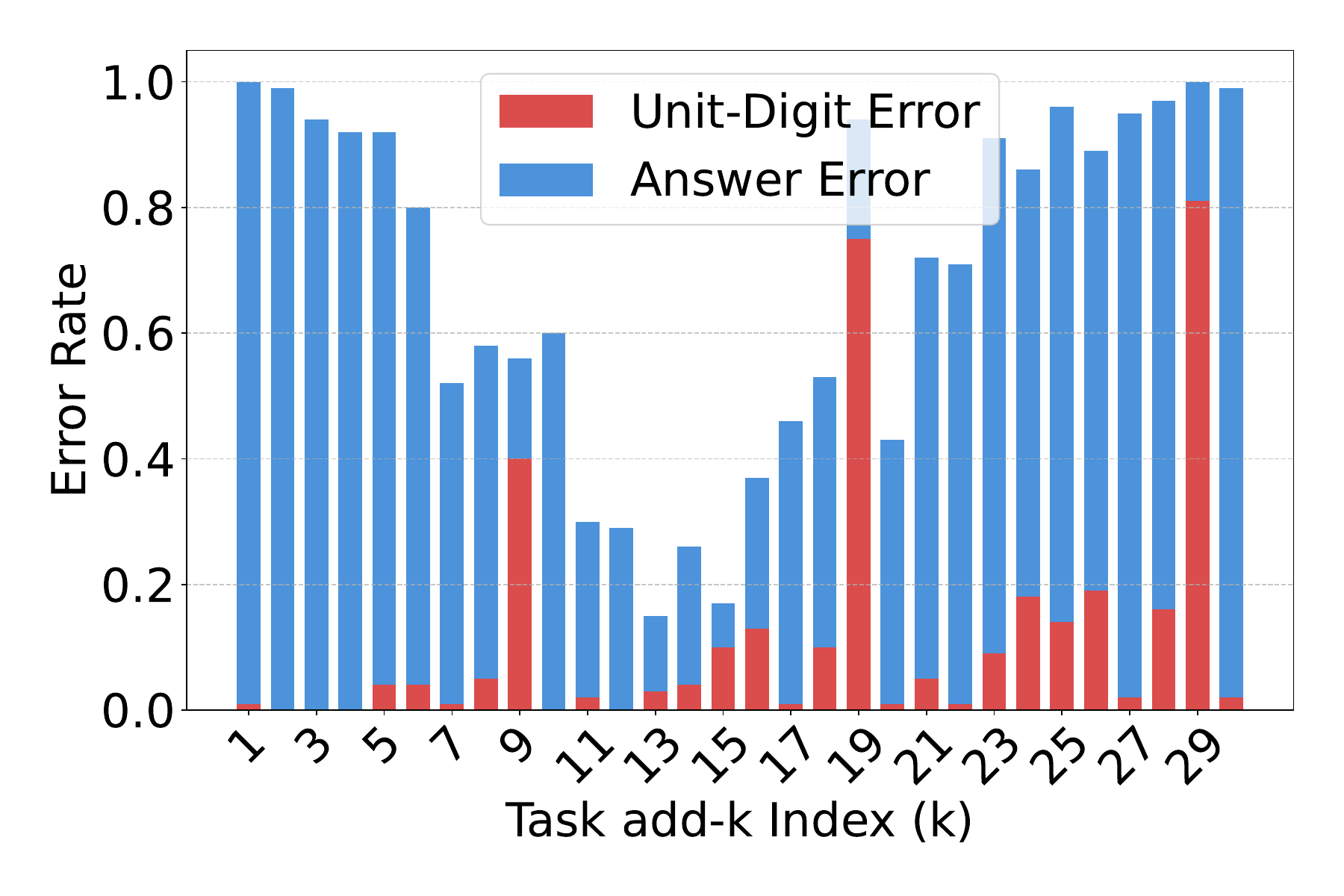}
        \caption{Head 2 onto unit subspace}
    \end{subfigure}
    \hfill
    \begin{subfigure}{0.48\linewidth}
        \includegraphics[width=\linewidth]{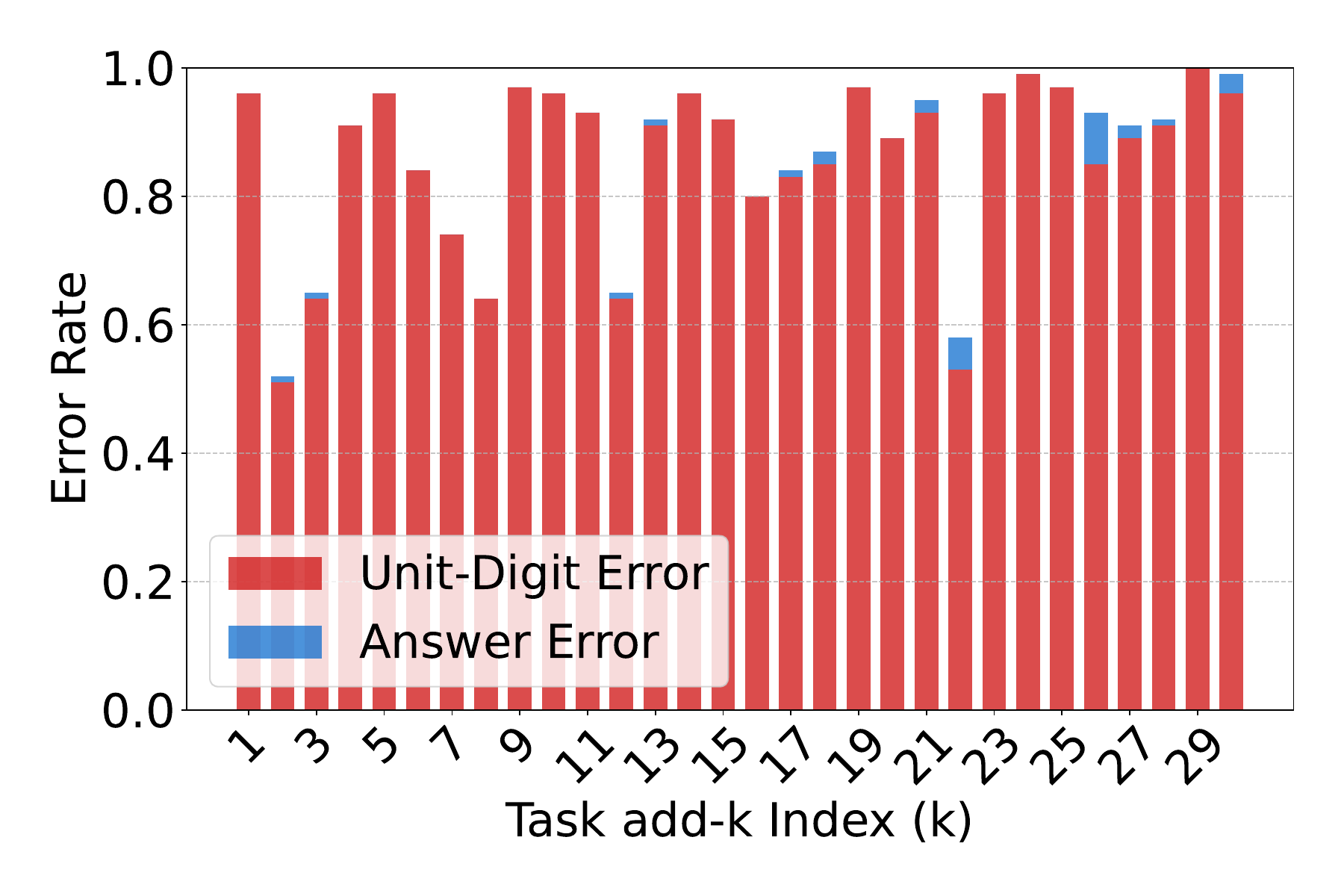}
        \caption{Head 2 out of unit subspace}
    \end{subfigure}

    \vspace{1em}
    \begin{subfigure}{0.48\linewidth}
        \includegraphics[width=\linewidth]{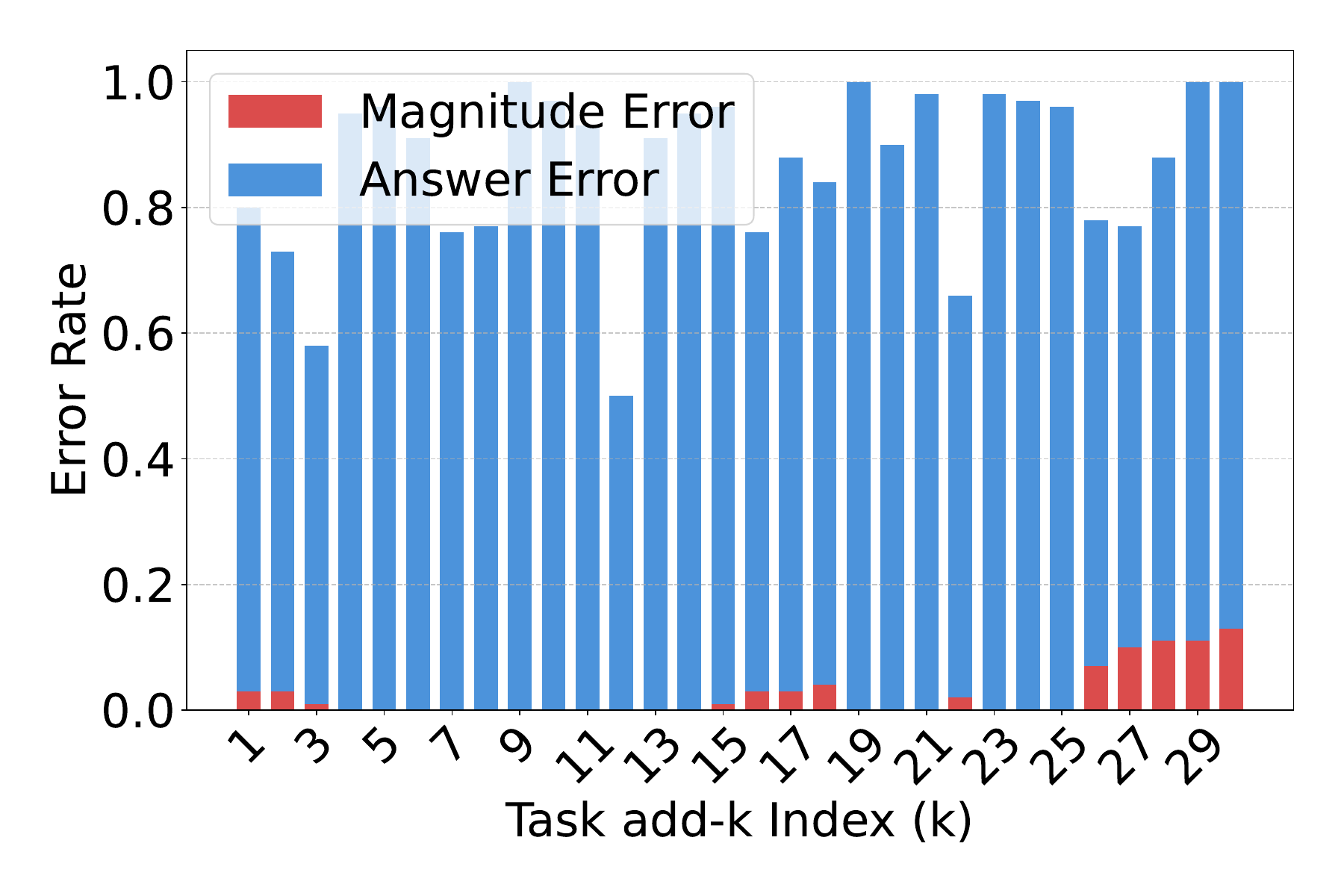}
        \caption{Head 2 onto magnitude subspace}
    \end{subfigure}
    \hfill
    \begin{subfigure}{0.48\linewidth}
        \includegraphics[width=\linewidth]{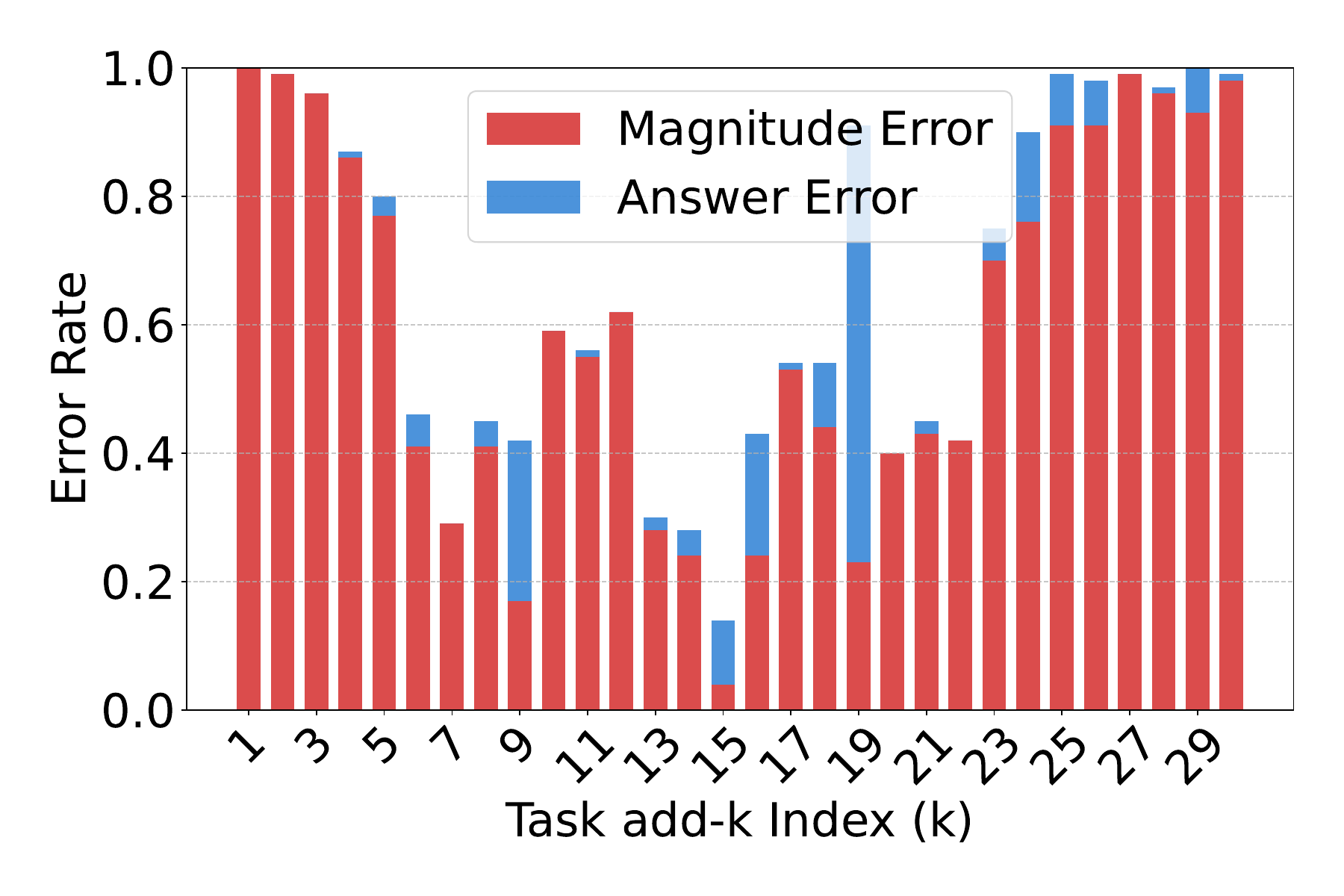}
        \caption{Head 2 out of magnitude subspace}
    \end{subfigure}

    \caption{Validation of hypotheses (i)–(iii) for head 2. Each row shows the projection effects for the parity, unit, and magnitude subspaces, respectively.}
    \label{fig:head2_combined}
\end{figure}

\begin{figure}[htbp]
    \centering
    \begin{subfigure}{0.48\linewidth}
        \includegraphics[width=\linewidth]{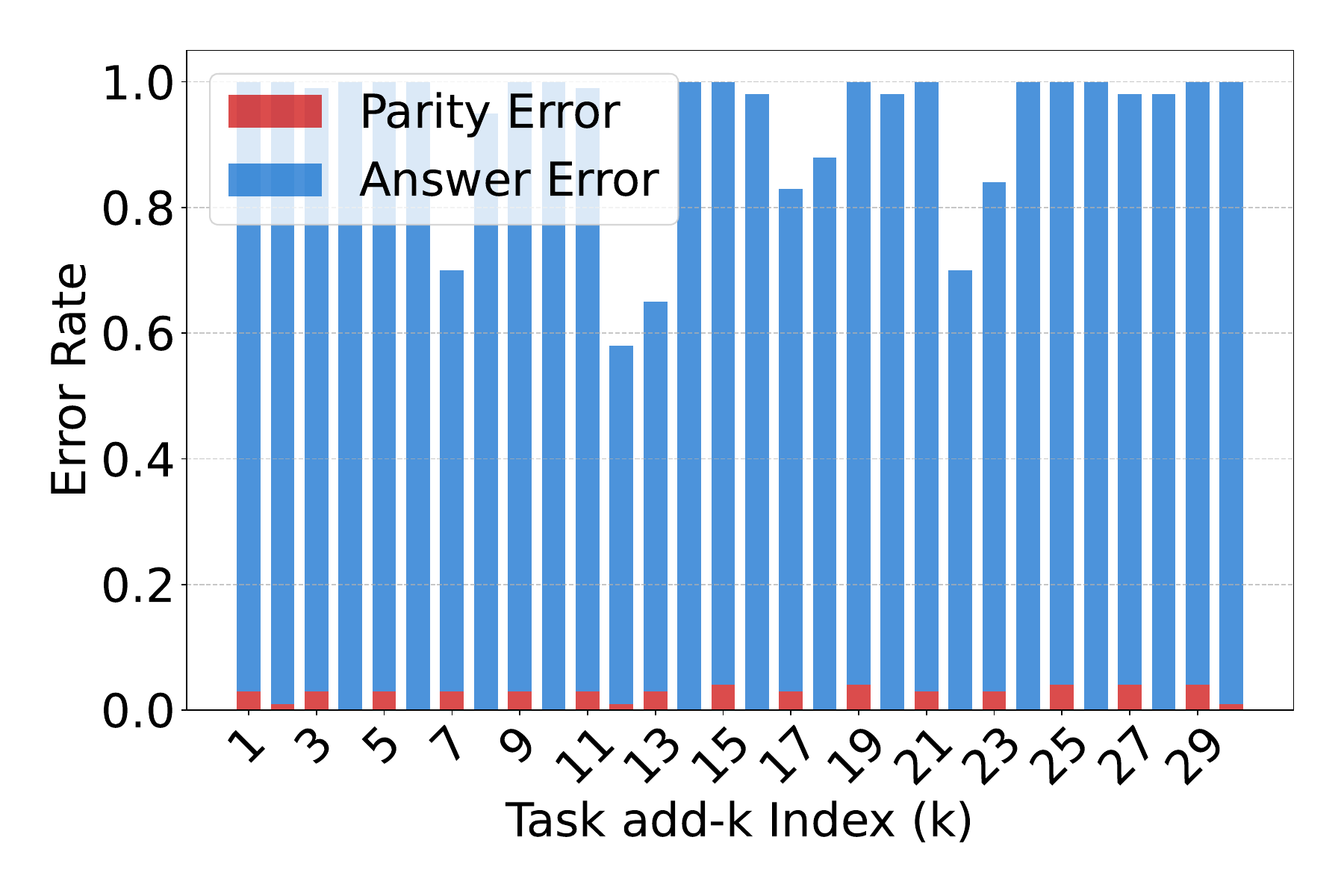}
        \caption{Head 3 onto parity direction}
    \end{subfigure}
    \hfill
    \begin{subfigure}{0.48\linewidth}
        \includegraphics[width=\linewidth]{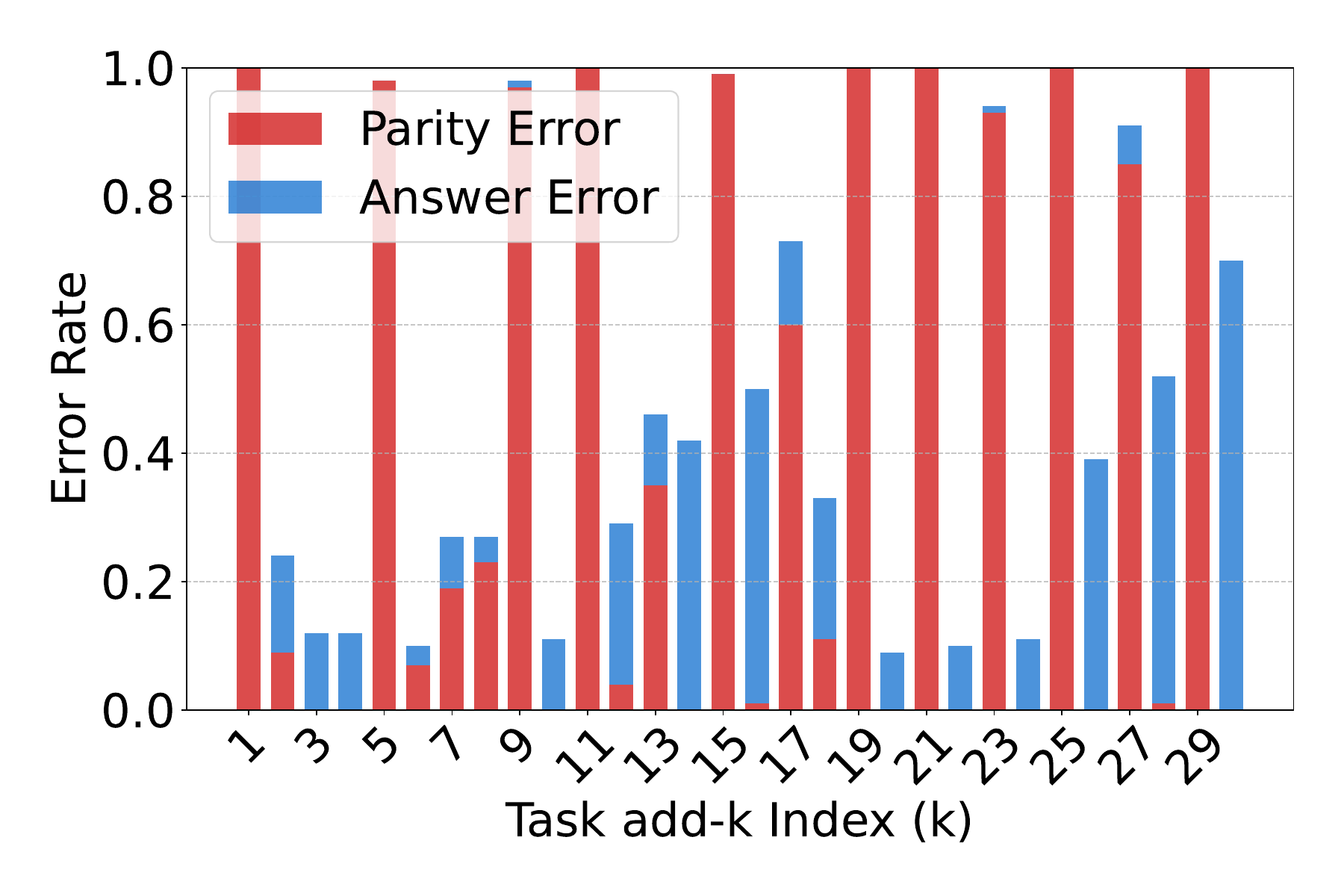}
        \caption{Head 3 out of parity direction}
    \end{subfigure}

    \vspace{1em}
    \begin{subfigure}{0.48\linewidth}
        \includegraphics[width=\linewidth]{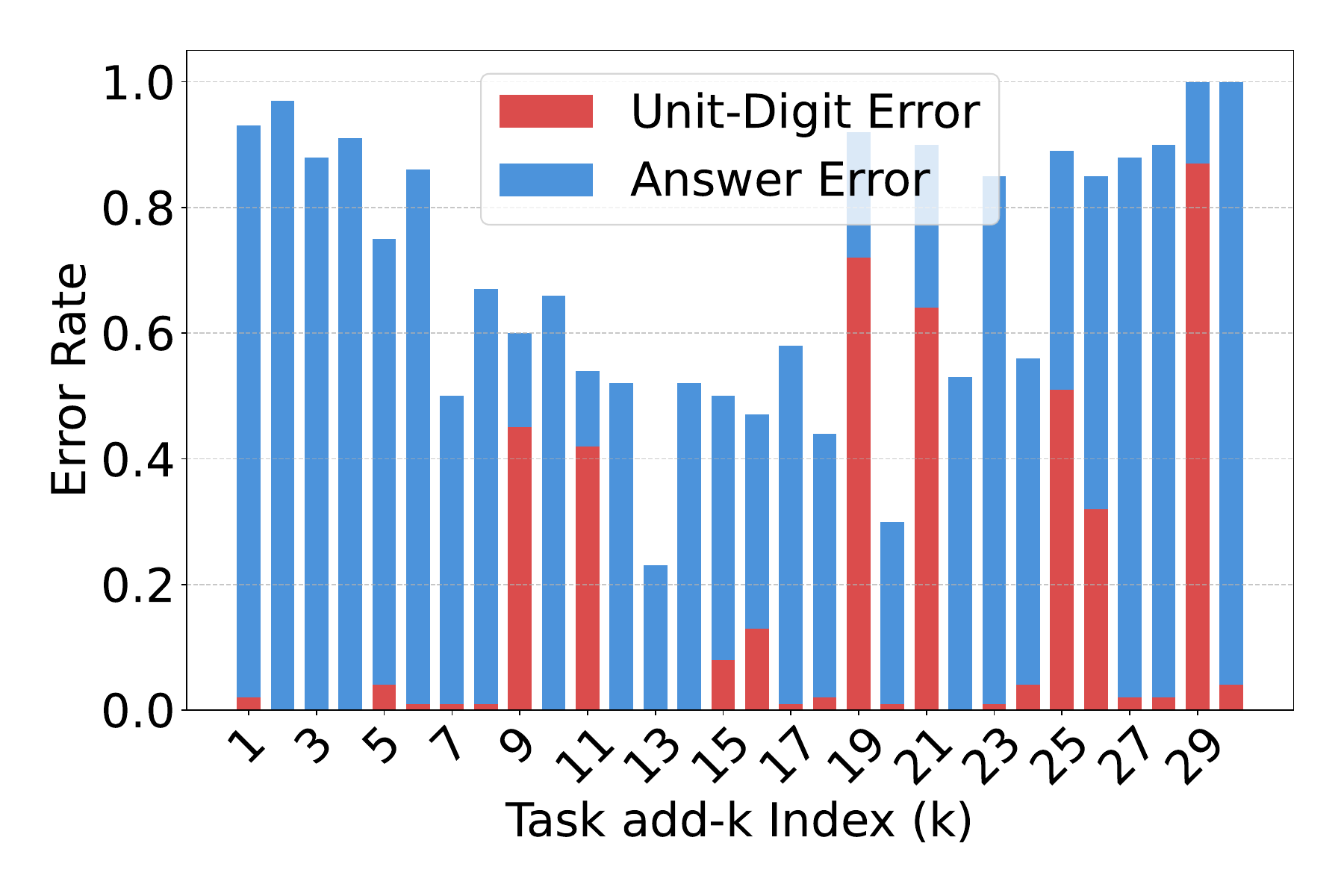}
        \caption{Head 3 onto unit subspace}
    \end{subfigure}
    \hfill
    \begin{subfigure}{0.48\linewidth}
        \includegraphics[width=\linewidth]{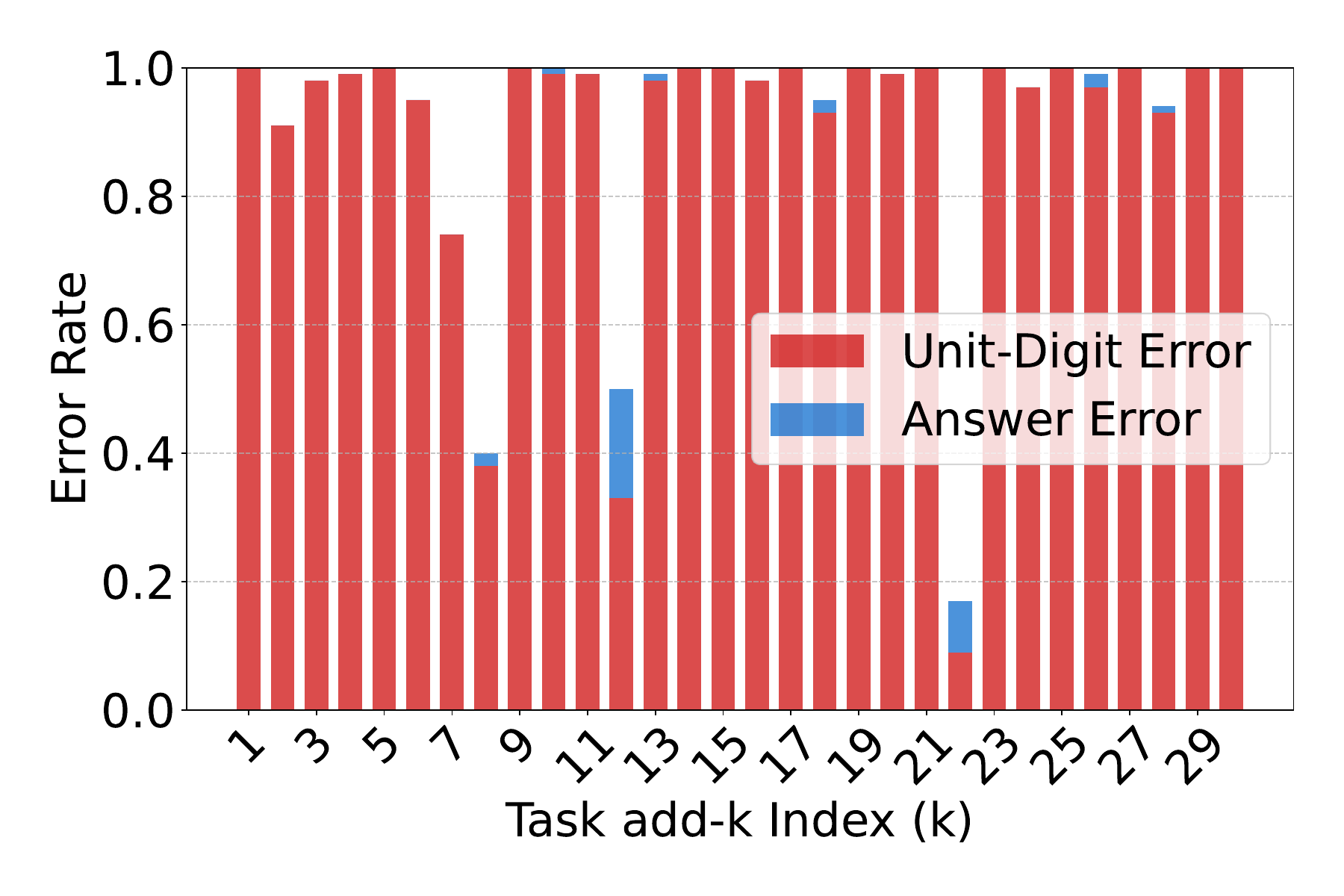}
        \caption{Head 3 out of unit subspace}
    \end{subfigure}

    \vspace{1em}
    \begin{subfigure}{0.48\linewidth}
        \includegraphics[width=\linewidth]{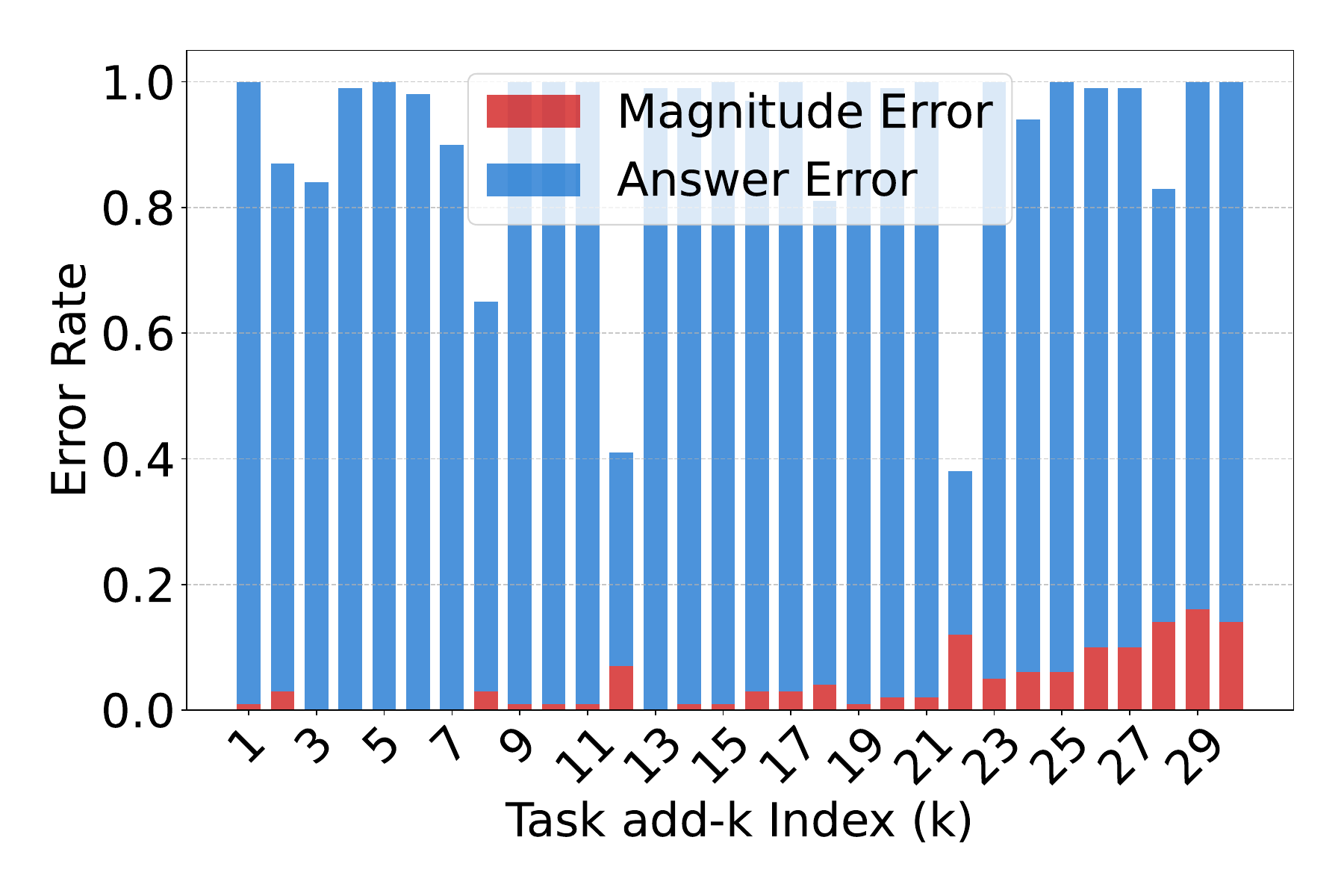}
        \caption{Head 3 onto magnitude subspace}
    \end{subfigure}
    \hfill
    \begin{subfigure}{0.48\linewidth}
        \includegraphics[width=\linewidth]{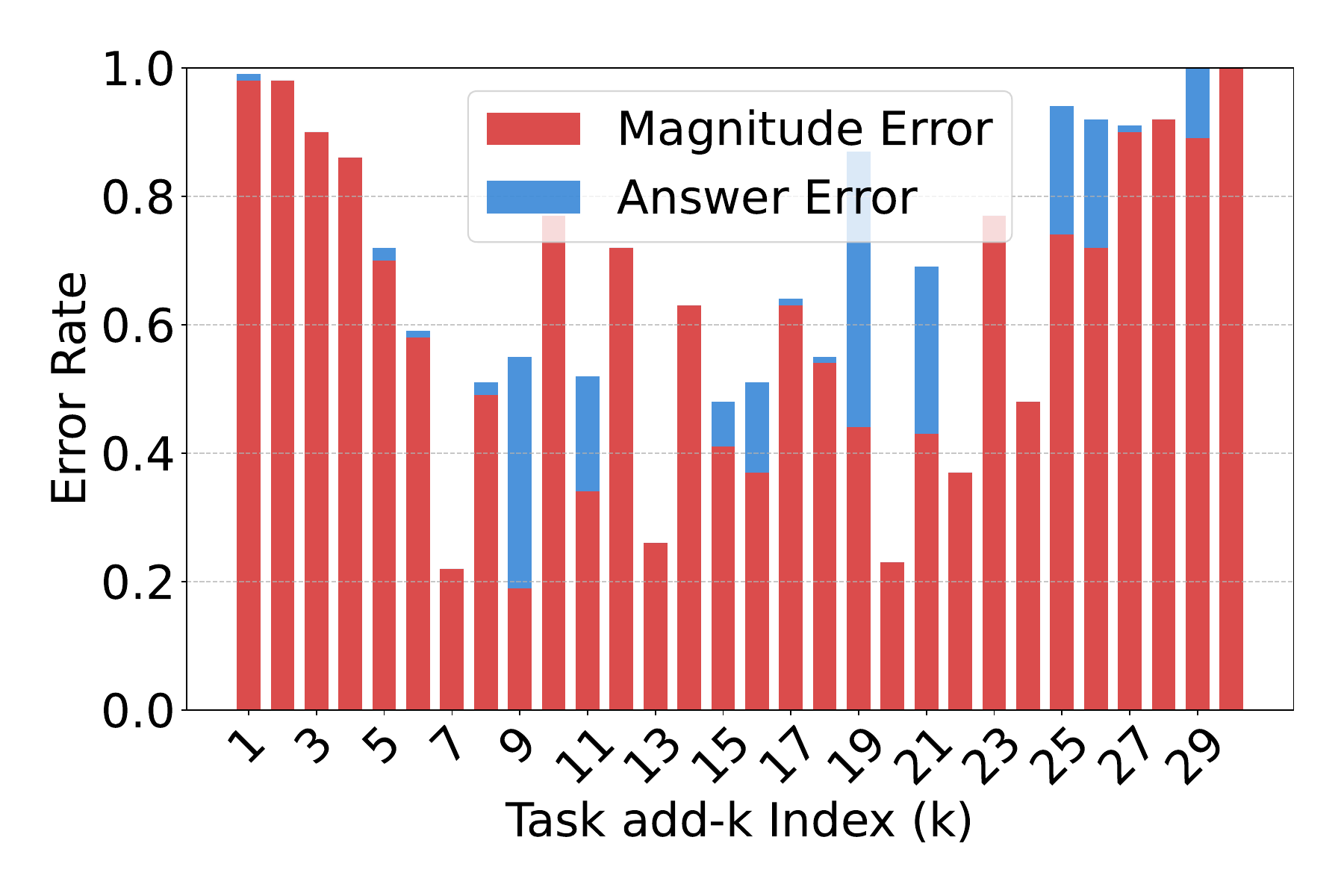}
        \caption{Head 3 out of magnitude subspace}
    \end{subfigure}

    \caption{Validation of hypotheses (i)–(iii) for head 3. The evidence is slightly weaker than for heads 1 and 2, consistent with head 3’s lower intervention accuracy (\S\ref{sec:3head}).}
    \label{fig:head3_combined}
\end{figure}

We first show the experimental results validating hypothesis (ii) for head 1 (Figure~\ref{fig:onto_out_unit}), and then show analogous results for hypotheses (i) and (iii) and for the other two heads (Figures~\ref{fig:head1_combined}--\ref{fig:head3_combined}). Projecting out of the parity direction does not substantially increase errors for either the parity or the final answer across all tasks, perhaps because parity is easy to obtain (e.g., random choice leads to $0.5$ accuracy) or parity is redundantly encoded elsewhere in the six-dimensional subspace.

\clearpage

\section{Supplement for \S\ref{sec:individual-extractor}}
This appendix provides the figures supporting \S\ref{sec:individual-extractor}.

\subsection{Additional figures for \S\ref{sec:individual-extractor}}\label{app: individual-extractor}

\begin{figure}[htbp]
    \centering
    \includegraphics[width=0.95\linewidth]{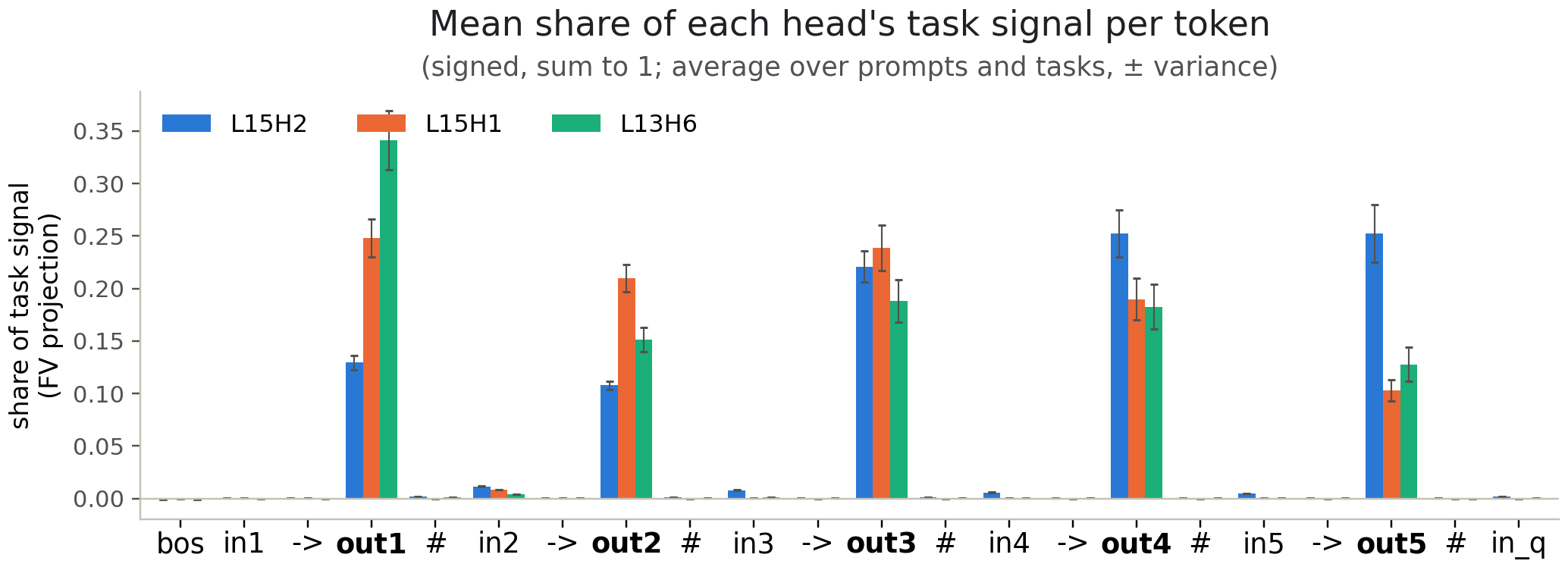}
    \caption{Mean share of each main head's task signal contributed by each token position on five-shot \textit{add-k} prompts (signed shares sum to one; averaged over prompts and the $25$ training tasks; error bars show variance). The demonstration label tokens $y_i$ (bold) carry essentially the entire task signal for all three main heads.}
    \label{fig:label_share}
\end{figure}

First, Figure~\ref{fig:label_share} reports the per-position decomposition of each main head's task signal on standard five-shot prompts.

\begin{figure}[htbp]
    \centering

    \begin{subfigure}{0.7\linewidth}
        \centering
        \includegraphics[width=\linewidth]{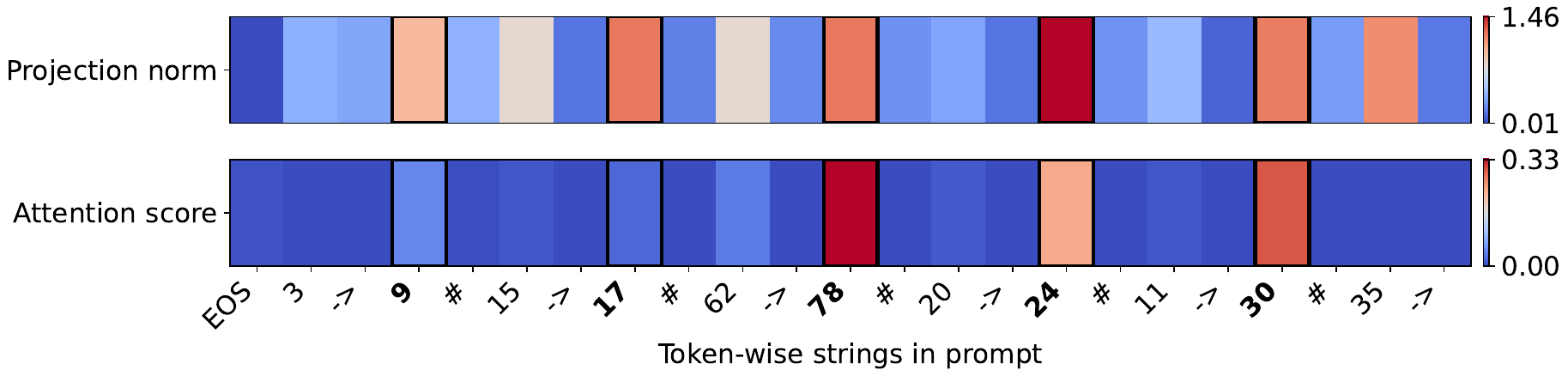}
        \includegraphics[width=\linewidth]{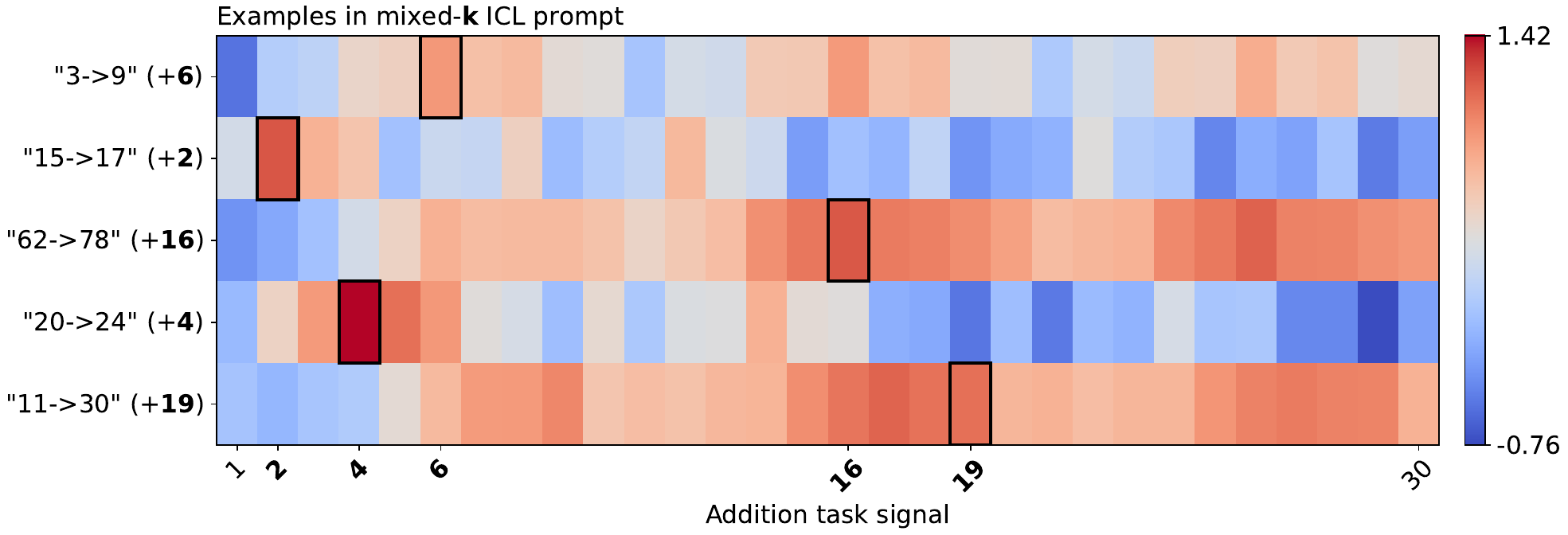}
        \caption{Head $(15,2)$}
        \label{fig:extractor_15_2}
    \end{subfigure}\hfill

    \begin{subfigure}{0.7\linewidth}
        \centering
        \includegraphics[width=\linewidth]{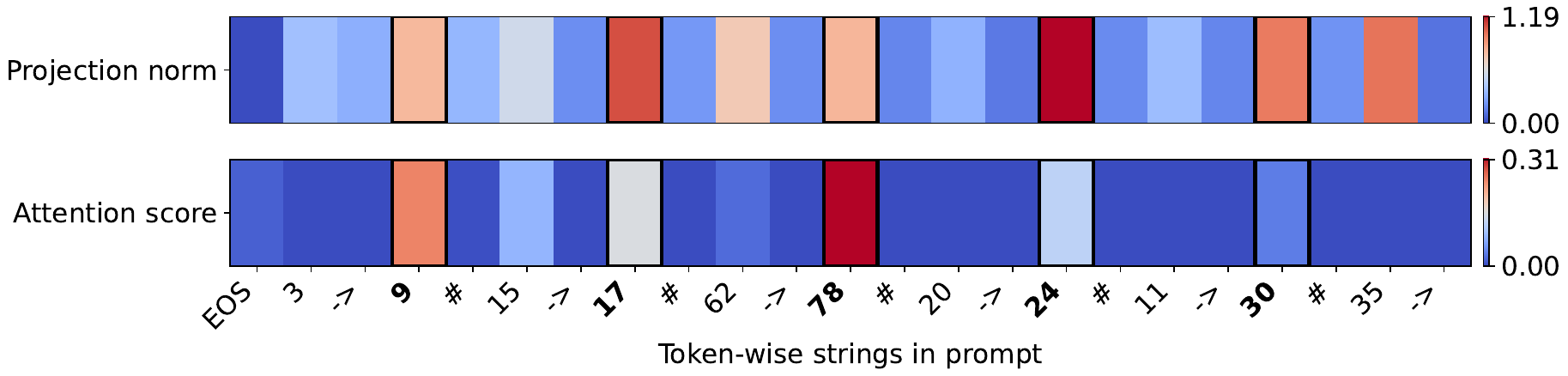}
        \includegraphics[width=\linewidth]{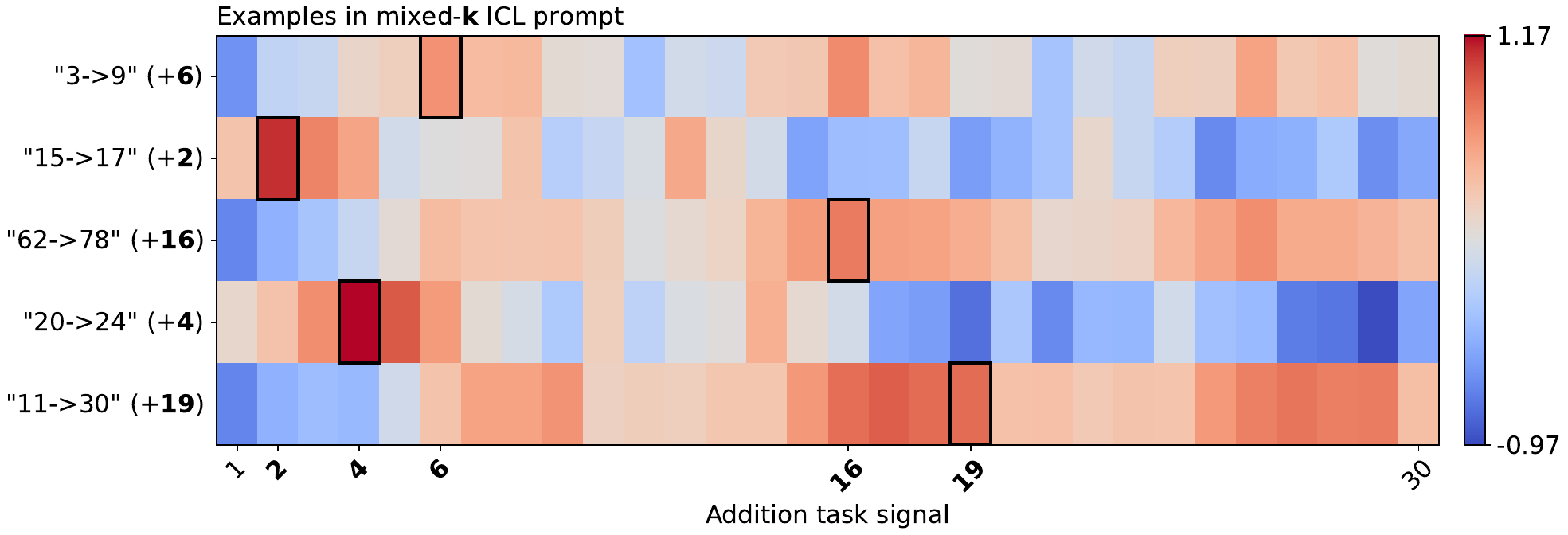}
        \caption{Head $(15,1)$}
        \label{fig:extractor_15_1}
    \end{subfigure}\hfill

    \begin{subfigure}{0.7\linewidth}
        \centering
        \includegraphics[width=\linewidth]{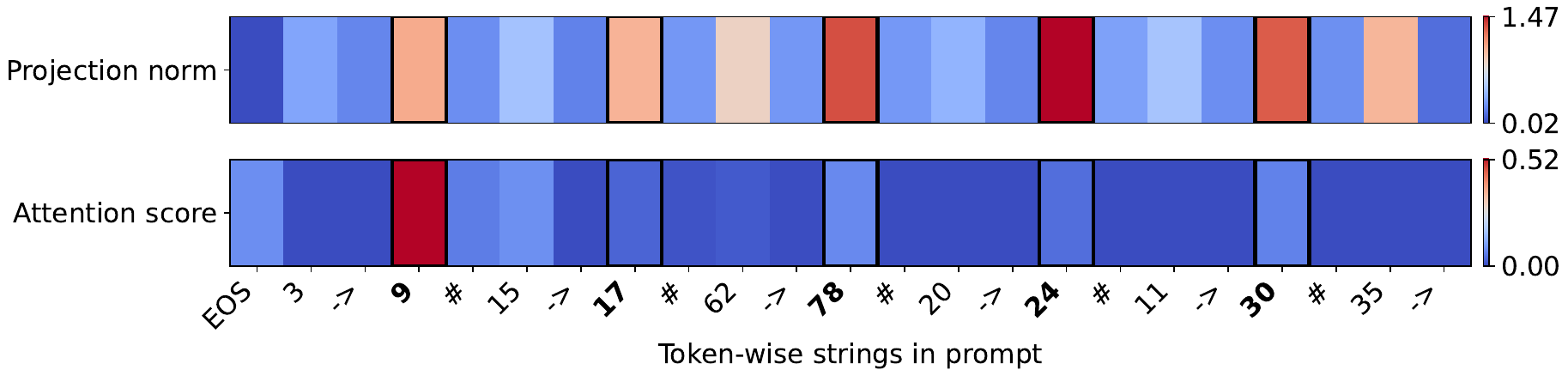}
        \includegraphics[width=\linewidth]{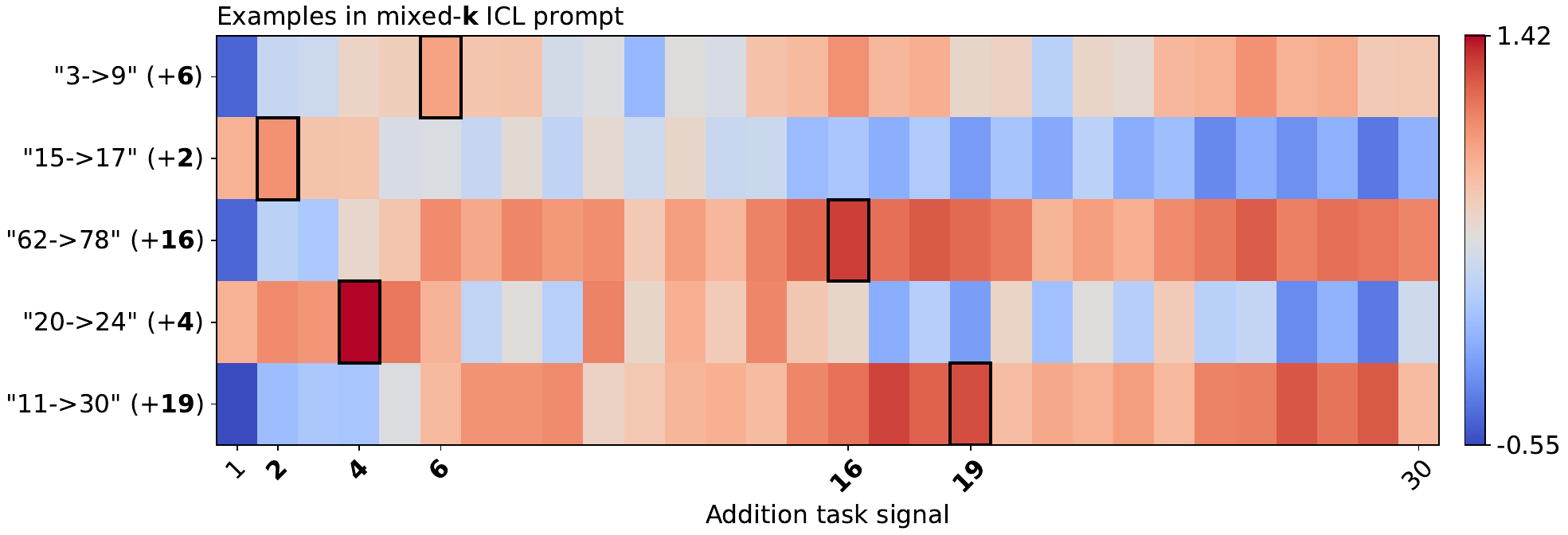}
        \caption{Head $(13,6)$}
        \label{fig:extractor_13_6}
    \end{subfigure}

    \caption{Llama-3-8B-Instruct.
    For each head, the \textbf{top panel} shows the two factors of each previous token's task-signal contribution $\alpha_t W_hO_hV_h z_t$: (1) the norm of the extracted information $\|W_hO_hV_h z_t\|$ and (2) the attention weight $\alpha_t$; both consistently peak at the tokens $t=y_i$ (in bold).
    The \textbf{bottom panel} shows the inner product between the projected signal from each $y_i$ and the head vector $\tilde{h}_k$ (normalized to unit norm), which peaks at $k=y_i-x_i$, indicating that each head extracts $y_i-x_i$ from its corresponding demonstration $x_i\!\to\!y_i$.}
    \label{fig:extractor_Llama-3_8b}
\end{figure}

Next, we turn to the signals extracted from each token of a mixed-$k$ ICL prompt (``3$\rightarrow$9 \# 15$\rightarrow$17 \# 62$\rightarrow$78 \# 20$\rightarrow$24 \# 11$\rightarrow$30 \# 35$\rightarrow$''). All three heads behave similarly: the extracted signal is strongest at the $y_i$ tokens, and its direction encodes $y_i-x_i$ (Figure~\ref{fig:extractor_Llama-3_8b}).

\clearpage

\end{document}